\documentclass[10pt,journal,compsoc]{IEEEtran}

\usepackage[utf8]{inputenc}
\usepackage[T1]{fontenc}
\usepackage{hyperref}
\usepackage{url}
\usepackage{booktabs}
\usepackage{amsfonts}
\usepackage{nicefrac}
\usepackage{microtype}
\usepackage{xcolor,color}
\usepackage{amsmath}
\usepackage{amssymb}
\usepackage{amsthm}
\usepackage{nicefrac}
\usepackage{bm}
\usepackage{multirow}
\usepackage{makecell}
\usepackage[ruled]{algorithm2e}
\usepackage{caption,subcaption}
\usepackage{enumitem}
\usepackage{mathtools}
\usepackage{wrapfig}
\usepackage[normalem]{ulem}
\usepackage{pifont}
\usepackage{wasysym}

\newcommand{\cmark}{\ding{51}}
\newcommand{\xmark}{\ding{55}}

\theoremstyle{plain}
\newtheorem{theorem}{Theorem}[section]
\newtheorem{proposition}[theorem]{Proposition}

\newtheorem{definition}[theorem]{Definition}

\DeclareMathOperator*{\argmin}{arg\,min}

\newcommand{\figref}[1]{Fig.~\ref{#1}}
\newcommand{\tabref}[1]{Tab.~\ref{#1}}
\newcommand{\secref}[1]{Sec.~\ref{#1}}

\newcommand{\algref}[1]{Alg.~\ref{#1}}
\newcommand{\equref}[1]{Equ.~(\ref{#1})}
\newcommand{\propref}[1]{Prop.~\ref{#1}}

\SetKwComment{Comment}{$\triangleright$\ }{}

\ifCLASSOPTIONcompsoc
  \usepackage[nocompress]{cite}
\else
  \usepackage{cite}
\fi

\begin{document}
\title{Selective Cross-Task Distillation}

\author{Su~Lu,
        Han-Jia~Ye,
        and~De-Chuan~Zhan
\IEEEcompsocitemizethanks{\IEEEcompsocthanksitem S. Lu, H.-J. Ye, and D.-C. Zhan are with State Key Laboratory for Novel Software Technology, Nanjing University, Nanjing, 210023, China.\protect\\
E-mail: \{lus, yehj\}@lamda.nju.edu.cn, zhandc@nju.edu.cn
}
\thanks{Manuscript received XXXXXX XX, 20XX; revised XXXXXX XX, 20XX.}}

\markboth{{IEEE} Transactions on Pattern Analysis and Machine Intelligence,~Vol.~XX, No.~X, XXXXXX~20XX}
{Lu \MakeLowercase{\textit{et al.}}: Selective Cross-Task Distillation}

\IEEEtitleabstractindextext{
\begin{abstract}
The outpouring of various pre-trained models empowers knowledge distillation by providing abundant teacher resources, but there lacks a developed mechanism to utilize these teachers adequately.
With a massive model repository composed of teachers pre-trained on diverse tasks, we must surmount two obstacles when using knowledge distillation to learn a new task.
First, given a fixed computing budget, it is not affordable to try each teacher and train the student repeatedly, making it necessary to seek out the most contributive teacher precisely and efficiently.
Second, semantic gaps exist between the teachers and the target student since they are trained on different tasks. Thus, we need to extract knowledge from a general label space that may be different from the student's.
Faced with these two challenges, we study a new setting named selective cross-task distillation that includes teacher assessment and generalized knowledge reuse.
We bridge the teacher's label space and the student's label space through optimal transport. The transportation cost from the teacher's prediction to the student's prediction measures the relatedness between two tasks and acts as an objective for distillation.
Our method reuses cross-task knowledge from a distinct label space and efficiently assesses teachers without enumerating the model repository.
Experiments demonstrate the effectiveness of our proposed method.
\end{abstract}

\begin{IEEEkeywords}
Knowledge Distillation, Model Reuse, Transfer Learning, Model Transferability
\end{IEEEkeywords}}

\maketitle

\IEEEpeerreviewmaketitle

\IEEEraisesectionheading{\section{Introduction}}
\label{Section:introduction}
\IEEEPARstart{K}{nowledge} distillation~\cite{KD-Survey} is a promising model reuse technique proposed by~\cite{KD}, and it has been proven to be effective in compressing models~\cite{Compression} and improving model performance~\cite{Gift}. The rapid spring of advanced deep learning algorithms~\cite{SimCLR,MoCo,SupCon} and network architectures~\cite{ResNet,Transformer,ViT,GPT-3,MLP_Mixer} brings the availability of plentiful pre-trained models. With these abundant teacher resources, there are opportunities for knowledge distillation to be applied to more practical applications. Under such a background, new challenges arise when we want to distill knowledge from a repository of pre-trained models to the maximum extent.

Consider product recognition systems~\cite{Product1,Product2} in supermarkets. Since different branches sell different goods, models in different branches have different class sets. When a new branch opens up, all existing models from other branches may be used as teachers to train the new model to improve learning efficiency. However, we do not know which is the most valuable, and existing models have different label spaces from our target task. These dilemmas reveal the insufficiency of standard knowledge distillation~\cite{KD} in two aspects. \textbf{(1)} Standard knowledge distillation often assumes that a teacher is given in advance. However, we must seek out the most contributive teacher efficiently with a repository of pre-trained models. \textbf{(2)} Standard knowledge distillation requires the label space of the teacher and the student to be identical, which is too strict to be satisfied for a teacher chosen from the massive repository.

A naive solution to \textbf{the first problem} is trying each model as the teacher to train the student repeatedly and scoring each teacher by the ground-truth accuracy of the corresponding student. Teacher selection now amounts to ranking all pre-trained models according to their scores. Since the model repository can be huge, it is not affordable to perform this operation in practice, and we need to rank all the pre-trained models without repetitive training. This problem is highly related to model transferability~\cite{Transferability,Transferability-Survey}. Existing works often design an efficient metric that evaluates the relatedness between a source task and the target task~\cite{NCE,LEEP,LogME,B-Tuning,Zoo-Tuning}, which enables teacher assessment without cumbersome training of the student. However, these methods are method-agnostic, i.e., the metrics are independent from the model reuse procedure, which makes the selected models sub-optimal.

\textbf{The second problem} is caused by support mismatch between the teacher's label distribution and the student's prediction. Lying at the core of standard knowledge distillation~\cite{KD} is the simple idea of distribution matching via KL divergence. This approach does not work when the teacher and the student target two different label spaces. An alternative is ignoring the teacher's classifier and reusing its embedding network to assist the student~\cite{RKD,Fitnets,ReFilled,MGD,WCoRD,SP,IRG}. However, the potential information contained in the teacher's predictions is wasted, resulting in an insufficient utilization of the pre-trained models.

\begin{figure}[t]
	\centering
	\includegraphics[width=\linewidth]{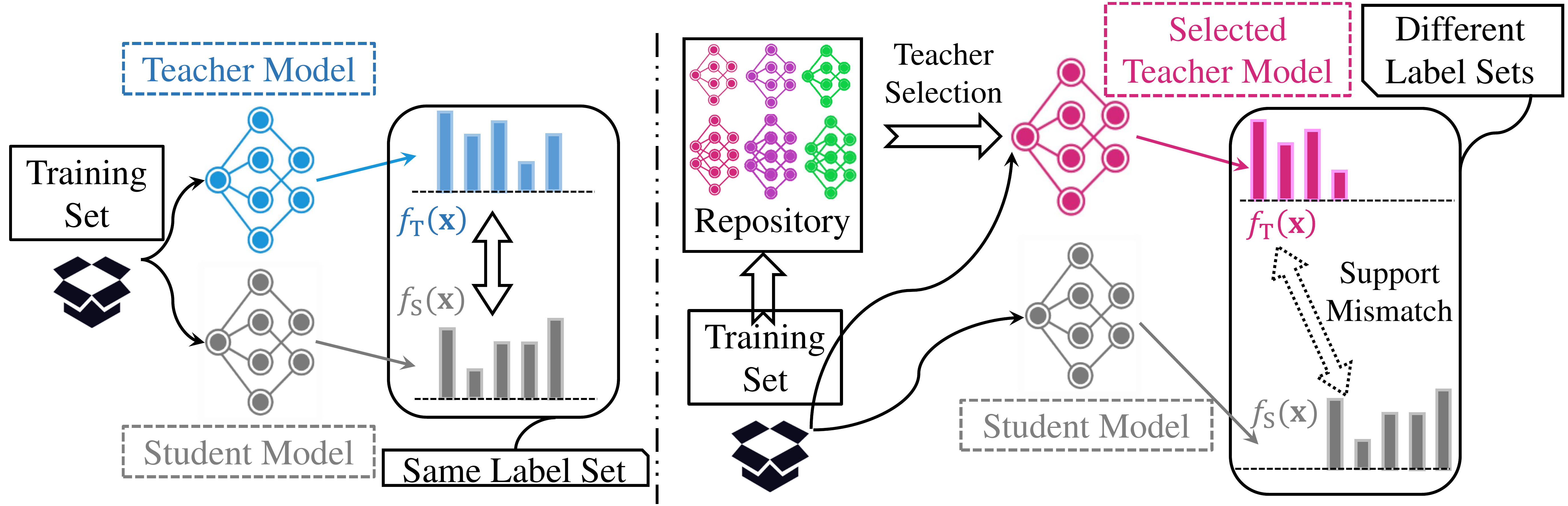}
	\caption{\small Comparison between two settings. \textbf{Left}: Standard knowledge distillation extracts knowledge from a given teacher and assists the training of a student. It requires the teacher and the student to target a same task. \textbf{Right}: When we have access to a group of teachers, we need to select one or several teachers from the repository and perform generalized knowledge reuse.}
	\label{Figure:kd_vs_sctd}
\end{figure}

To tackle these two problems, we study a new setting called ``selective cross-task distillation'' where we have a model repository containing a group of accessible teachers specializing in various tasks. Selective cross-task distillation contains two main phases. In the first phase, a student must select one or several teachers from the model repository. Intuitively, the chosen teacher should target a similar task as the student so that its supervision is contributive. The second phase is generalized knowledge reuse which means transferring the knowledge of the selected teacher to the target task even though they are pre-trained on different label sets. \figref{Figure:kd_vs_sctd} gives a comparison between standard knowledge distillation and our new setting.

This paper proposes a unifying objective that jointly solves the abovementioned problems. Although the teacher and the student target different classes, we can match their predictions by considering the relationship between the two class sets. This inspires us to utilize optimal transport~\cite{ComputationalOT,OT-ML} to bridge the semantic gap and measure the relatedness between a teacher and the target student. Specifically, we first construct a cost matrix between two class sets using the teacher's embedding network. We compute the semantic distance between two classes as the Euclidean distance between their corresponding class centers. The representations of class centers in both teacher's and student's class sets are obtained using the teacher's embedding network. This step reuses the comparison ability of the teacher's strong embedding network to capture the semantic similarities between different classes, which only requires two label spaces to be relevant but not identical~\cite{ReFilled,RKD,IRG,AML}. After that, we use Sinkhorn distance~\cite{SinkhornDistance} between the teacher's output and the student's output as an objective for teacher assessment and generalized knowledge reuse. In the first phase, we traverse the model repository to seek the most relevant teacher with the nearest Sinkhorn distance to the target student, and this process is free of model training and hence efficient. In the second phase, we optimize the student's parameters to pull it closer to the teacher in Sinkhorn distance, which successfully transfers the teacher's generalized knowledge to a distinct label space.

In the experiment part, we empirically verify the effectiveness of our method in teacher assessment and generalized knowledge reuse. For \textbf{the first problem}, it is shown that our method induces an evaluation metric that highly correlates to the student's performance. Thus, it efficiently picks out the most contributive teacher from the model repository. For \textbf{the second problem}, we test our method in both cross-task and standard knowledge distillation, and our method consistently improves the student's performance.

In summary, our contributions are threefold:
\begin{itemize}
    \item We study {\em a novel setting} called selective cross-task distillation, which tackles the problem of teacher assessment and generalized knowledge reuse simultaneously;
    \item We propose {\em an effective and efficient method} for selective cross-task distillation based on optimal transport;
    \item {\em Abundant experiment results} verify our claim points.
\end{itemize}

After the related literature and preliminary in \secref{Section:related} and \secref{Section:preliminary}, we give a detailed introduction and necessary explanations about our method in \secref{Section:reuse} and \secref{Section:assessment}. Finally, we list the experiment results and discussions in \secref{Section:experiment}, followed by the future work in \secref{Section:future}.

\section{Related Work}
\label{Section:related}
In this section, we describe three lines of works that are related to ours, i.e., knowledge distillation, optimal transport for knowledge reuse, and assessment of pre-trained models.

\begin{table*}[!t]
\centering
\caption{\small Comparison of the existing settings with our proposed setting, i.e., selective cross-task distillation. KD is short for knowledge distillation. $\mathcal{C}_{\mathrm{T}}$ and $\mathcal{C}_{\mathrm{S}}$ stand for the class set of teacher and student respectively.}
\begin{tabular}{@{}l|cc|cc|cc@{}}
\toprule
\multirow{2}{*}{Problem Setting}    & \multicolumn{2}{c|}{$\mathcal{C}_{\mathrm{T}}$ v.s. $\mathcal{C}_{\mathrm{S}}$}          & \multicolumn{2}{c|}{Teacher} & \multicolumn{2}{c}{Knowledge Type} \\
& $\mathcal{C}_{\mathrm{T}}=\mathcal{C}_{\mathrm{S}}$ & $\mathcal{C}_{\mathrm{T}}\neq\mathcal{C}_{\mathrm{S}}$ & Fixed        & Suitable       & Embedding & Classifier \\ \midrule
Standard KD with Logits~(e.g.,~\cite{KD})
& \cmark & \xmark & \cmark & \xmark & \cmark & \cmark \\
Standard KD with Representations~(e.g.,~\cite{Fitnets})
& \cmark & \xmark & \cmark & \xmark & \cmark & \xmark \\
Cross-Task KD~(e.g.,~\cite{ReFilled})
& \cmark & \cmark & \cmark & \xmark & \cmark & \xmark \\
KD with Specially Designed Teachers~(e.g.,~\cite{Nasty})
& \cmark & \xmark & \cmark & \cmark & \cmark & \cmark \\ \midrule
Selective Cross-Task Distillation~(ours)
& \cmark & \cmark & \cmark & \cmark & \cmark & \cmark \\ \bottomrule
\end{tabular}
\label{Table:setting}
\end{table*}

\subsection{Knowledge Distillation}
\label{Subsection:related_kd}
Knowledge distillation is an essential model reuse approach that extracts knowledge from a teacher and assists a student's training. It was first proposed in~\cite{KD} and has been proven to be an effective technique for improving student performance. Recent advances in knowledge distillation involve exploring new types of knowledge~\cite{Fitnets,ALP-KD,AT,AFD,RKD,ReFilled,Jacobian,VID}, developing new training styles~\cite{BAN,SSKD,TA,KD-SVD,MultiTeacher,SFTN,ONE,Snapshot,DML}, studying the reasons for its success~\cite{Understand1,Statistical,Understand2}, and applying it to new applications~\cite{Progressive,ST,Cross-Lingual}. This paper studies a new setting called selective cross-task distillation, which is highly related to but different from conventional knowledge distillation. Now we discuss \textbf{three key questions} to clarify the position of our work in the existing literature.

\textbf{(1) What is the relationship between the teacher's and student's tasks?} Most existing works~\cite{KD,Fitnets,RKD,VID,Jacobian} follow the ``same task assumption'', i.e., the teacher model is trained on the same task as the student, which restricts the application field of knowledge distillation. As an exception, several recent works~\cite{ReFilled,GKD,CrossDistill} have started to explore cross-task knowledge distillation where the same task assumption is replaced by a weaker ``relevant task assumption'', which means the teacher's task shares semantic relatedness with the student's task.

Recall that our motivation to study the new setting is to utilize the massive model repository to the maximum extent. Thus, we must enable cross-task distillation since it is unlikely to obtain a teacher from the repository that perfectly fits our target task. In this paper, we propose a method for generalized cases, which means the teacher's label space can be the same as, overlapped with, or totally different from the student's label space.

\textbf{(2) How to obtain a good teacher?} Knowledge distillation society often assumes the existence of a suitable teacher model. It may be a deeper neural network~\cite{KD,Fitnets,ReFilled,GKD} or a previous generation of the student model along the training process~\cite{BAN,Snapshot,LabelRefinery}. Despite the rapid development of knowledge distillation algorithms~\cite{TA,ONE,SFTN,SSKD}, how to obtain a good teacher remains an open problem, which raised attention recently~\cite{Nasty,Semiparametric}. \cite{Statistical} claimed that an ideal teacher should approximate the true label distribution. However, this insight hardly guides practice because the gap between the current teacher and the Bayesian optimal solution is unknown.

In real-world applications, sometimes we construct the teacher model by ourselves, e.g., model compression, but sometimes we just fetch a teacher from existing models. In this case, the student's performance is determined by which teacher we select. This paper tackles this selection problem through efficient model assessment. We aim to pick out the most contributive teacher from the repository without trying each one to train the student repeatedly. An alternative to teacher selection is directly distilling knowledge from the ensemble of all the teachers~\cite{ONE,Snapshot,FedDF}, which costs lots of computing resources and is not practical for large model repository. We will give a more detailed discussion in \secref{Section:assessment}.

\textbf{(3) What kind of knowledge is reused during distillation?} In standard knowledge distillation~\cite{KD} and other subsequent works~\cite{TA,DMC}, we reuse the teacher's soft labels to help the student via distribution matching. This method does not apply to cross-task knowledge distillation since the teacher's label set differs from the student's. Intermediate features~\cite{Fitnets,AT,LIT} and relationships between instances~\cite{RKD,IRG,ReFilled,GKD} are also considered dark knowledge and used for cross-task knowledge transfer. Although this kind of knowledge does not depend on labels and is easier to transfer across class sets, existing works neglect the potential supervision contained in the teacher's classifier~(head) and only focus on the embedding network, resulting in an insufficient utilization of the teacher model.

In our method, we simultaneously reuse the teacher's representations and predictions and close the semantic gap between teacher and student through optimal transport. We fuse two kinds of knowledge, i.e., instance representations and semantic labels, in a unifying objective for both teacher assessment and generalized knowledge reuse.  

To summarize, this paper studies a novel setting named selective cross-task distillation. We compare our setting to existing works in Table~\ref{Table:setting}. We can see that our setting adapts to both same-task and cross-task scenario, and simultaneously utilizes the teacher's embedding network and classifier. Moreover, the teacher is selected from a massive repository to fit the target task to the maximum extent.

\subsection{Optimal Transport for Knowledge Reuse}
\label{Subsection:related_ot}
Optimal transport theory~\cite{ComputationalOT} is widely adopted in many machine learning problems, including generative modeling~\cite{WGAN,ImproveGAN}, representation learning~\cite{WMD,DeepEMD}, and model fusion~\cite{OT-Fusion}. Despite the widespread application of optimal transport, we mainly focus on those works that use optimal transport for knowledge reuse and discuss the relationships and differences between our method and these works. In detail, two sub-fields of knowledge reuse are of interest, i.e., domain adaptation and representation distillation. These two sub-fields have the highest similarities with our setting. 

\textbf{(1) Domain Adaptation with Optimal Transport.} Domain adaptation~\cite{EasyDA,UniversalDA,TCA} is a particular case of transfer learning that utilizes labeled data in one or more related source domains to execute new tasks in a target domain~\cite{DA-Survey}. Under this setting, \cite{OTDA} and \cite{Deepjdot} transported source features to target features and trained a classifier for the target domain. Similarly, \cite{WDGRL} solves the optimal transport problem by the dual formulation. \cite{OT-DA1} matched both representations and semantic labels simultaneously, while~\cite{OT-DA2} is a direct extension of~\cite{OT-DA1} with multiple source domains. \cite{OT-DA3} proposed a new way to construct the cost matrix through solving a weighted Kantorovich problem to reduce the wrong pair-wise transport procedure. These works set the cost matrix in optimal transport as the distance between two representations to characterize the semantic relationships between instances, but the source and target distributions are naively set to uniform distributions, which is sub-optimal. In other words, they only use one key component in optimal transport (cost matrix) while ignoring the counterpart (source and target distributions).

\textbf{(2) Representation Distillation with Optimal Transport.} Representation distillation~\cite{Fitnets,AT,LIT,ReFilled,GKD,RKD,IRG} means reusing the teacher's embedding network to assist the student's learning. We have discussed this line of work in~\secref{Subsection:related_kd}. Generally speaking, the idea of using optimal transport in representation distillation is very similar to that of domain adaptation. The cost matrix is constructed to capture the relationships between the teacher's and student's representations. In~\cite{OT-Compression,MGD}, the authors minimized the transportation cost from teacher's representations to student's representations to pull them closer. \cite{WCoRD} extends this approach to contrastive learning. However, the information contained in the teacher's classifier is totally overlooked. This flaw makes it hard to use the source and target distributions in optimal transport.

Our method takes advantage of optimal transport more adequately compared to existing algorithms. In detail, we set the source and target distributions in optimal transport as the teacher's and student's predictions, and the Euclidean distances between class centers are used to estimate the ground metric. In this way, the teacher's representation ability is utilized to depict the correlation between class sets while its classifier directly acts as the source distribution. 

\subsection{Assessment of Pre-Trained Models.}
\label{Subsection:related_ptm}
More and more pre-trained models are available nowadays, and digging into the potential of these models is attracting increasing interest. Since fine-tuning is common practice in model reuse, a natural question arises, i.e., which pre-trained model will bring us the best performance on the target task after fine-tuning? Owing to the high computing expense, it is intractable to fine-tune all the pre-trained models to find the best one, which requires us to explore efficient approaches to evaluate the a model's transferability.

The most popular idea is to design a metric to assess the correlation between a pre-trained model and a target task. In~\cite{NCE,LEEP}, the authors first assigned pseudo labels to the instances from the target task by the source model. They then evaluated the transferability of the source model to the target task by the expected cross-entropy between the pseudo and true labels. \cite{N-LEEP} was an extension of~\cite{LEEP} that does not need the head classifier of the pre-trained model, and it used the cluster indexes generated by a Gaussian mixture model~\cite{GMM} to replace the pseudo labels. Thus, this method fits unsupervised scenarios and regression scenarios.  \cite{LogME} and~\cite{B-Tuning} are two recent works that built a proxy model on the source features and used the maximum evidence to assess a pre-trained model. \cite{GBC} directly checked the separability of source features by Bhattacharyya distance~\cite{B-Distance} to verify their benefit to the target task. Other similar papers include~\cite{OTCE}, which simultaneously considered domain and task differences between the source model and the target task. \cite{WhichModel} gave a systematic summary of algorithms on pre-trained model assessment.

In this paper, we study a novel setting called selective cross-task distillation, where we need to find the most contributive teacher from a model repository. This step is similar to model assessment. However, our method differs from the works listed above in two main aspects. \textbf{(1)} The methods to reuse a pre-trained model are different. Existing works mainly focus on model fine-tuning, but we study a more flexible setting, i.e., knowledge distillation. In knowledge distillation, teacher and student architectures and training algorithms can differ. In fact, fine-tuning can be seen as a specific kind of self-distillation~\cite{BAN}. \textbf{(2)} All these heuristic metrics are method-agnostic. They are designed without considering the subsequent procedure of model reuse. Ideally, the metric for teacher selection should be relevant to the distillation algorithm. In this paper, teacher assessment and knowledge reuse are two coupled processes that both rely on the semantic gaps between label distributions.

\section{Preliminary}
\label{Section:preliminary}
In this section, we first describe some preliminary concepts about supervised classification and standard knowledge distillation. After that, we give the formulation of our proposed new setting, i.e., selective cross-task distillation.

\subsection{Supervised Classification}
\label{Subsection:preliminary_supervised}
In a $C$-class classification task, we denote the training set by $\mathcal{D}=\{(\mathbf{x}_i,\mathbf{y}_i)\}_{i=1}^{N}$, where $\mathbf{x}_i\in\mathbb{R}^{D}$ and $\mathbf{y}_i\in\{0,1\}^{C}$ are instances and the corresponding one-hot labels. A model $f(\mathbf{x}):\mathbb{R}^{D}\rightarrow\mathbb{R}^{C}$ receives an instance as input and outputs a $C$-dimensional logit vector. Taking deep neural network as an example, $f(\mathbf{x})$ can be written as $W^\top\phi(\mathbf{x})$ where $\phi(\mathbf{x}):\mathbb{R}^{D}\rightarrow\mathbb{R}^{d}$ is the feature extractor and $W\in\mathbb{R}^{d}\times C$ contains parameters of the classifier. We can empirically optimize some loss function $\ell$ on the training set as~\equref{Equation:classification}:
\begin{equation}
f^\star=\argmin_{f}\frac{1}{N}\sum_{i=1}^{N}\ell(\mathbf{y}_i,f(\mathbf{x}_i))\;.
\label{Equation:classification}
\end{equation}

\subsection{Standard Knowledge Distillation}
\label{Subsection:preliminary_kd}
If we already have a well-trained model $f_{\mathrm{T}}$ on label set $\mathcal{C}_{\mathrm{T}}$, we can extract ``dark knowledge'' from it~\cite{Understand1,Statistical} to assist the training of a new model $f_{\mathrm{S}}$ on label set $\mathcal{C}_{\mathrm{S}}$. The standard distillation deals with the situation where $\mathcal{C}_{\mathrm{T}}=\mathcal{C}_{\mathrm{S}}$. The most well-known formulation~\cite{KD} is~\equref{Equation:standard_kd}:
\begin{equation}
\min_{f_{\mathrm{S}}}\frac{1}{N}\sum_{i=1}^{N}\ell(\mathbf{y}_i,f_{\mathrm{S}}(\mathbf{x}_i))+\lambda\mathbb{KL}(\bm{\rho}_{\tau}(f_{\mathrm{T}}(\mathbf{x}_i)),\bm{\rho}_{\tau}(f_{\mathrm{S}}(\mathbf{x}_i)))\;.
\label{Equation:standard_kd}
\end{equation}
Here $\bm{\rho}_{\tau}$ is the softmax function with temperature $\tau$,
\begin{equation}
\bm{\rho}_{\tau}(f(\mathbf{x}))_{c}=\frac{\exp(f(\mathbf{x})_{c}/\tau)}{\sum_{c^\prime=1}^{C}\exp(f(\mathbf{x})_{c^\prime}/\tau)}\;,c\in[C]\;,
\end{equation}
$C$ is the number of classes and the dimension of $f(\mathbf{x})$, and $\lambda\geq 0$ in~\equref{Equation:standard_kd} is a hyper-parameter that balances two terms. This method considers the teacher's output class distribution as a soft label, and uses KL divergence to pull student's output closer to it. The teacher model usually has a larger capacity~\cite{TA}, and its output can contain more information than one-hot label such as class correlations and instance relationships~\cite{Understand2}. Trained with both instance labels and teacher's supervision, the student model can obtain a higher accuracy and converge faster.

\begin{algorithm}[t]
\caption{\small The whole process of our proposed method for selective cross-task distillation.}
\KwIn{Number of teachers $H$, Model repository $\{f_\mathrm{T}^h\}_{h=1}^{H}$, Training set of the student $\mathcal{D}=\{(\mathbf{x}_i,\mathbf{y}_i)\}_{i=1}^{N}$.}
\KwOut{Well-trained student $f_{\mathrm{S}}$.}
Obtain the index of optimal teacher $h^\star$ by teacher assessment \Comment*[r]{Described in \secref{Section:assessment}}
Fetch the selected teacher $f_{\mathrm{T}}^{h^\star}$\;
Train $f_{\mathrm{S}}$ on training set $\mathcal{D}$ with $f_{\mathrm{T}}^{h^\star}$ by generalized knowledge reuse \Comment*[r]{Described in \secref{Section:reuse}}
Return $f_{\mathrm{S}}$\;
\label{Algorithm:sctd}
\end{algorithm}

\subsection{Selective Cross-Task Distillation}
\label{Subsection:preliminary_sctd}
In selective cross-task distillation, we have a model repository that contains $H$ teachers $\{f_{\mathrm{T}}^{h}\}_{h=1}^{H}$ trained on diverse tasks. Let $\mathcal{C}_{\mathrm{T}}^{h}$ be the label set of $h$-th teacher, and $\mathcal{C}_{\mathrm{T}}^{h}\neq\mathcal{C}_{\mathrm{S}}$ usually holds. Our goal is to select one teacher from the model repository and reuse its general knowledge to assist the training of the student model $f_{\mathrm{S}}$. Ideally, the selected teacher will result in a student that has the lowest generalization error. However, we do not have access to the test set during training phase. Thus, we turn to optimize the student's performance on training set with a precisely chosen teacher, as shown in~\equref{Equation:sctd}:
\begin{equation}
\min_{h,f_{\mathrm{S}}}\frac{1}{N}\sum_{i=1}^{N}\ell(\mathbf{y}_i,f_{\mathrm{S}}(\mathbf{x}_i))+\lambda\mathbb{D}(f_{\mathrm{T}}^h(\mathbf{x}_i),f_{\mathrm{S}}(\mathbf{x}_i))\;.
\label{Equation:sctd}
\end{equation}
Two variables $h$ and $f_{\mathrm{S}}$ are coupled together in \equref{Equation:sctd}. Note that $h$ is the index of teacher, but it is unrealistic to enumerate $h\in[H]$ and solve \equref{Equation:sctd} repeatedly. Thus, we decompose selctive cross-task distillation into two sub-problems, i.e., teacher assessment and generalized knowledge reuse. Teacher assessment aims to use a practical metric to rank all the teachers efficiently and select the best one for the subsequent generalized knowledge reuse, and we expect the selected teacher to produce a good student. The core of generalized knowledge reuse is the implementation of $\mathbb{D}$, which matches two outputs in different label spaces.

In selective cross-task distillation, we first perform teacher assessment and then distill its generalized knowledge. The whole process of our proposed method for selective cross-task distillation is described in \algref{Algorithm:sctd}.

As we can see in \algref{Algorithm:sctd}, generalized knowledge reuse comes after teacher assessment and selection. However, in the rest of this paper, we first introduce generalized knowledge reuse in \secref{Section:reuse} and then study teacher assessment in \secref{Section:assessment}. We write the paper in a different order from the algorithm process because the teacher assessment metric is logically based on the distillation method.

\section{Generalized Knowledge Reuse}
\label{Section:reuse}
After introducing the basic concepts in \secref{Section:preliminary}, we are ready to present details of our method. In this section, we first discuss how to reuse the knowledge of a teacher that may have a different label space from the student. After that, we study the optimization properties of the proposed method.

\subsection{Sinkhorn Distance for Distribution Matching}
\label{Subsection:reuse_sinkhorn}
The main obstacle to matching the outputs of teacher and student is support mismatch, which makes KL divergence not applicable in generalized knowledge reuse. However, if we consider the semantic relationship between $\mathcal{C}_{\mathrm{T}}$ and $\mathcal{C}_{\mathrm{S}}$, the teacher's output can still be instructive. For example, suppose $f_{\mathrm{T}}$ classifies cat and dog while $f_{\mathrm{S}}$ differs tiger from dog. We can approximately use the teacher's predicted probability of an instance being a cat as a reference because cats and tigers share similar appearance characteristics.

Sinkhorn distance~\cite{SinkhornDistance} is a regularized variant of OT distance, and it is widely used for matching two distributions in various applications~\cite{OT-Embedding,LatentOT,OTCE} because it considers the metric space of probabilities. This property enables us to use Sinkhorn distance to measure the discrepancy between two predictions in different label spaces. To this end, a key component is the cost matrix that describes the similarities between semantic information encoded by probability dimensions. We first state the definition of Sinkhorn distance and then study the method to generate the cost matrix.

\subsubsection{Sinkhorn Distance}
\label{Subsubsection:reuse_sinkhorn_sinkhorn}
\begin{definition}[Sinkhorn Distance]
Let $\mathcal{P}_{R}$ be the set of $R$-dimensional probability simplexes, i.e., $\mathcal{P}_{R}=\{\mathbf{p}\in\mathbb{R}^{R}|p_r\geq 0, \sum_{r=1}^{R}p_r=1\}$. Let $\bm{\mu}\in\mathcal{P}_{R_1}$ and $\bm{\nu}\in\mathcal{P}_{R_2}$ be two discrete probability distributions. Define the set of transport polytopes as $\Pi(\bm{\mu},\bm{\nu})=\{T\in\mathbb{R}_{+}^{R_1\times R_2}|T\mathbf{1}_{R_2}=\bm{\mu},T^\top\mathbf{1}_{R_1}=\bm{\nu}\}$ which contains all legal transportation plans from $\bm{\mu}$ to $\bm{\nu}$. Let $\mathbb{H}(T)=-\sum_{m=1}^{R_1}\sum_{n=1}^{R_2}T_{mn}(\log T_{mn}-1)$ be discrete entropy of transportation plan $T$.\footnote{$\mathbb{H}(T)$ can also be defined as $-\sum_{m,n}T_{mn}\log T_{mn}$ equivalently since $\sum_{m,n}T_{mn}=1$. Our form induces a more convenient dual problem mathematically, which is easier for optimization analysis.} Given a cost matrix $M\in\mathbb{R}_{+}^{R_1\times R_2}$, Sinkhorn distance between $\bm{\mu}$ and $\bm{\nu}$ is defined as
\begin{equation}
S_{\epsilon}(\bm{\mu},\bm{\nu})=\min_{T\in\Pi(\bm{\mu},\bm{\nu})}\langle T,M\rangle-\epsilon\mathbb{H}(T)\;.
\label{Equation:sinkhorn_distance}
\end{equation} 
\label{Definition:sinkhorn_distance}
\end{definition}

In \equref{Equation:sinkhorn_distance}, $\epsilon$ is a hyper-parameter controlling the strength of regularization term $\mathbb{H}(T)$. This term smooths the objective and forces the transportation plan to spread over the space rather than focusing on a few dimensions~\cite{ComputationalOT}.

Let $\mathbf{p}_{\mathrm{T}}$ and $\mathbf{p}_{\mathrm{S}}$ be the teacher's and student's predicted label distributions of an instance. Computing $S_{\epsilon}(\mathbf{p}_{\mathrm{T}},\mathbf{p}_{\mathrm{S}})$ amounts to finding an optimal plan to transport the teacher's confidence in source class labels to the student's confidence in target class labels. 

\begin{figure}[t]
	\centering
	\includegraphics[width=\linewidth]{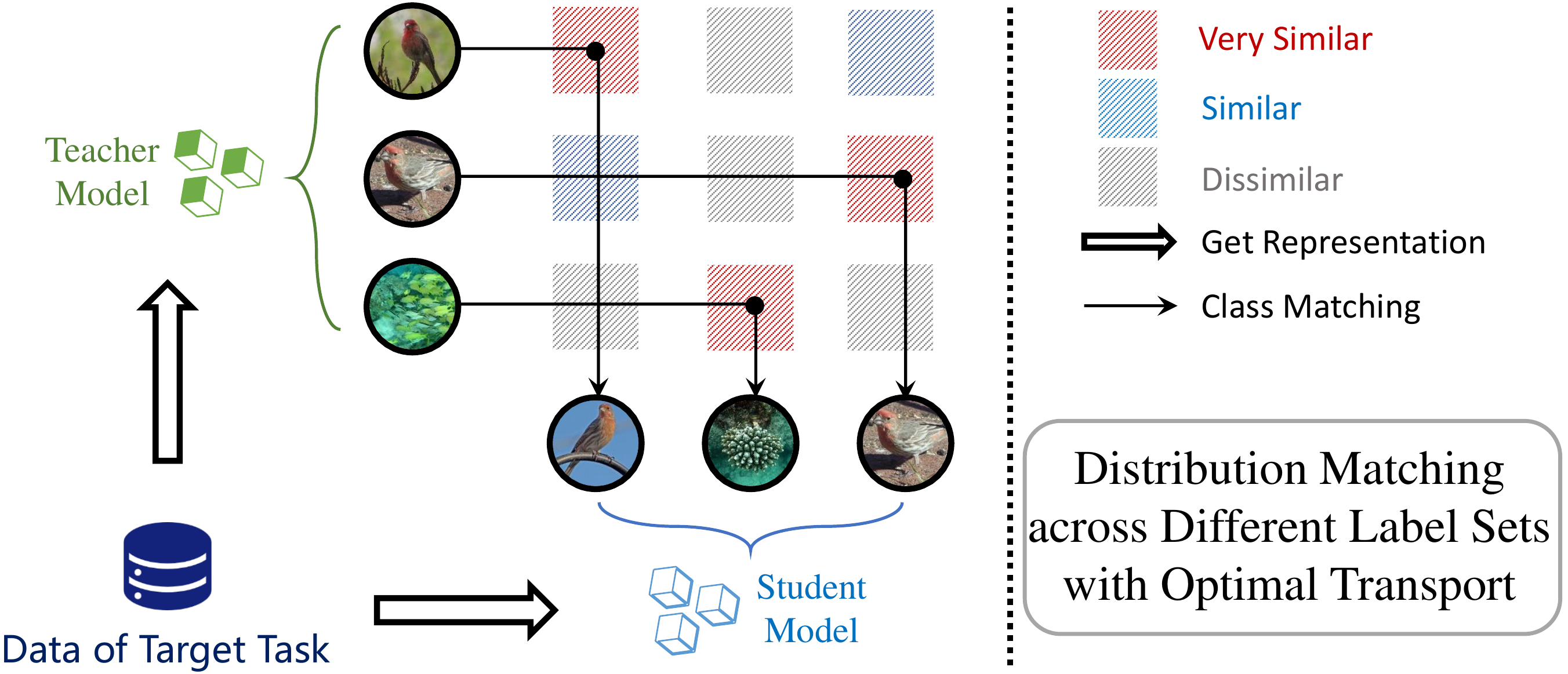}
	\caption{\small An illustration of the proposed method for distribution matching across two different label sets. Given a cost matrix that encodes the semantic distances between source classes and target classes, a dimension of the teacher's prediction tends to be transported to target dimensions with high similarities.}
	\label{Figure:ot_matching}
\end{figure}

An ideal cost matrix $M$ can capture the semantic distances between source classes and target classes. In other words, if the similarity between a source class and a target class is high, e.g., cats v.s. tigers, the corresponding value in $M$ is small. Thus, a source dimension will have priority to be transported to a target dimension corresponding to a semantically similar class since this leads to a lower transportation cost. If a teacher is related to current task, values in $M$ should be small in general, resulting in a small $S_{\epsilon}(\mathbf{p}_{\mathrm{T}},\mathbf{p}_{\mathbf{S}})$. On the other hand, the transportation cost will be very high is the task gap between the teacher and student cannot be ignored. \figref{Figure:ot_matching} shows the procedure of distribution matching across two different label sets.

\subsubsection{Construction of Cost Matrix}
\label{Subsubsection:reuse_sinkhorn_cost}
In our method, we reuse the teacher's representation network to get the class centers of both student's classes~(target classes) and teacher's classes~(source classes). Let $\phi_{\mathrm{T}}(\mathbf{x})$ be the representation network of teacher, $\mathcal{D}$ be the training set of student, and we can compute the student's class centers by averaging the instance representations from each class:
\begin{equation}
\begin{aligned}
\mathbf{e}_{\mathrm{S},n}=\frac{1}{|\mathcal{D}^{n}|}\sum_{(\mathbf{x},\mathbf{y})\in\mathcal{D}^{n}}\phi_{\mathrm{T}}(\mathbf{x})\;, \\
\mathcal{D}^{n}=\{(\mathbf{x},\mathbf{y})\in\mathcal{D}\mid \mathbf{y}(n)=1\}\;.
\end{aligned}
\label{Equation:center_student}
\end{equation}
Here $\mathbf{e}_{\mathrm{S},n}$ is the $n$-th class center in $\mathcal{C}_{\mathrm{S}}$. As for the teacher's class centers $\{\mathbf{e}_{\mathrm{T},m}\}$, we store them after training the teacher and directly fetch them to compute the cost matrix. After obtaining both $\{\mathbf{e}_{\mathrm{S},n}\}$ and $\{\mathbf{e}_{\mathrm{T},m}\}$, we can compute Euclidean distances between class centers in the embedding space to determine the values in the cost matrix $M$:
\begin{equation}
M_{mn}=\|\mathbf{e}_{\mathrm{T},m}-\mathbf{e}_{\mathrm{S},n}\|_{2},m\in\left[|\mathcal{C}_{\mathrm{T}}|\right],n\in\left[|\mathcal{C}_{\mathrm{S}}|\right]\;.
\label{Equation:cost_matrix}
\end{equation}

\textbf{Discussion on Teacher's Class Centers.} In \equref{Equation:cost_matrix}, we assume that the teacher's class centers $\{\mathbf{e}_{\mathrm{T}}\}$ are available, which to some extent disobeys the common practice that the student cannot visit the training data of the teacher. An alternative is using the teacher's classifier $W_{\mathrm{T}}\in\mathbb{R}^{d\times |\mathcal{C}_{\mathrm{T}}|}$~(weights of the last linear layer) to replace the class centers approximately,
\begin{equation}
\mathbf{e}_{\mathrm{T},m}=\mathrm{norm}\left(W_{\mathrm{T}}(:,m)\right),m\in\left[|\mathcal{C}_{\mathrm{T}}|\right]\;,
\label{Equation:center_teacher}
\end{equation}
where $\mathrm{norm}(\cdot)$ normalizes the corresponding vector. Now we explain the rationality of this approximation. When we want to classify an instance $\mathbf{x}$ with the teacher model $f_{\mathrm{T}}(\cdot)=W^\top_{\mathrm{T}}\phi_{\mathrm{T}}(\cdot)$,
there are two approaches:
\begin{itemize}
    \item \textbf{I.} Using the whole model. We first obtain the representation of $\mathbf{x}$ by $\phi_{\mathrm{T}}(\cdot)$ and then compute the inner product of $W_{\mathrm{T}}$ and $\phi_{\mathrm{T}}(\mathbf{x})$ to get the logits of instance $\mathbf{x}$. That is, the logit of the $m$-th class is $W_{\mathrm{T}}(:,m)^\top\phi_{\mathrm{T}}(\mathbf{x})$;
    \item \textbf{II.} Using the representation network and class centers. If we have computed the teacher's class centers $\{\mathbf{e}_{\mathrm{T},m}\}$, we can build a Nearest Class Mean~(NCM) classifier to put $\mathbf{x}$ into the category of its nearest class center~\cite{ProtoNet,MatchNet}. This means the logits of $\mathbf{x}$ are computed as $\{-\|\phi_{\mathrm{T}}(\mathbf{x})-\mathbf{e}_{\mathrm{T},m}\|_{2}^2\}$.
\end{itemize}
We expand the logit of the $m$-th class in the second method:
\begin{equation}
\begin{aligned}
  & -\|\phi_{\mathrm{T}}(\mathbf{x})-\mathbf{e}_{\mathrm{T},m}\|_2^2 \\
= & -\underbrace{\phi_{\mathrm{T}}(\mathbf{x})^\top\phi_{\mathrm{T}}(\mathbf{x})}_{\text{constant}} - \underbrace{\mathbf{e}_{\mathrm{T},m}^\top\mathbf{e}_{\mathrm{T},m}}_{\text{norm}} + \underbrace{2\mathbf{e}_{\mathrm{T},m}^\top\phi_{\mathrm{T}}(\mathbf{x})}_{\text{inner product}}\;.
\end{aligned}
\label{Equation:ncm}
\end{equation}
Given an instance $\mathbf{x}$, the label predicted by two approaches should be same intuitively since they are based on the same model. In the first approach, the $m$-th logit is computed as the inner product of $W_{\mathrm{T}}(:,m)$ and the instance representation. In \equref{Equation:ncm}, the second term only depends on the norm of the class center, and the third term is the inner product of the class center and the instance representation. Thus, if the norm of class center is fixed, the label predicted by the first approach is $\max_{m}W_{\mathrm{T}}(:,m)^\top\phi_{\mathrm{T}}(\mathbf{x})$, and the label predicted by the second approach is $\max_{m}\mathbf{e}_{\mathrm{T},m}^\top\phi_{\mathrm{T}}(\mathbf{x})$. Replacing class centers with $W_{\mathrm{T}}$ ensures the identity between two predicted labels, which explains the rationality of \equref{Equation:center_teacher}.

We further empirically verify our claim point. We train a WideResNet-(40,2)~\cite{WideResNet} on CIFAR-10~\cite{CIFAR} and report the test accuracies of three methods in \tabref{Table:test_method}. Accuracies of three methods are close to each other, which means our approximation of the teacher's class centers is reasonable.

\begin{table}[t]
\caption{\small Test accuracies of three different test methods, i.e., using the whole model~(\textbf{I}), using NCM with true class centers~(\textbf{II}), and using NCM with approximate class centers. More implementation details can be found in Appendix 1.1.}
\begin{tabular}{@{}l|ccc@{}}
\toprule
Test Method     & Whole Model & NCM - True & NCM - Approximate \\ \midrule
Accuracy ($\%$) & 95.15       & 94.47      & 94.53             \\ \bottomrule
\end{tabular}
\label{Table:test_method}
\end{table}

\textbf{Discussion on Cost Matrix.} (1) We do not require the teacher to cover the student's classes. A common case in knowledge distillation is that the teacher's task is related to the student's task. Although there often exist representation drifts between different class sets, $M$ only needs to describe relative relationships between classes, and we can transfer the teacher's comparison ability across classes~\cite{ReFilled,MatchNet,ProtoNet}. (2) Intuitively, it is better that the teacher shares a similar class set with the student. However, when the task gap between teacher and student is enormous, the student's performance will not degenerate compared to vanilla training. We empirically verify this claim point in \secref{Subsubsection:experiment_gkr_gap}.

\subsubsection{Summary}
\label{Subsubsection:reuse_sinkhorn_summary}
In summary, given a teacher model $f_{\mathrm{T}}(\cdot)=W_{\mathrm{T}}^\top\phi_{\mathrm{T}}(\cdot)$, we first construct the cost matrix $M$ by $\phi_{\mathrm{T}}(\cdot)$. The teacher's class centers $\{\mathbf{e}_{\mathrm{T},m}\}$ can be approximated by $W_{\mathrm{T}}$ while the student's class centers $\{\mathbf{e}_{\mathrm{S},n}\}$ are obtained through $\phi_{\mathrm{T}}$. The comparison ability of $\phi_{\mathrm{T}}$ is encoded in $M$ and used to capture semantic similarities between the teacher's classes and student's classes. Given the cost matrix $M$, we first get the teacher's and student's output probabilities,
\begin{equation}
\begin{aligned}
\mathbf{p}_{\mathrm{T}} & =\bm{\rho}_{\tau}(f_{\mathrm{T}}(\mathbf{x}_i))\;, \\
\mathbf{p}_{\mathrm{S}} & =\bm{\rho}_{\tau}(f_{\mathrm{S}}(\mathbf{x}_i))\;,
\end{aligned}
\label{Equation:prediction}
\end{equation}
and then minimize the Sinkhorn distance $S_{\epsilon}(\mathbf{p}_{\mathrm{T}},\mathbf{p}_{\mathrm{S}})$ to train the student with the assistance of the teacher model,
\begin{equation}
\min_{f_{\mathrm{S}}}\frac{1}{N}\sum_{i=1}^{N}\ell(\mathbf{y}_i,f_{\mathrm{S}}(\mathbf{x}_i))+\lambda S_{\epsilon}(\mathbf{p}_{\mathrm{T}},\mathbf{p}_{\mathrm{S}})\;,
\label{Equation:generalized_kd}
\end{equation}
where the Sinkhorn distance can be written as
\begin{equation}
S_{\epsilon}(\mathbf{p}_{\mathrm{T}},\mathbf{p}_{\mathrm{S}})=\min_{T\in\Pi(\mathbf{p}_{\mathrm{T}},\mathbf{p}_{\mathrm{S}})}\langle T,M\rangle-\epsilon\mathbb{H}(T)\;.
\label{Equation:sinkhorn_primal}
\end{equation}

\equref{Equation:generalized_kd} replaces KL divergence in standard knowledge distillation with Sinkhorn distance, bridging the support gap between $\mathbf{p}_{\mathrm{T}}$ and $\mathbf{p}_{\mathrm{S}}$. In our method, the teacher's representations and predictions are simultaneously reused, which is a more adequate utilization compared to feature-based methods~\cite{RKD,MGD,WCoRD,OT-Compression}. Experiments in \secref{Subsubsection:experiment_gkr_ctkd} further verify the superiority of our method. \algref{Algorithm:reuse} summarizes the procedure of generalized knowledge reuse.

\begin{algorithm}[t]
\caption{\small Generalized knowledge reuse process of our proposed method.}
\KwIn{A selected teacher $f_{\mathrm{T}}$, Training set of the student $\mathcal{D}=\{(\mathbf{x}_i,\mathbf{y}_i)\}_{i=1}^{N}$, Hyper-parameters $\epsilon$, $\tau$, and $\lambda$.}
\KwOut{Well-trained student $f_{\mathrm{S}}$.}
Randomly initialize $f_{\mathrm{S}}$\;
Compute the teacher's class centers with \equref{Equation:center_teacher}\;
Compute the student's class centers with \equref{Equation:center_student}\;
Compute the cost matrix $M$ with \equref{Equation:cost_matrix}\;
\While{not converge}
{
    $L_1\leftarrow\sum_{i=1}^{N}\ell(\mathbf{y}_i,f_{\mathrm{S}}(\mathbf{x}_i))$\;
    $L_2\leftarrow\sum_{i=1}^{N}S_{\epsilon}(\bm{\rho}_{\tau}(f_{\mathrm{T}}(\mathbf{x}_i)),\bm{\rho}_{\tau}(f_{\mathrm{S}}(\mathbf{x}_i)))$
    \Comment*[r]{Implemented as \algref{Algorithm:sinkhorn}}
    $L\leftarrow \frac{1}{N}\left(L_1 + \lambda L_2\right)$\;
    Use total loss $L$ to update $f_{\mathrm{S}}$\;
}
Return $f_{\mathrm{S}}$\;
\label{Algorithm:reuse}
\end{algorithm}

\subsection{Optimization}
\label{Subsection:reuse_optimization}
Now we study the optimization properties of \equref{Equation:generalized_kd}. It is a bi-level optimization problem~\cite{Bilevel} because it involves a nested optimization problem~(\equref{Equation:sinkhorn_primal}). Our target is learning the student, and we need to compute the gradient of $S_{\epsilon}$ w.r.t. model parameters through inner optimization. Moreover, solving the inner problem once can only provide a single update for the student, so it is essential to estimate the complexity of inner optimization. Thus, there are two questions we need to answer: (1) How to compute the gradient w.r.t. model parameters? (2) How much time will it take in the inner optimization to compute the gradient?

\begin{algorithm}[t]
\caption{\small Sinkhorn algorithm.}
\KwIn{Cost matrix $M$, Teacher's output probability distribution $\mathbf{p}_{\mathrm{T}}$, Student's output probability distribution $\mathbf{p}_{\mathrm{S}}$, Hyper-parameter $\epsilon$, Maximum number of iterations $I$.}
\KwOut{Sinkhorn loss $S_{\epsilon}(\mathbf{p}_{\mathrm{T}},\mathbf{p}_{\mathrm{S}})$.}
$\mathbf{u}\leftarrow\mathbf{1}_{|\mathcal{C}_{\mathrm{T}}|}$ \Comment*[r]{Initialize $\mathbf{u}$}
$\mathbf{v}\leftarrow\mathbf{1}_{|\mathcal{C}_{\mathrm{S}}|}$ \Comment*[r]{Initialize $\mathbf{v}$}
$K\leftarrow\exp\left(-\frac{M}{\epsilon}\right)$\;
$T\leftarrow\mathrm{diag}(\mathbf{u})K\mathrm{diag}(\mathbf{v})$\;
$i\leftarrow 0$\;
\While{not converge}
{
    $\mathbf{u}\leftarrow\mathbf{p}_{\mathrm{T}}./(K\mathbf{v})$ \Comment*[r]{Implement \equref{Equation:iteration}}
    $\mathbf{v}\leftarrow\mathbf{p}_{\mathrm{S}}./(K^\top\mathbf{u})$ \Comment*[r]{Implement \equref{Equation:iteration}}
    $T\leftarrow\mathrm{diag}(\mathbf{u})K\mathrm{diag}(\mathbf{v})$\;
    $i\leftarrow i+1$\;
    \If{$i \geq I$}
    {
        Break;
    }
}
Return $\langle T,M\rangle-\epsilon\mathbb{H}(T)$\;
\label{Algorithm:sinkhorn}
\end{algorithm}

\subsubsection{Gradient Computation}
\label{Subsubsection:reuse_optimization_gradient}
Computing the gradient of $S_{\epsilon}$ is intractable, and we solve this problem from the dual of \equref{Equation:sinkhorn_primal}, i.e., \equref{Equation:sinkhorn_dual}:
\begin{equation}
\max_{\bm{\alpha},\bm{\beta}}\bm{\alpha}^\top\mathbf{p}_{\mathrm{T}}+\bm{\beta}^\top\mathbf{p}_{\mathrm{S}}-\epsilon\sum_{m,n}\exp\left(\frac{\bm{\alpha}_{m}+\bm{\beta}_{n}-M_{mn}}{\epsilon}\right)\;.
\label{Equation:sinkhorn_dual}
\end{equation}
With the existence of the entropy regularization term $-\epsilon\mathbb{H}(T)$, the target function of \equref{Equation:sinkhorn_distance} is $\epsilon$-strictly convex, resulting in strong duality between \equref{Equation:sinkhorn_primal} and \equref{Equation:sinkhorn_dual}. Thus, the optimal values of \equref{Equation:sinkhorn_primal} and \equref{Equation:sinkhorn_dual} are identical. Let $\bm{\alpha}^\star$ and $\bm{\beta}^\star$ be any pair of optimal solution to \equref{Equation:sinkhorn_dual}, we can derive the gradient of $S_{\epsilon}(\mathbf{p}_{\mathrm{T}},\mathbf{p}_{\mathrm{S}})$:
\begin{equation}
\frac{\partial S_{\epsilon}(\mathbf{p}_{\mathrm{T}},\mathbf{p}_{\mathrm{S}})}{\partial\mathbf{p}_{\mathrm{S}}}=\frac{\partial {\bm{\alpha}^\star}^\top\mathbf{p}_{\mathrm{T}}+{\bm{\beta}^\star}^\top\mathbf{p}_{\mathrm{S}}}{\partial \mathbf{p}_{\mathrm{S}}}=\bm{\beta}^\star\;.
\label{Equation:gradient_ps}
\end{equation}
Applying chain rule to \equref{Equation:gradient_ps} further induces
\begin{equation}
\frac{\partial S_{\epsilon}(\mathbf{p}_{\mathrm{T}},\mathbf{p}_{\mathrm{S}})}{\partial f_{\mathrm{S}}(\mathbf{x})}=(\bm{\beta}^\star-\langle \bm{\beta}^\star,\mathbf{p}_{\mathrm{S}}\rangle)\odot\mathbf{p}_{\mathrm{S}}\;,\footnote{$\odot$ means element-wise multiplication.}
\end{equation}
which forms the following proposition.

\begin{proposition}[Gradient of Sinkhorn Distance]
Let $\mathbf{p}_{\mathrm{T}}=\bm{\rho}_{\tau}(f_{\mathrm{T}}(\mathbf{x}))$ and $\mathbf{p}_{\mathrm{S}}=\bm{\rho}_{\tau}(f_{\mathrm{S}}(\mathbf{x}))$ be the output probability distributions of teacher and student respectively. Let $\bm{\beta}^\star$ be the optimal solution to \equref{Equation:sinkhorn_dual}. The gradient of $S_{\epsilon}(\mathbf{p}_{\mathrm{T}},\mathbf{p}_{\mathrm{S}})$ w.r.t. $\mathbf{p}_{\mathrm{S}}$ and $f(\mathbf{x})$ can be written as $\frac{\partial S_{\epsilon}(\mathbf{p}_{\mathrm{T}},\mathbf{p}_{\mathrm{S}})}{\partial \mathbf{p}_{\mathrm{S}}}=\bm{\beta}^\star$ and $\frac{\partial S_{\epsilon}(\mathbf{p}_{\mathrm{T}},\mathbf{p}_{\mathrm{S}})}{\partial f(\mathbf{x})}=(\bm{\beta}^\star-\langle \bm{\beta}^\star,\mathbf{p}_{\mathrm{S}}\rangle)\odot\mathbf{p}_{\mathrm{S}}$ respectively.
\label{Proposition:sinkhorn_gradient}
\end{proposition}

\propref{Proposition:sinkhorn_gradient} answers our first question. To update the student's parameters, we need to obtain the optimal solution to \equref{Equation:sinkhorn_dual}. In the next part, we study the time complexity of our proposed method based on a classic algorithm.

\begin{figure}[t]
	\centering
	\includegraphics[width=\linewidth]{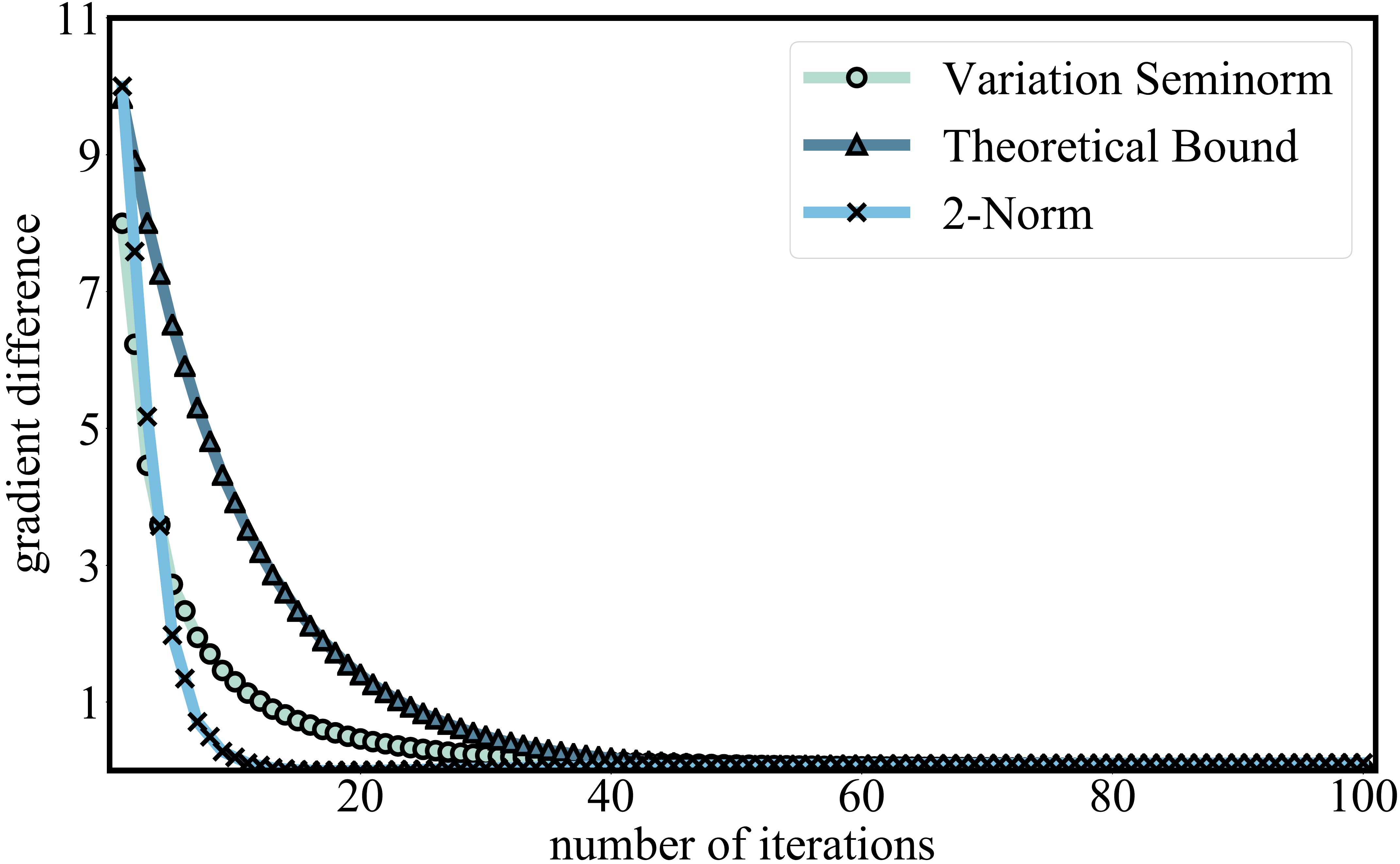}
	\caption{\small An empirical evaluation of \propref{Proposition:convergence}. We randomly sample $256$ instances from CIFAR-100 and form $256$ OT problems. We illustrate the convergence curves of $\|\nabla_{\mathbf{p}_{\mathrm{S}}}^{(t)}{\mathrm{S}}_{\epsilon}(\mathbf{p}_{\mathrm{T}},\mathbf{p}_{\mathrm{S}})-\nabla_{\mathbf{p}_{\mathrm{S}}}{\mathrm{S}}_{\epsilon}(\mathbf{p}_{\mathrm{T}},\mathbf{p}_{\mathrm{S}})\|_{\mathrm{var}}$, theoretical bound in \propref{Proposition:convergence}, and $\|\nabla_{\mathbf{p}_{\mathrm{S}}}^{(t)}{\mathrm{S}}_{\epsilon}(\mathbf{p}_{\mathrm{T}},\mathbf{p}_{\mathrm{S}})-\nabla_{\mathbf{p}_{\mathrm{S}}}{\mathrm{S}}_{\epsilon}(\mathbf{p}_{\mathrm{T}},\mathbf{p}_{\mathrm{S}})\|_{2}$. The values are averaged over $256$ problems. See Appendix 1.2 for details.}
	\label{Figure:convergence}
\end{figure}

\subsubsection{Complexity Analysis}
\label{Subsubsection:reuse_optimization_complexity}
In~\cite{SinkhornDistance}, the authors reformulated our target problem~(\equref{Equation:sinkhorn_dual}) as a matrix scaling problem and proposed a practical iterative algorithm, i.e., Sinkhorn algorithm, to solve it. We first review Sinkhorn algorithm proposed by~\cite{SinkhornDistance} and then study the complexity of using this algorithm to compute the gradient of $S_{\epsilon}(\mathbf{p}_{\mathrm{T}},\mathbf{p}_{\mathrm{S}})$.

The Lagrangian of \equref{Equation:sinkhorn_primal} can be written as
\begin{equation}
\mathcal{L}(T,\bm{\alpha},\bm{\beta})=\langle T,M\rangle-\epsilon\mathbb{H}(T)-\bm{\alpha}^\top(T\mathbf{1})-\bm{\beta}^\top(T^\top\mathbf{1})\;,
\label{Equation:lagrangian}
\end{equation}
where $\bm{\alpha}\in\mathbb{R}^{|\mathcal{C}_{\mathrm{T}}|}$ and $\bm{\beta}\in\mathbb{R}^{|\mathcal{C}_{\mathrm{S}}|}$ are Lagrangian multipliers. The first order condition of \equref{Equation:lagrangian} yields that
\begin{equation}
\frac{\partial\mathcal{L}(T,\bm{\alpha},\bm{\beta})}{\partial T_{mn}}=M_{mn}+\epsilon\log T_{mn}-\bm{\alpha}_{m}-\bm{\beta}_{n}=0\;,
\label{Equation:first_order_condition}
\end{equation}
which leads to the following expression:
\begin{equation}
T_{mn}=\exp(\bm{\alpha}_{m}/\epsilon)\exp(-M_{mn}/\epsilon)\exp(\bm{\beta}_{n}/\epsilon)\;.
\label{Equation:expression}
\end{equation}
Let $T_{\epsilon}^\star$ be the optimal solution to \equref{Equation:sinkhorn_primal}, we can derive the matrix form of $T_{\epsilon}^\star$ by \equref{Equation:expression}:
\begin{equation}
T_{\epsilon}^\star=\mathrm{diag}(\mathbf{u})K\mathrm{diag}(\mathbf{v})\;,
\end{equation}
where $K=\exp(-M/\epsilon)$, $\mathbf{u}=\exp(\bm{\alpha}/\epsilon)$, and $\mathbf{v}=\exp(\bm{\beta}/\epsilon)$.

Therefore, solving \equref{Equation:sinkhorn_primal} amounts to obtaining vectors $\mathbf{u},\mathbf{v}\succeq\mathbf{0}$. This can be solved through the Sinkhorn's fixed point iteration proposed by~\cite{Sinkhorn-Knopp}:
\begin{equation}
(\mathbf{u},\mathbf{v})\leftarrow(\mathbf{p}_{\mathrm{T}}./(K\mathbf{v}),\mathbf{p}_{\mathrm{S}}./(K^\top\mathbf{u}))\;.
\label{Equation:iteration}
\end{equation}
\algref{Algorithm:sinkhorn} is the scheme of this practical algorithm.
Each Sinkhorn iteration requires us to perform two matrix-vector multiplications, costing $\mathcal{O}(|\mathcal{C}_\mathrm{T}||\mathcal{C}_{\mathrm{S}}|)$ time. Another important question is how many iterations do we need to acquire an accurate gradient to update the student model. \propref{Proposition:convergence} gives the convergence rate of the approximate gradient of $S_{\epsilon}(\mathbf{p}_{\mathrm{T}},\mathbf{p}_{\mathrm{S}})$ w.r.t. $\mathbf{p}_{\mathrm{S}}$, whose proof is based on~\cite{ComputationalOT}.

\begin{proposition}[Convergence Rate of Gradient]
Let $\nabla_{\mathbf{p}_{\mathrm{S}}}^{(t)}S_{\epsilon}(\mathbf{p}_{\mathrm{T}},\mathbf{p}_{\mathrm{S}})$ be the approximate gradient w.r.t. the student's output probability after $t$ Sinkhorn iterations. Let $M$ be the cost matrix in \equref{Equation:sinkhorn_primal} and $K=\exp\left(-M/\epsilon\right)$. Vector sequence $\nabla_{\mathbf{p}_{\mathrm{S}}}^{(t)}S_{\epsilon}(\mathbf{p}_{\mathrm{T}},\mathbf{p}_{\mathrm{S}})$ has a linear convergence rate $\kappa(K)^2$ in variation seminorm, and $\kappa(K)\in[0,1)$ is a constant about $K$:
\begin{equation}
\frac
{\|\nabla_{\mathbf{p}_{\mathrm{S}}}^{(t+1)}S_{\epsilon}(\mathbf{p}_{\mathrm{T}},\mathbf{p}_{\mathrm{S}})-\nabla_{\mathbf{p}_{\mathrm{S}}}S_{\epsilon}(\mathbf{p}_{\mathrm{T}},\mathbf{p}_{\mathrm{S}})\|_{\mathrm{var}}}
{\|\nabla_{\mathbf{p}_{\mathrm{S}}}^{(t)}S_{\epsilon}(\mathbf{p}_{\mathrm{T}},\mathbf{p}_{\mathrm{S}})-\nabla_{\mathbf{p}_{\mathrm{S}}}S_{\epsilon}(\mathbf{p}_{\mathrm{T}},\mathbf{p}_{\mathrm{S}})\|_{\mathrm{var}}}\leq\kappa(K)^{2}\;.
\end{equation}
\label{Proposition:convergence}
\end{proposition}

The detailed proof of \propref{Proposition:convergence} can be found in Appendix~2. Since variation seminorm is a norm between vectors defined up to additive constant, \propref{Proposition:convergence} shows that the gap between the approximate gradient and the true gradient decreases exponentially as more iterations are performed. This means we can obtain an approximation near true gradient with only a small number of iterations, and this is empirically verified in \figref{Figure:convergence}. Given a threshold $\delta$ of the difference between $\nabla_{\mathbf{p}_{\mathrm{S}}}^{(t)}S_{\epsilon}(\mathbf{p}_{\mathrm{T}},\mathbf{p}_{\mathrm{S}})$ and $\nabla_{\mathbf{p}_{\mathrm{S}}}S_{\epsilon}(\mathbf{p}_{\mathrm{T}},\mathbf{p}_{\mathrm{S}})$,
\begin{equation}
\begin{aligned}
 & \|\nabla_{\mathbf{p}_{\mathrm{S}}}^{(t)}S_{\epsilon}(\mathbf{p}_{\mathrm{T}},\mathbf{p}_{\mathrm{S}})-\nabla_{\mathbf{p}_{\mathrm{S}}}S_{\epsilon}(\mathbf{p}_{\mathrm{T}},\mathbf{p}_{\mathrm{S}})\|_{\mathrm{var}} \\
\leq & \|\nabla_{\mathbf{p}_{\mathrm{S}}}^{(0)}S_{\epsilon}(\mathbf{p}_{\mathrm{T}},\mathbf{p}_{\mathrm{S}})-\nabla_{\mathbf{p}_{\mathrm{S}}}S_{\epsilon}(\mathbf{p}_{\mathrm{T}},\mathbf{p}_{\mathrm{S}})\|_{\mathrm{var}}\cdot\kappa(K)^{2t} \leq \delta\;,
\end{aligned}
\end{equation}
which means
\begin{equation}
\begin{aligned}
t & \geq \frac{1}{2}\log_{\kappa(K)}\frac{\delta}{\|\nabla_{\mathbf{p}_{\mathrm{S}}}^{(0)}S_{\epsilon}(\mathbf{p}_{\mathrm{T}},\mathbf{p}_{\mathrm{S}})-\nabla_{\mathbf{p}_{\mathrm{S}}}S_{\epsilon}(\mathbf{p}_{\mathrm{T}},\mathbf{p}_{\mathrm{S}})\|_{\mathrm{var}}} \\
 & = \frac{\log\delta-\log\|\nabla_{\mathbf{p}_{\mathrm{S}}}^{(0)}S_{\epsilon}(\mathbf{p}_{\mathrm{T}},\mathbf{p}_{\mathrm{S}})-\nabla_{\mathbf{p}_{\mathrm{S}}}S_{\epsilon}(\mathbf{p}_{\mathrm{T}},\mathbf{p}_{\mathrm{S}})\|_{\mathrm{var}}}{2\log\kappa(K)} \\
 & = \mathcal{O}\left(\log\delta\right)\;.
\end{aligned}
\end{equation}
Since performing a single Sinkhorn iteration costs us $\mathcal{O}(|\mathcal{C}_{\mathrm{T}}||\mathcal{C}_{\mathrm{S}}|)$ time, the overall time complexity of optimizing the model for one step is $\mathcal{O}(|\mathcal{C}_{\mathrm{T}}||\mathcal{C}_{\mathrm{S}}|\log\delta)$. In practice, we can either control the change of the approximate gradients after two iterations or the number of maximum iterations.

\section{Teacher Assessment}
\label{Section:assessment}
Until now we have discussed how to reuse the knowledge of a teacher model that may have a different label space from the student and have studied the optimization properties of the proposed method. Standard knowledge distillation often assumes that a teacher is given in advance. However, in selective cross-task distillation, we need to pick out the most contributive teacher from the repository.

To assess a pre-trained model, recent works~\cite{NCE,LEEP,LogME,N-LEEP,OTCE,GBC} have proposed several empirical metrics to evaluate the relevance between the pre-trained model and the target task. All the existing works assume that the model reuse method is fine-tuning, and we extend this line of research to the field of knowledge distillation. In fact, fine-tuning can be seen as a special case of self-distillation~\cite{BAN,OwnTeacher}, which means we are studying a more general setting.

Intuitively, an ideal metric should be binding with the subsequent model reuse procedure, and we equip our distillation algorithm with a complementary teacher selection mechanism. Since we minimize the Sinkhorn distance to reuse a cross-task teacher, we also use Sinkhorn distance to measure the relatedness of teacher models.

\begin{table*}[t]
\centering
\caption{\small Comparison between three ways to compute the assessment metric. Approximation \textbf{I} means replacing $f_{\mathrm{S}}^{h}$ in \equref{Equation:metric} with $f_{\mathrm{S}}$. Approximation \textbf{II} means replacing $f_{\mathrm{S}}^{h}$ in \equref{Equation:metric} with $f_{\mathrm{F}}^{h}$. $\mathcal{O}(\Gamma)$ is the expected time complexity of training a deep network while $\mathcal{O}(\gamma)$ is the expected time complexity of training a linear model. $\mathcal{O}(\Xi)$ is the expected time complexity of performing the forward process of the teacher model.}
\begin{tabular}{@{}l|ccc@{}}
\toprule
Metric            & \equref{Equation:metric}                             & Approximation \textbf{I}          & Approximation \textbf{II}                            \\ \midrule
Model(s) to Train & $H$ deep networks $\{f_{\mathrm{S}}^{h}\}_{h=1}^{H}$ & $1$ deep network $f_{\mathrm{S}}$ & $H$ linear models $\{f_{\mathrm{F}}^{h}\}_{h=1}^{H}$ \\
Time Complexity   & $\mathcal{O}(H(\Xi+\Gamma))$                               & $\mathcal{O}(\Gamma)$             & $\mathcal{O}(H(\Xi+\gamma))$                               \\
Quality           & Highest                                              & Lowest                            & High                                                 \\ \bottomrule
\end{tabular}
\label{Table:metric_comparison}
\end{table*}

\begin{algorithm}[t]
\caption{\small Teacher assessment process of our proposed method~(Approximation \textbf{II}).}
\KwIn{Number of teachers $H$, Model repository $\{f_{\mathrm{T}}^{h}\}_{h=1}^{H}$, Student's traing set $\{(\mathbf{x}_i,\mathbf{y}_i)\}_{i=1}^{N}$.}
\KwOut{Selected teacher index $h^\star$.}
$h^\star\leftarrow 0$\;
$\mathcal{M}^\star \leftarrow +\infty$\;
\For{$h\in[H]$}
{
    Obtain $\{(\phi_{\mathrm{T}}^{h}(\mathbf{x}_i),\mathbf{y}_i)\}_{i=1}^{N}$\;
    Fit linear model $f_{\mathrm{F}}^{h}$ on $\{(\phi_{\mathrm{T}}^{h}(\mathbf{x}_i),\mathbf{y}_i)\}_{i=1}^{N}$\;
    $\mathcal{M}(f_{\mathrm{T}}^{h})\leftarrow\frac{1}{N}\sum_{i=1}^{N}S_{\epsilon}(\bm{\rho}_{\tau}(f_{\mathrm{T}}^{h}(\mathbf{x}_i)),\bm{\rho}_{\tau}(f_{\mathrm{F}}^{h}(\mathbf{x}_i)))$ \Comment*[r]{Implemented as \algref{Algorithm:sinkhorn}}
    \If{$\mathcal{M}(f_{\mathrm{T}}^{h})<\mathcal{M}^\star$}
    {
        $\mathcal{M}^\star\leftarrow\mathcal{M}(f_{\mathrm{T}}^{h})$\;
        $h^\star\leftarrow h$\;
    }
}
Return $h^\star$\;
\label{Algorithm:assessment}
\end{algorithm}

Assume that we have access to a model repository containing $H$ teachers, i.e.,  $\mathcal{T}=\{f_{\mathrm{T}}^h\}_{h=1}^{H}$. To be consistent with the generalized knowledge reuse procedure, we define the metric to assess a candidate teacher $f_{\mathrm{T}}^{h}$ as \equref{Equation:metric}:
\begin{equation}
\mathcal{M}(f_{\mathrm{T}}^h)=\frac{1}{N}\sum_{i=1}^{N}S_{\epsilon}(\bm{\rho}_{\tau}(f_{\mathrm{T}}^{h}(\mathbf{x}_i)),\bm{\rho}_{\tau}(f_{\mathrm{S}}^{h}(\mathbf{x}_i)))\;,
\label{Equation:metric}
\end{equation}
where $f_{\mathrm{S}}^{h}$ is the student model optimized under the supervision of instance labels and $f_{\mathrm{T}}^{h}$. \equref{Equation:metric} measures the task gap between a candidate teacher and the student. Based on the definition of Sinkhorn distance, we can conclude that if a candidate teacher is related to the target task, the value of \equref{Equation:metric} will be small.

Although \equref{Equation:metric} is an ideal metric that is consistent with our distillation loss, this metric is not practical because it is not affordable to train $f_{\mathrm{S}}^{h}$ for each $f_{\mathrm{T}}^{h}$ in $\mathcal{T}$ when $H$ is large. Thus, we consider two approximations of \equref{Equation:metric}:
\begin{itemize}
    \item \textbf{I.} Replacing $f_{\mathrm{S}}^{h}$ with $f_{\mathrm{S}}$. A naive alternative is computing the Sinkhorn distance between the $h$-th teacher and the student trained on the training set without the assistance of $f_{\mathrm{T}}$. In this case, we only need to train $f_{\mathrm{S}}$ once. However, the connection between $f_{\mathrm{T}}^{h}$ and $f_{\mathrm{S}}$ is neglected, making the metric sub-optimal;
    \item \textbf{II.} Replacing $f_{\mathrm{S}}^{h}$ with $f_{\mathrm{F}}^{h}$. To maintain the connection between teacher and student, we propose to train a ``fictitious'' student $f_{\mathrm{F}}^{h}$ for each teacher to compute the metric. In detail, we first extract the instance representations by $\phi_{\mathrm{T}}^{h}$ and then train a linear model $f_{\mathrm{F}}^{h}$ on the dataset $\{(\phi_{\mathrm{T}}^{h}(\mathbf{x}_i),\mathbf{y}_i)\}_{i=1}^{N}$. In our metric, the cost matrix $M$ is computed between the $h$-th teacher's label space and the student's label space. If $f_{\mathrm{T}}^{h}$ targets a same label space as the student, $f_{\mathrm{F}}^{h}$ trained on $\{(\phi_{\mathrm{T}}^{h}(\mathbf{x}_i),\mathbf{y}_i)\}_{i=1}^{N}$ will have similar predictions to the true student $f_{\mathrm{S}}^{h}$, and the Sinkhorn distance between their outputs will be small. Otherwise, the values in the cost matrix $M$ will be large, and the output distributions of the teacher $f_{\mathrm{T}}^{h}$ and the fictitious student $f_{\mathrm{F}}^{h}$ will be distant.
\end{itemize}

We compare the vanilla metric and two approximations in \tabref{Table:metric_comparison}. Intuitively, the metric computed exactly as \equref{Equation:metric} will have the highest quality since we train the student many times to fit each teacher model in the repository. However, its time complexity scales with the number of candidate teachers $H$, which is not acceptable for a huge repository. Approximation \textbf{I} trains only one student network, but the relationship between $f_{\mathrm{S}}$ and $f_{\mathrm{T}}^{h}$ is not explicitly considered. Approximation \textbf{II} trains $H$ linear models, achieving a better trade-off between quality and efficiency. We further empirically compare these three metrics in \secref{Subsubsection:experiment_ta_approximation}. In practice, we choose Approximation \textbf{II}, whose scheme is summarized in \algref{Algorithm:assessment}.

\textbf{Discussion on Assessment Metric.} (1) The computation of $\mathcal{M}(f_{\mathrm{T}}^h)$ using Approximation \textbf{II} is efficient and practical. We first compute the cost matrix $M$ using the teacher's embedding network and simultaneously store all the instance representations $\{\phi_{\mathrm{T}}^{h}(\mathbf{x}_i)\}_{i=1}^{N}$. This forward process exists in all the related works~\cite{NCE,LEEP,LogME,B-Tuning,N-LEEP,OTCE,GBC} and does not cost too much time since we do not store gradients during the forward process. After that, in virtue of the simplicity of linear classifiers, the computation of assessment metrics is efficient. (2) In Approximation \textbf{II}, we use a fictitious student $f_{\mathrm{F}}^{h}$ to replace the true student $f_{\mathrm{S}}^{h}$, and it is necessary to study the influence of this substitution. In fact, the gap between the fictitious and true student does not affect teacher assessment too much. Although the complexity of a linear classifier is limited, it is built upon the representations extracted by the teacher. Moreover, we do not require the fictitious student to mimic the true student perfectly but only use it to evaluate the relatedness between a teacher and the target student, i.e., we only want the teachers selected by two metrics to be identical. In \secref{Subsubsection:experiment_ta_approximation}, we check the gap between the fictitious and true student, and show that they lead to similar evaluation metrics.

\textbf{Assessment v.s. Ensemble.} Apart from scoring each teacher and picking the best one, an alternative is directly reusing all the teachers in the repository. For example, we can aggregate the features of a group of teachers and distill knowledge from the ensemble~\cite{ONE,Zoo-Tuning,FedDF}. However, this approach requires us to perform forward process of all the teachers when training a student, which makes the training cost $H$ times larger than before. Moreover, it is not trivial to aggregate a group of teachers trained on diverse tasks. Thus, teacher assessment is usually more practical than teacher ensemble in real-world applications.

\section{Experiment}
\label{Section:experiment}
\begin{figure}[t]
	\centering
	\begin{minipage}[b]{\linewidth}
		\centering
		\includegraphics[width=\linewidth]{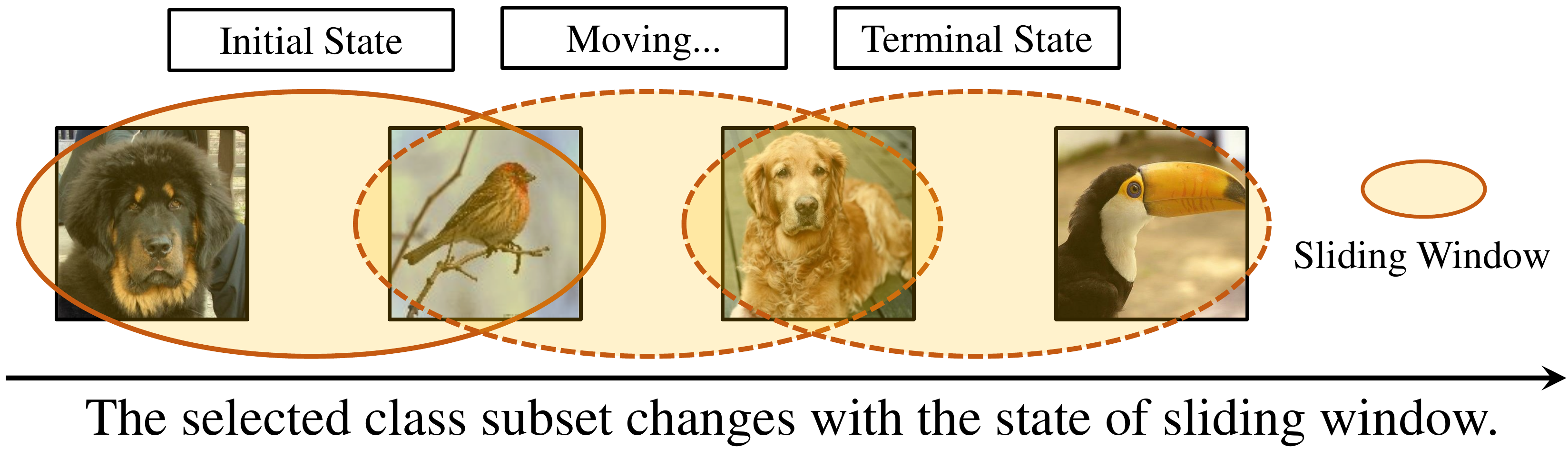}
		\subcaption{\small ``Sliding Window'' protocol. While the sliding window moves forward, the selected class subset changes.}
		\label{Figure:single_slide}
	\end{minipage}
	\begin{minipage}[b]{\linewidth}
		\centering
		\includegraphics[width=\linewidth]{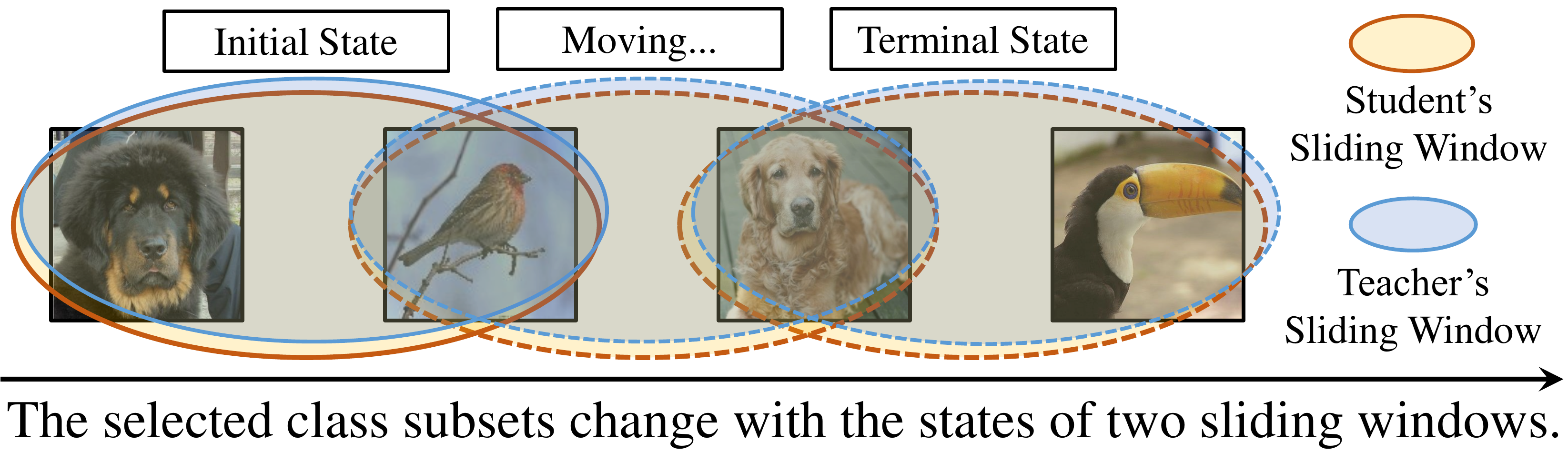}
		\subcaption{\small ``Double Sliding Window'' protocol. Two windows are used to select class subsets for teacher and student.}
		\label{Figure:double_slide}
	\end{minipage}
	\caption{\small Two split methods used for generalized knowledge reuse and teacher assessment respectively.}
	\label{Figure:slide}
\end{figure}

Since selective cross-task distillation includes two main stages, i.e., teacher assessment and generalized knowledge reuse, our experiments contain two main parts. We conduct the experiments mainly on CIFAR-100~\cite{CIFAR} and Caltech-UCSD Birds-200-2011~(CUB)~\cite{CUB}.

\begin{figure*}[!t]
	\centering
	\begin{minipage}[b]{0.24\linewidth}
		\centering
		\includegraphics[width=\linewidth]{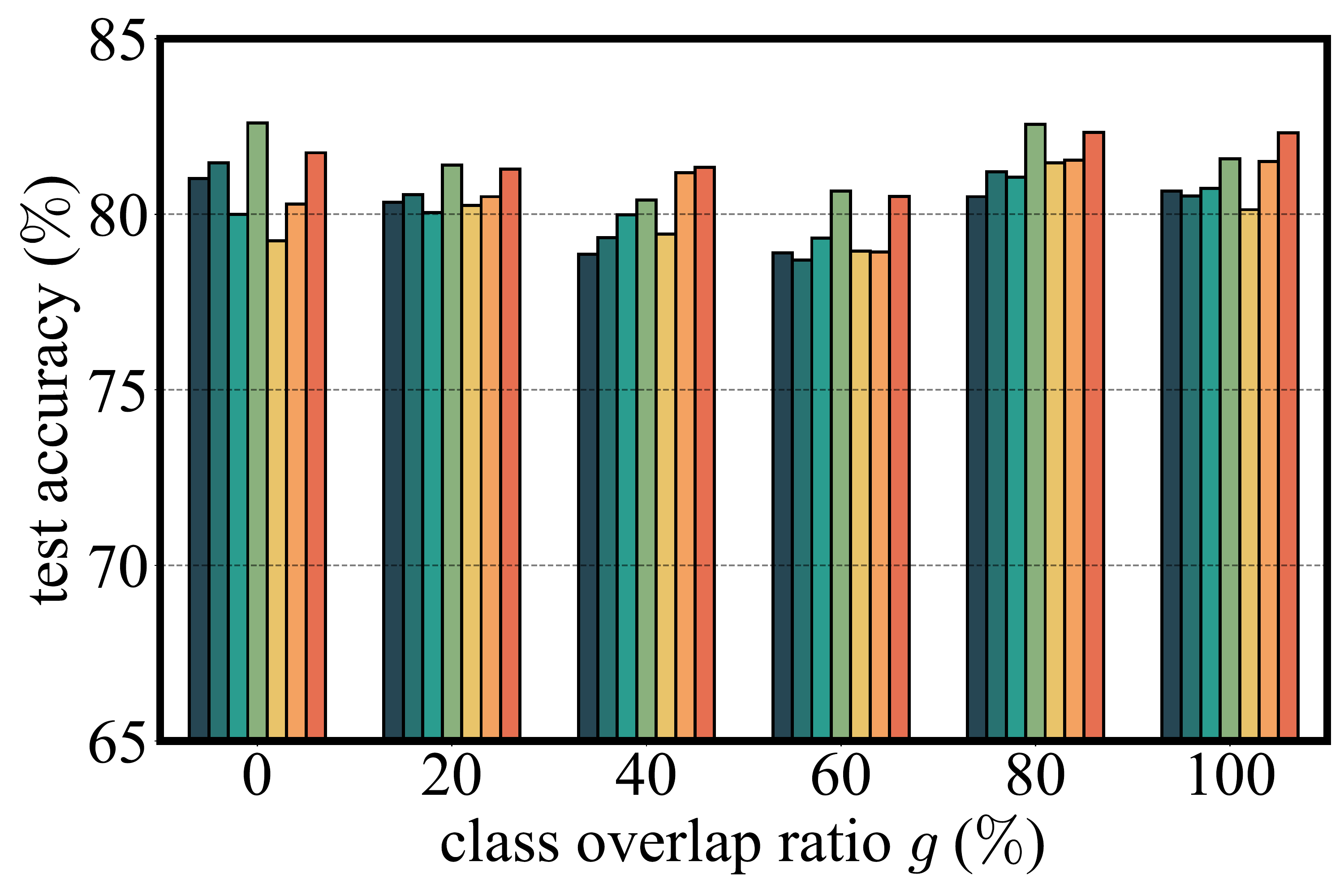}
		\subcaption{\small Student=(40,2)}
	\end{minipage}
	\begin{minipage}[b]{0.24\linewidth}
		\centering
		\includegraphics[width=\linewidth]{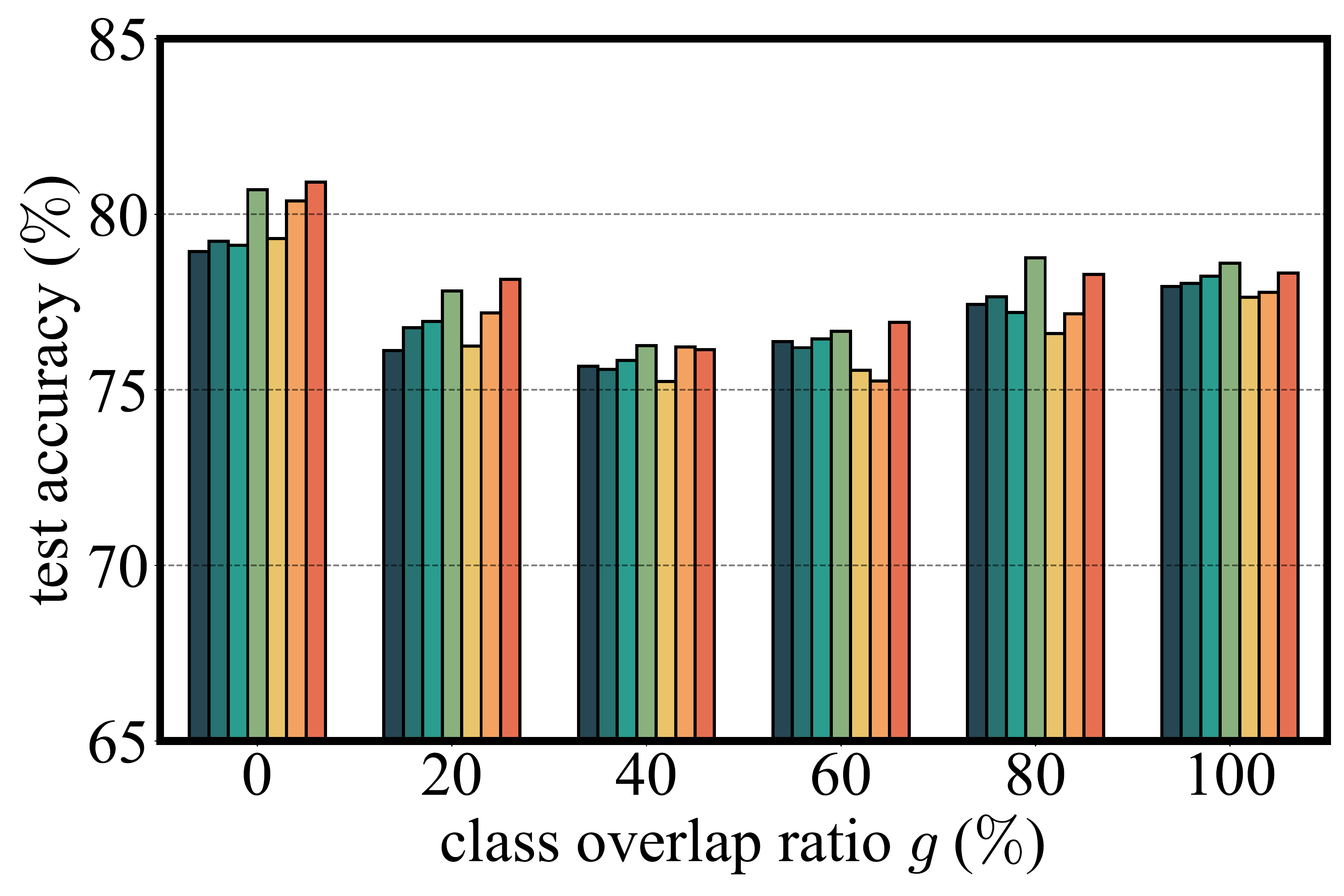}
		\subcaption{\small Student=(16,2)}
	\end{minipage}
	\begin{minipage}[b]{0.24\linewidth}
		\centering
		\includegraphics[width=\linewidth]{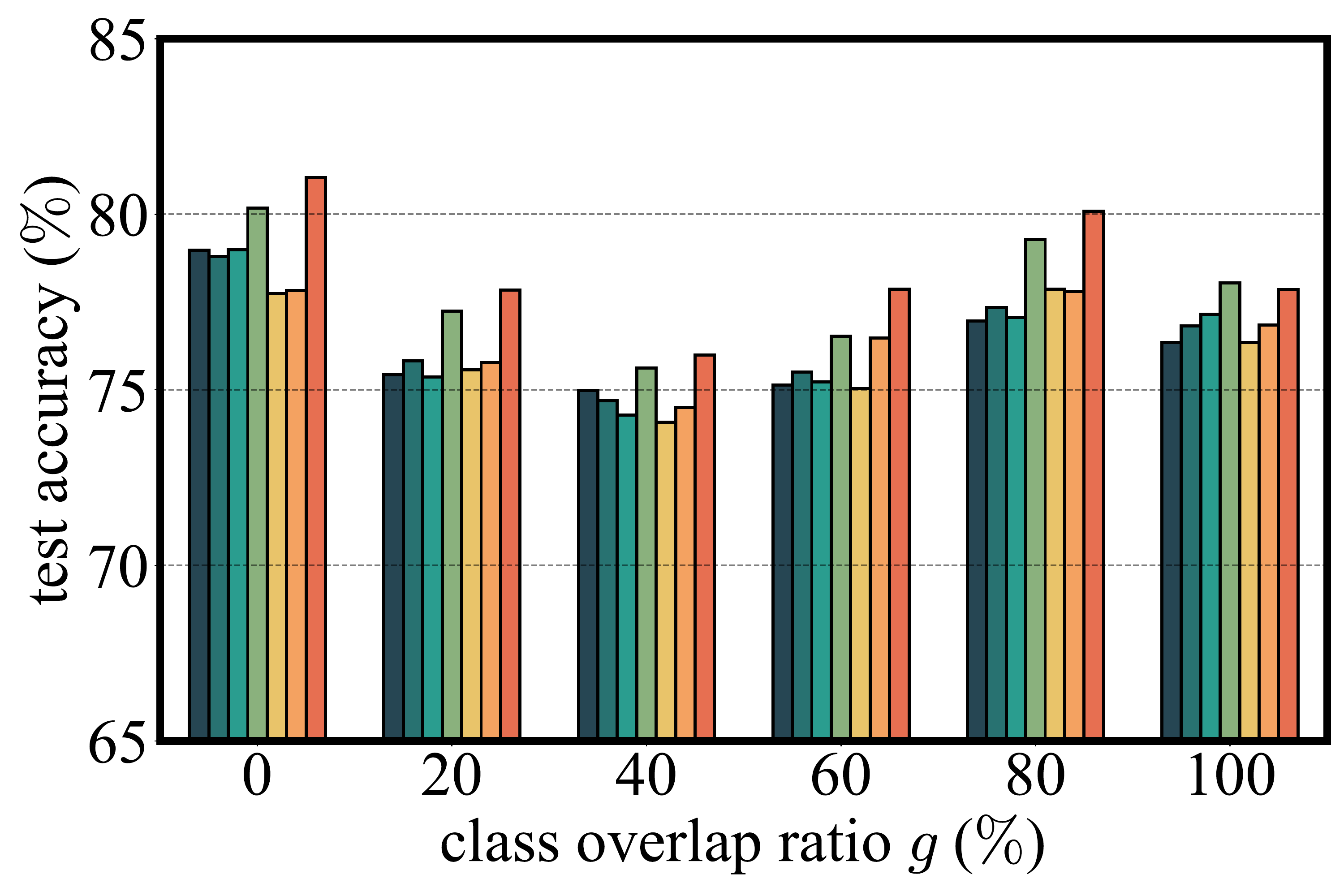}
		\subcaption{\small Student=(40,1)}
	\end{minipage}
	\begin{minipage}[b]{0.24\linewidth}
		\centering
		\includegraphics[width=\linewidth]{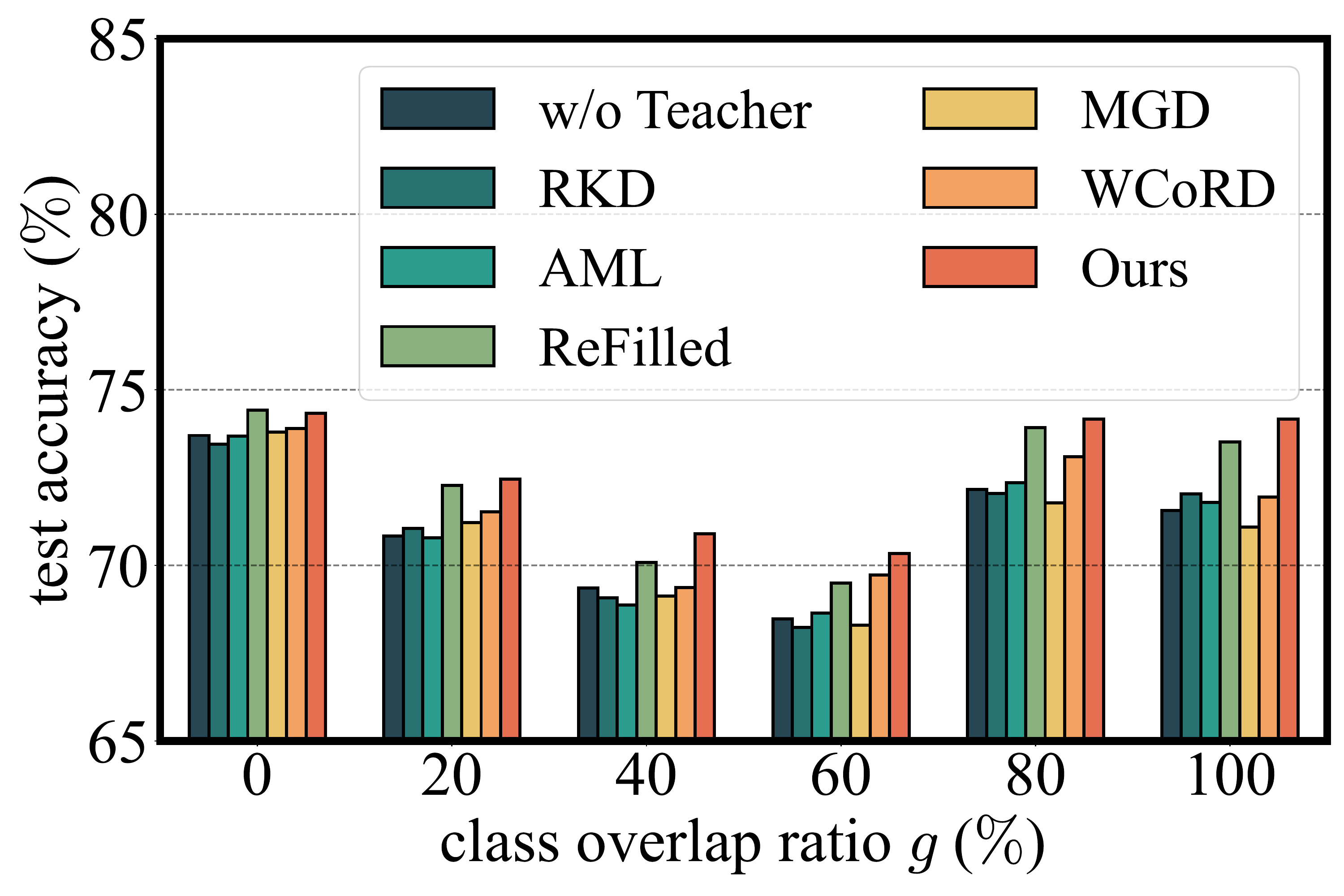}
		\subcaption{\small Student=(16,1)}
	\end{minipage}
	
	\begin{minipage}[b]{0.24\linewidth}
		\centering
		\includegraphics[width=\linewidth]{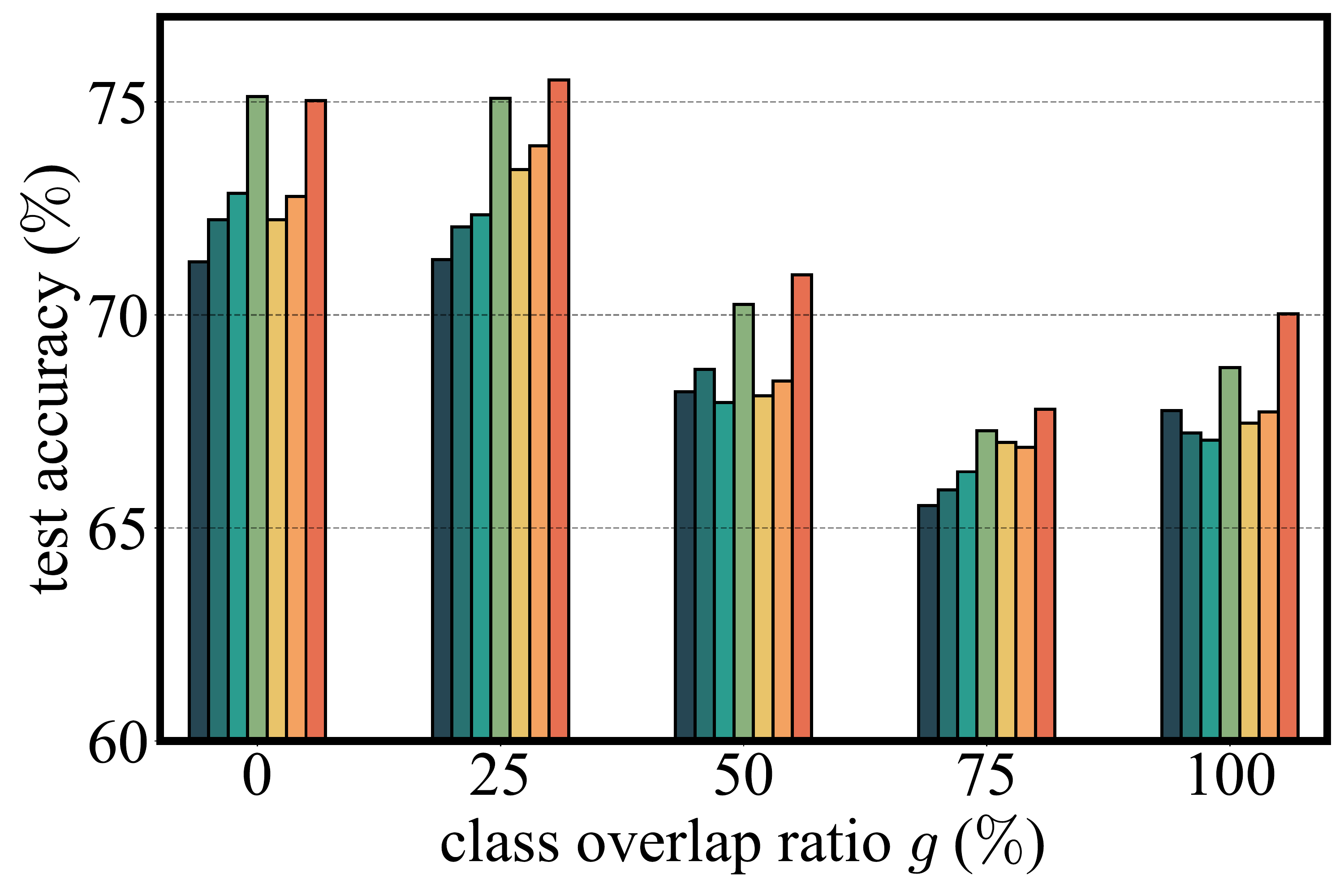}
		\subcaption{\small Student=1.0}
	\end{minipage}
	\begin{minipage}[b]{0.24\linewidth}
		\centering
		\includegraphics[width=\linewidth]{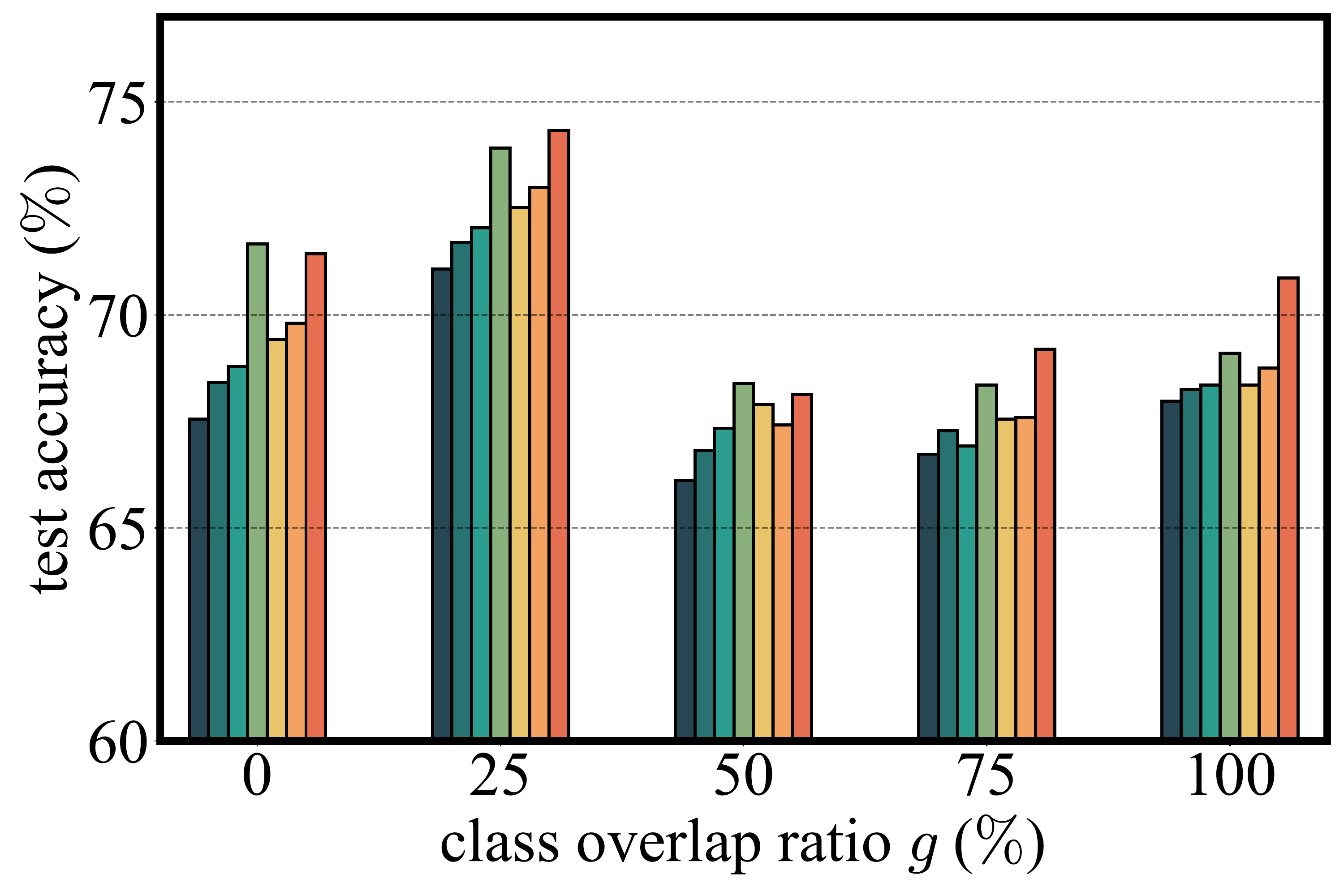}
		\subcaption{\small Student=0.75}
	\end{minipage}
	\begin{minipage}[b]{0.24\linewidth}
		\centering
		\includegraphics[width=\linewidth]{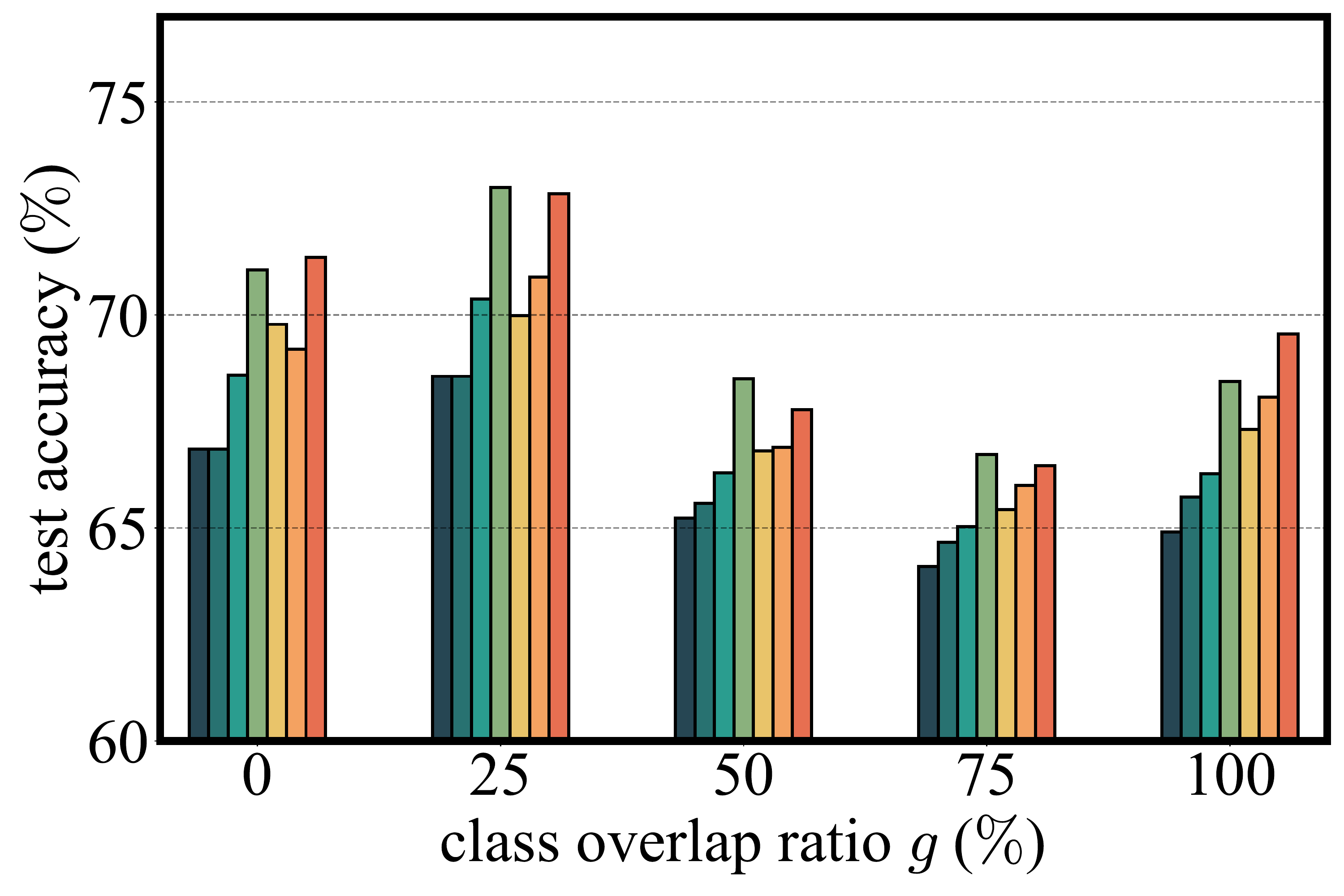}
		\subcaption{\small Student=0.5}
	\end{minipage}
	\begin{minipage}[b]{0.24\linewidth}
		\centering
		\includegraphics[width=\linewidth]{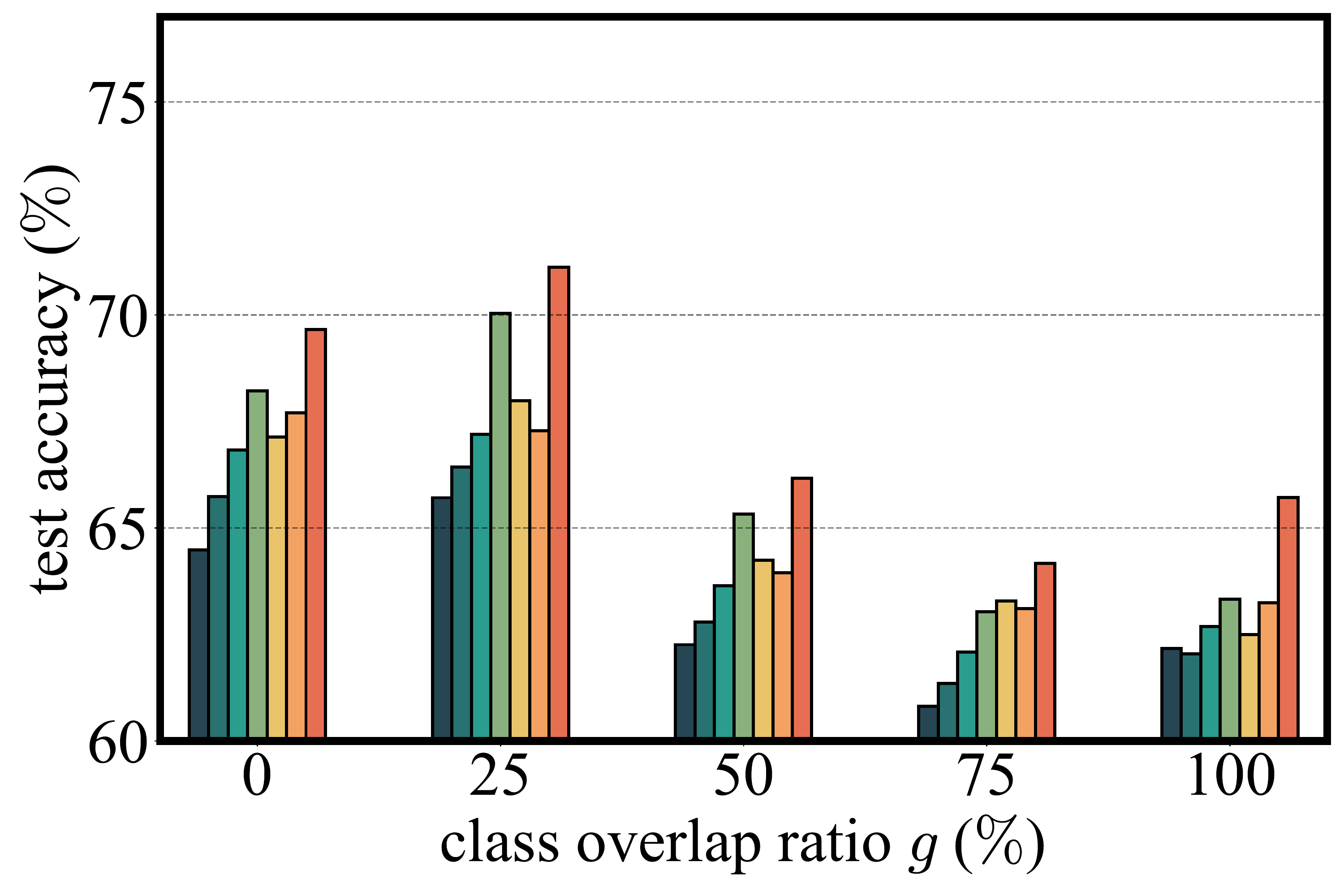}
		\subcaption{\small Student=0.25}
	\end{minipage}
	\caption{\small Results of generalized knowledge distillation on CIFAR-100 (top) and CUB (bottom). Class overlap ratio changes from $100\%$ to $0\%$. Teacher architecture is WideResNet-(40,2) for CIFAR-100 and MobileNet-1.0 for CUB. All the values for drawing this figure can be found in Appendix 3.1.}
	\label{Figure:gkd}
\end{figure*}

\subsection{Generalized Knowledge Reuse} \label{Subsection:experiment_gkr}
In the first part, we assume that {\em a fixed teacher is given} and check the generalized knowledge reuse ability of our method. In detail, we want to study following questions: 
\textbf{(1)} Can our proposed method reuse the knowledge of a generalized teacher?
\textbf{(2)} Will our method also perform well in standard knowledge distillation?
\textbf{(3)} Can cost matrix $M$ capture class relationships well?
\textbf{(4)} Ablation study on the cost matrix;
\textbf{(5)} What will happen if task gap between teacher and student is too large?
\textbf{(6)} What are the influences of hyper-parameters $\tau$, $\lambda$, and $\epsilon$ in \equref{Equation:generalized_kd}?

\subsubsection{Sliding Window Protocol}
\label{Subsubsection:experiment_gkr_sliding}
First of all, we introduce a special dataset split method called ``sliding window'' protocol, as shown in \figref{Figure:single_slide}. Taking CUB as an example, we first sort its $200$ classes randomly and then use a sliding window that covers $100$ classes to select a class subset. Initially, the window covers the first $100$ classes, moving $25$ classes forward every step until covering the last $100$ classes. This procedure generates $5$ different label sets. In this subsection, we fix the class subset of the teacher as the initial one~(first $100$ classes) and range the class subset of the student, which means the task gap between teacher and student increases when the sliding window moves forward. We can use the class overlap ratio $g$ to measure the task gap between teacher and student. For CUB, $g\in\{100\%,75\%,50\%,25\%,0\%\}$. For CIFAR-100, $g\in\{100\%,80\%,60\%,40\%,20\%,0\%\}$.

\subsubsection{Implementation Details}
\label{Subsubsection:experiment_gkr_implementation}
In this subsection, the teacher is trained on a fixed class subset~(determined by the initial sliding window). Students are trained on different class subsets. We extract the training instances belonging to its class subset to train it and test in on the corresponding test instances. For CIFAR-100, we instantiate the teacher and student as WideResNets~\cite{WideResNet}. Teacher is WideResNet-(40,2) while students are WideResNet-$\{$(40,2), (16,2), (40,1), (16,1)$\}$. For CUB, teacher is MobileNet-1.0~\cite{MobileNet} while students are MobileNet-$\{$1.0, 0.75, 0.5, 0.25$\}$.

Given a teacher or a student, we train the model for $200$ epochs. The model is optimized using SGD optimizer with initial learning rate $0.1$. For CIFAR-100, the learning rate is multiplied by $0.2$ after $50$, $100$, and $150$ epochs. For CUB, the learning rate is multiplied by $0.2$ after $150$, $170$, $180$ epochs. The batch size is $256$ for CIFAR-100 and $128$ for CUB. As for the optimizer hyper-parameters, weight decay is set to $0.0005$ and momentum is set to $0.9$. Hyper-parameter $\lambda$ in \equref{Equation:generalized_kd} is set to $10$ for CIFAR-100 and $100$ for CUB. The temperature $\tau$ of softmax function is set to $3$. Smoothing factor $\epsilon$ in Sinkhorn distance is set to $0.1$.

\subsubsection{Generalized Knowledge Distillation}
\label{Subsubsection:experiment_gkr_ctkd}
\textbf{Baseline and Compared Methods.}
A natural baseline for any knowledge distillation algorithm is training the student model without using the teacher. Since standard knowledge distillation~\cite{KD} does not work when the teacher and student target at different label spaces, we compare our method to several methods that can be applied to our setting:
\begin{itemize}
    \item Feature-based distillation methods including RKD~\cite{RKD} and AML~\cite{AML}, which can be applied to cross-task setting in form because they do not rely on labels;
    \item A method specially designed for cross-task distillation, i.e., ReFilled~\cite{ReFilled}, which utilizes the comparison ability of a teacher model to boost the student;
    \item Knowledge reuse methods based on optimal transport including MGD~\cite{MGD} and WCoRD~\cite{WCoRD}.
\end{itemize}

Results are shown in \figref{Figure:gkd}. Among all the compared methods, ReFilled~\cite{ReFilled} is the only one designed for cross-task knowledge distillation, and it achieves competitive performance. All these compared methods ignore the teacher's classifier and only utilizes its representation network, which is the main reason for their unsatisfying performances. Our proposed method simultaneously reuses $\phi_{\mathrm{T}}$ and the classifier $W_{\mathrm{T}}$, achieving best results on most cases.

A counter-intuitive phenomenon in \figref{Figure:gkd} is that the test accuracy is not positive correlated with class overlap ratio $g$ since a larger $g$ means a smaller task gap between the teacher and the student. However, the test set also changes with $g$, which makes the accuracies across different overlap ratios incomparable actually. Besides, we can see that the variation tendency of test accuracy with overlap ratio is consistent across different student architectures, and this may mean some test sets are harder than others inherently.

\begin{table}[t]
    \centering
    \caption{\small Average test accuracies on CIFAR-100. Teacher architecture is WideResNet-(40,2). When the student shares a same architecture as the teacher, we also use self-distillation~(BAN) to learn the student, and the test accuracy is $75.41\%$ on CIFAR-100. $\dag$ indicates self-supervised methods while $\pounds$ indicates methods based on optimal transport. Best results are in \textbf{bold}.}
    \begin{tabular}{@{}l|cccc@{}}
		\toprule
		(Depth, Width)           & (40, 2)     & (16, 2)    & (40, 1)    & (16, 1)    \\ \midrule
		Teacher                  & \multicolumn{4}{c}{74.44; +BAN~\cite{BAN}: 75.41}  \\
		Student                  & 74.44       & 70.15      & 68.97      & 65.44      \\ \midrule
		KD~\cite{KD}             & 75.47       & 71.87      & 70.46      & 66.54      \\
		FitNet~\cite{Fitnets}    & 74.29       & 70.89      & 68.66      & 65.38      \\
		VID~\cite{VID}           & 75.25       & 73.31      & 71.51      & 66.32      \\
		RKD~\cite{RKD}           & 76.62       & 72.56      & 72.18      & 65.22      \\
		SFTN~\cite{SFTN}         & 76.93       & 75.23      & 72.04      & 67.41      \\
		AFD~\cite{AFD}           & 77.42       & 75.58      & 72.50      & 67.37      \\
		SEED$^\dag$~\cite{SEED}  & 76.28       & 73.40      & 71.83      & 67.75      \\
		SSKD$^\dag$~\cite{SSKD}  & 75.42       & 74.03      & 72.71      & 67.30      \\
		WCoRD$^{\dag\pounds}$~\cite{WCoRD}& 77.36 & 74.29   & 72.78      & 67.35         \\      
		MGD$^\pounds$~\cite{MGD} & 76.40       & 74.25      & 72.17      & 66.80         \\
		ReFilled~\cite{ReFilled} & 77.49       & 74.01      & 72.72      & 67.56      \\ \midrule
		Ours                  & \bf 78.03   & \bf 75.83      & \bf 73.94  & \bf 68.01  \\ \bottomrule
	\end{tabular}
	\label{Table:kd_cifar100}
\end{table}

\begin{table}
	\centering
    \caption{\small Average test accuracies on CUB. Teacher architecture is MobileNet-1.0. When the student shares a same architecture as the teacher, we also use self-distillation~(BAN) to learn the student, and the test accuracy is $76.87\%$ on CUB. $\dag$ indicates self-supervised methods while $\pounds$ indicates methods based on optimal transport. Best results are in \textbf{bold}.}
	\begin{tabular}{@{}l|cccc@{}}
    	\toprule
    	Width Multiplier         & 1.0         & 0.75       & 0.5        & 0.25          \\ \midrule
    	Teacher                  & \multicolumn{4}{c}{75.36; +BAN~\cite{BAN}: 76.87}     \\
    	Student                  & 75.36       & 74.87      & 72.41      & 69.72         \\ \midrule
    	KD~\cite{KD}             & 77.61       & 76.02      & 74.24      & 72.03         \\
    	FitNet~\cite{Fitnets}    & 75.10       & 75.03      & 72.17      & 69.09         \\
    	VID~\cite{VID}           & 77.03       & 76.91      & 75.62      & 72.23         \\
    	RKD~\cite{RKD}           & 77.72       & 76.80      & 74.99      & 72.55         \\
    	SFTN~\cite{SFTN}         & 77.64       & 77.90      & 77.34      & 73.55         \\
    	AFD~\cite{AFD}           & 78.67       & 78.11      & 77.42      & 73.60         \\		
    	SEED$^\dag$~\cite{SEED}  & 77.93       & 78.14      & 77.50      & 73.23         \\
    	SSKD$^\dag$~\cite{SSKD}  & 78.34       & 78.22      & 77.10      & 72.18         \\
    	WCoRD$^{\dag\pounds}$~\cite{WCoRD}& 79.02 & 78.20   & 77.83      & 74.22         \\      
    	MGD$^\pounds$~\cite{MGD} & 78.55       & 77.69      & 76.68      & 73.40         \\
    	ReFilled~\cite{ReFilled} & 79.33       & 78.52  & 76.90      & 74.04         \\ \midrule
    	Ours                  & \bf 79.87   & \bf 78.92      & \bf 78.43  & \bf 75.01     \\ \bottomrule
    \end{tabular}
	\label{Table:kd_cub200}
\end{table}

\begin{table}[t]
\centering
\caption{\small Average test accuracies on CIFAR-100. Teacher architecture is ResNet-50. Its accuracy on test set is $72.17$. Teacher and student share a same label space. $\dag$ indicates self-supervised methods while $\pounds$ indicates methods based on optimal transport. Best results are in \textbf{bold}.}
\begin{tabular}{@{}l|cccc@{}}
\toprule
(Depth, Width)           & (40, 2)     & (16, 2)    & (40, 1)    & (16, 1)    \\ \midrule
Teacher                  & \multicolumn{4}{c}{72.17} \\
Student                  & 74.44       & 70.15      & 68.97      & 65.44      \\ \midrule
KD~\cite{KD}             &   74.15     &  71.23     &  69.30     &  66.12     \\
FitNet~\cite{Fitnets}     & 74.00       & 70.22      & 68.06      & 64.72      \\
VID~\cite{VID}           & 74.25       & 72.83      & 71.00      & 65.76      \\
RKD~\cite{RKD}           & 74.82       & 71.99      & 71.35      & 65.20      \\
SFTN~\cite{SFTN}           & 74.22       &  72.13    & 70.20      & 64.37      \\
AFD~\cite{AFD}           & 74.58       & 71.99      & 71.05      & 64.74      \\
SEED$^\dag$~\cite{SEED}         & \bf 75.03       & 72.25      & 70.93      & 66.44      \\
SSKD$^\dag$~\cite{SSKD}         & 74.86       & 72.75      & 71.96      & 66.87      \\
WCoRD$^{\dag\pounds}$~\cite{WCoRD}           & 74.98       & 72.54      & 71.60      & 66.43      \\
MGD$^{\pounds}$~\cite{WCoRD}           & 74.50       & 71.87      &  71.22     & 66.15      \\
ReFilled~\cite{ReFilled} & 74.92       & 72.57      & 71.85      & 66.90      \\ \midrule
Ours                  & 74.96       & \bf 74.23      & \bf 72.66      & \bf 67.89      \\ \bottomrule
\end{tabular}
\label{Table:kd_family_cifar100}
\end{table}

\begin{table}[t]
\centering
\caption{\small Average test accuracies on CUB. Teacher architecture is ResNet-50. Its accuracy on test set is $74.29$. Teacher and student share a same label space. $\dag$ indicates self-supervised methods while $\pounds$ indicates methods based on optimal transport. Best results are in \textbf{bold}.}
\begin{tabular}{@{}l|cccc@{}}
\toprule
Width Multiplier           & 1.0     & 0.75    & 0.5    & 0.25    \\ \midrule
Teacher                  & \multicolumn{4}{c}{74.29} \\
Student                  & 75.36       & 74.87      & 72.41      & 69.72      \\ \midrule
KD~\cite{KD}             & 75.66       & 74.50      & 73.14   & 70.88      \\
FitNet~\cite{Fitnets}     & 75.20       & 74.28      & 72.66      & 70.52      \\
VID~\cite{VID}           & 75.82       & 74.14     & 72.50      & 70.92     \\
RKD~\cite{RKD}           & 75.29       & 74.34      & 72.87     & 71.02     \\
SFTN~\cite{SFTN}           &  75.77      & 74.69      & 72.90      & 71.45      \\
AFD~\cite{AFD}           & 75.43       & 74.20       & 72.62      & 71.08      \\
SEED$^\dag$~\cite{SEED}         & 76.23       & 75.12      & 73.49     & 71.88      \\
SSKD$^\dag$~\cite{SSKD}         & 76.34       & 75.22      & 73.80     & 72.35      \\
WCoRD$^{\dag\pounds}$~\cite{WCoRD}           & 76.34       & 74.88      &  74.17     &  73.53     \\
MGD$^{\pounds}$~\cite{WCoRD}           & 75.73       & 74.26      & 74.50      &  73.64     \\
ReFilled~\cite{ReFilled} & 76.24       & 75.50      & 74.97     & 73.87      \\ \midrule
Ours                  & \bf 76.73       & \bf 76.04     & \bf 75.67      & \bf 74.35      \\ \bottomrule
\end{tabular}
\label{Table:kd_family_cub200}
\end{table}

\subsubsection{Standard Knowledge Distillation}
\label{Subsubsection:experiment_gkr_kd}
Standard knowledge distillation is a special case of generalized knowledge distillation. In this part, we show that our proposed method can achieve competitive performance on standard knowledge distillation. We compare our method to several distillation methods on CIFAR-100 and CUB. Architectures of the teachers and the students are same as those in the previous part. Results are listed in \tabref{Table:kd_cifar100} and \tabref{Table:kd_cub200}. Our method achieves best results in most cases. Specifically, when teacher and student share a same architecture, we also try to use self-distillation~\cite{BAN} to learn a student. SEED~\cite{SEED} and SSKD~\cite{SSKD} are recently proposed self-supervised distillation methods, and we can see that our method is better because instance labels are used.

Another interesting topic is cross-family distillation, which means the architectures of teacher and student come from different families. Specifically, we set teacher to ResNet-50~\cite{ResNet} for both CIFAR-100 and CUB. Results are shown in \tabref{Table:kd_family_cifar100} and \tabref{Table:kd_family_cub200}. In \tabref{Table:kd_family_cifar100}, we can see that the accuracy of teacher (72.17) is lower than that of a WideResNet-(40,2) student (74.44). Thus, the improvements of all the distillation methods in the first column are limited. Our proposed method achieves best accuracy in most cases, showing that it can distill the knowledge of a cross-family teacher. Similar phenomenon can be observed in \tabref{Table:kd_family_cub200}.

\begin{table*}[t]
\caption{\small Ablation study about the cost matrix. Several kinds of cost matrices are considered. We report the student's test accuracies on both CIFAR-100 and CUB. Three kinds of cost matrices are considered. For CIFAR-100, class overlap ratio $g$ is set to $60\%$. For CUB, class overlap ratio $g$ is set to $50\%$. Best results are in \textbf{bold}.}
\centering
\begin{tabular}{@{}lcccc|lcccc@{}}
\toprule
\multicolumn{5}{c|}{CIFAR-100 ($g=60\%$)}              & \multicolumn{5}{c}{CUB ($g=50\%$)}               \\
(Depth, Width) & (40, 2) & (16, 2) & (40, 1) & (16, 1) & Width Multiplier & 1.0   & 0.75  & 0.5   & 0.25  \\ \midrule
\textbf{I.} Constant       & 76.42   & 71.50   & 73.33   & 66.69   & \textbf{I.} Constant         & 67.89 & 61.60 & 62.33 & 61.42 \\
\textbf{II.} Random         & 75.17   & 72.38   & 73.69   & 67.95   & \textbf{II.} Random            & 65.30 & 62.35 & 64.47 & 62.54 \\
\textbf{III.} Superclass        & 77.14   & 73.82   & 74.90   & 68.37    & \textbf{III.} Superclass           & - & - & - & - \\
\textbf{IV.} Shallow         & 78.15   & 75.33   & 75.29   & 69.04   & \textbf{IV.} Shallow           & 68.47 & 66.54 & 66.20 & 65.35 \\
Ours        & \bf 80.51   & \bf 76.92   & \bf 77.87   & \bf 70.34   & Ours          & \bf 70.94 & \bf 68.13 & \bf 67.78 & \bf 66.17 \\ \bottomrule
\end{tabular}
\label{Table:gkd_ablation}
\end{table*}

\begin{figure*}[t]
	\centering
	\begin{minipage}[b]{0.32\linewidth}
		\centering
		\includegraphics[width=\linewidth]{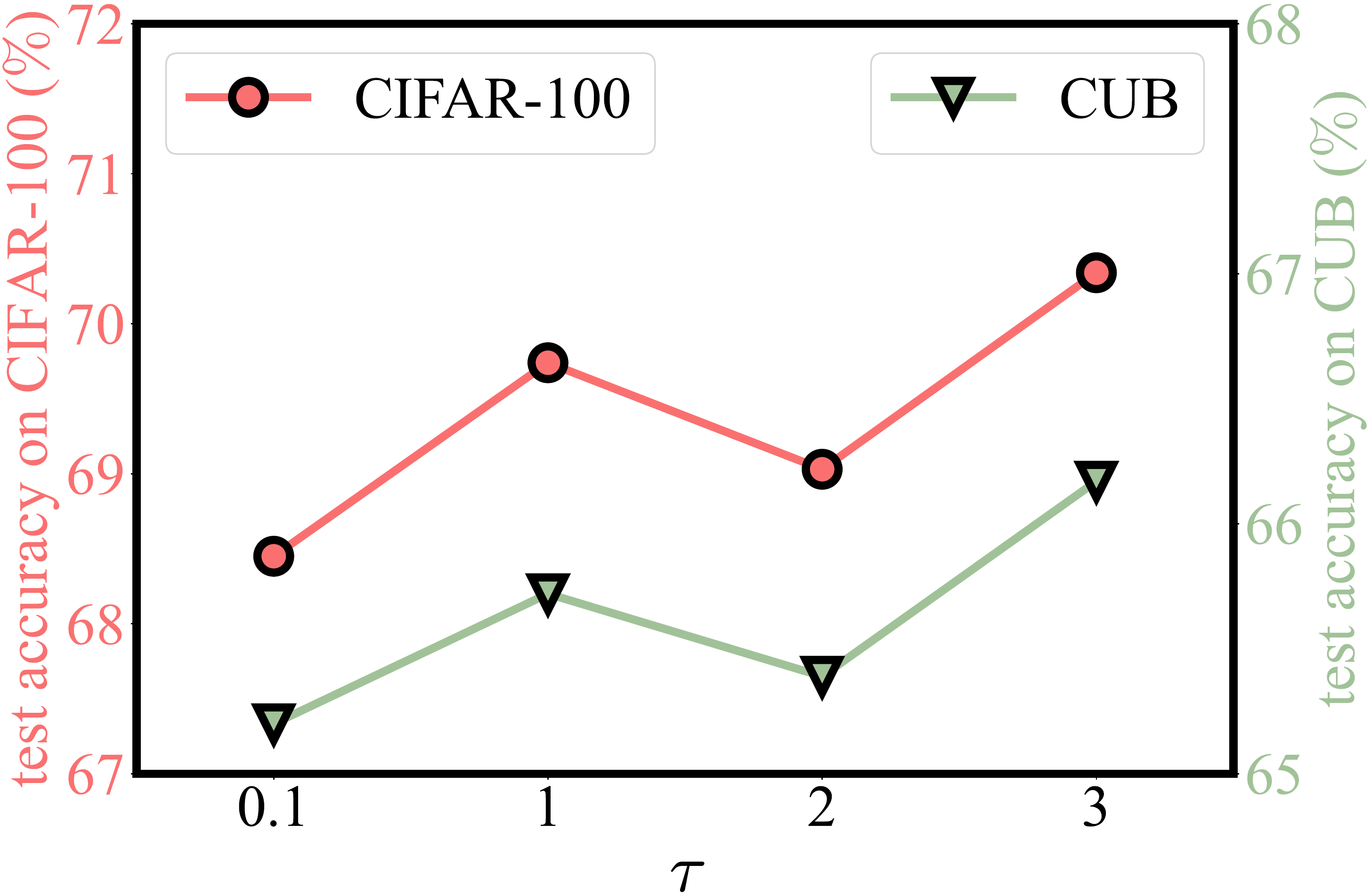}
		\subcaption{\small Influence of hyper-parameter $\tau$.}
		\label{Figure:gkd_hyper_tau}
	\end{minipage}
	\begin{minipage}[b]{0.32\linewidth}
		\centering
		\includegraphics[width=\linewidth]{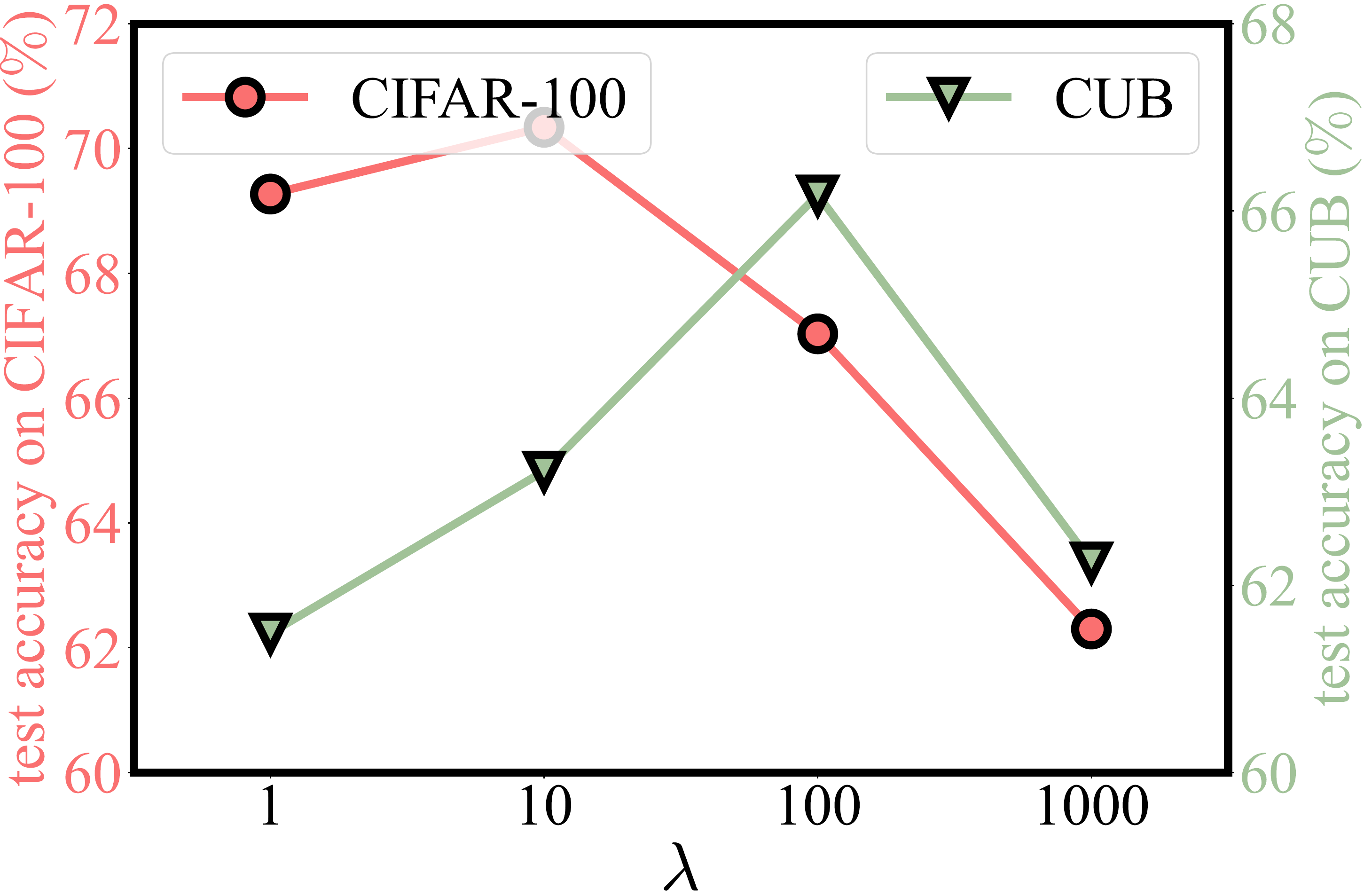}
		\subcaption{\small Influence of hyper-parameter $\lambda$.}
		\label{Figure:gkd_hyper_lambda}
	\end{minipage}
	\begin{minipage}[b]{0.32\linewidth}
		\centering
		\includegraphics[width=\linewidth]{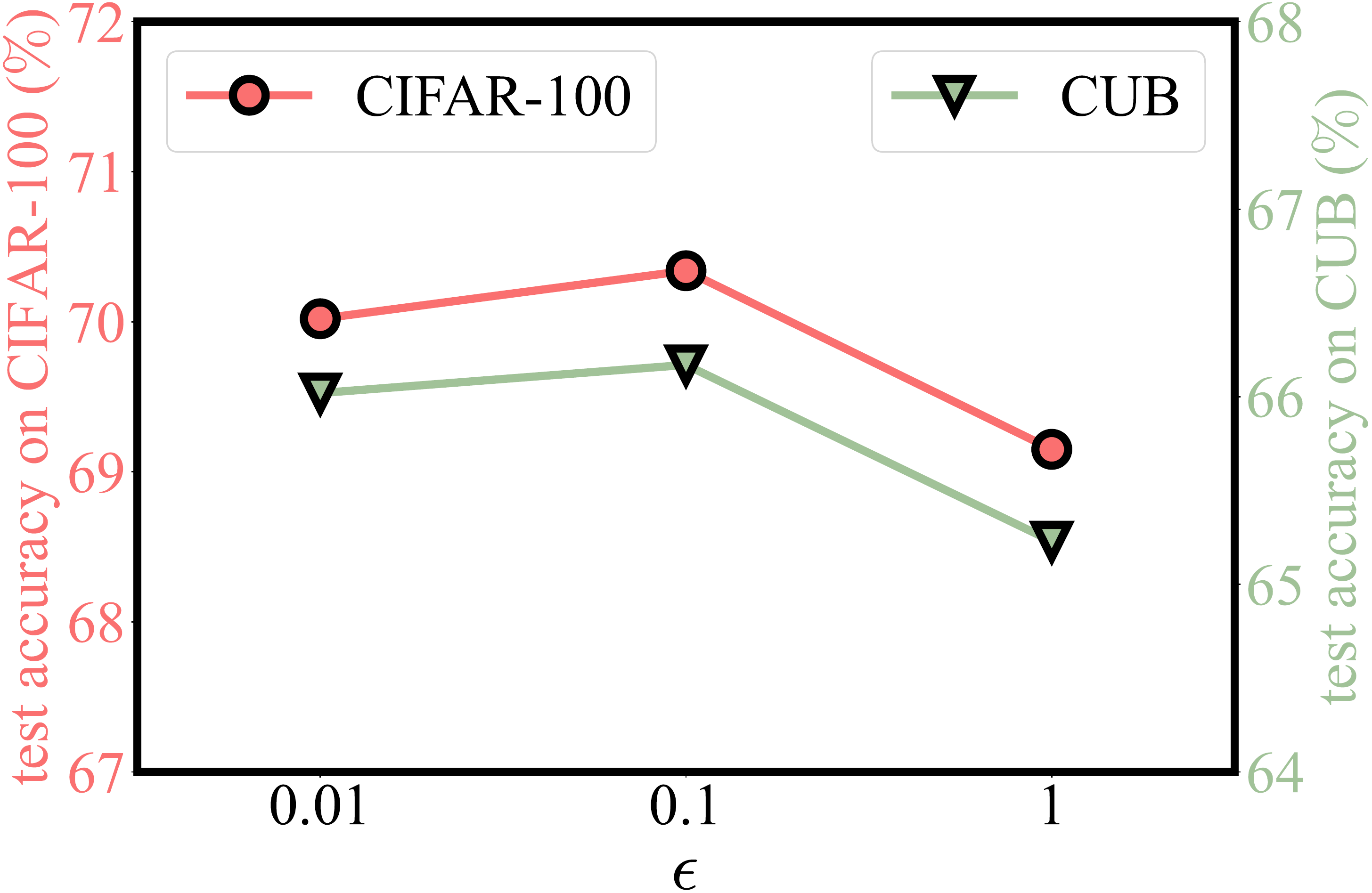}
		\subcaption{\small Influence of hyper-parameter $\epsilon$.}
		\label{Figure:gkd_hyper_epsilon}
	\end{minipage}
	\caption{\small Influence of three hyper-parameters. Without specification, $\tau$ is usually set to $3$ for both datasets. $\lambda$ is usually set to $10$ for CIFAR-100 and $100$ for CUB. $\epsilon$ is usually set to $0.1$ for both datasets.}
	\label{Figure:gkd_hyper}
\end{figure*}

\begin{table}[t]
    \centering
    \caption{\small NMI based on a randomly initialized embedding network and the teacher's embedding network $\phi_{\mathrm{T}}$.}
    \begin{tabular}{@{}l|cccccc@{}}
	\toprule
	$g (\%)$            & 100  & 80   & 60   & 40   & 20   & 0    \\ \midrule
	random              & 0.52 & 0.52 & 0.59 & 0.56 & 0.55 & 0.55 \\
	$\phi_{\mathrm{T}}$ & 0.80 & 0.72 & 0.77 & 0.76 & 0.79 & 0.81 \\ \bottomrule
	\end{tabular}
	\label{Table:clustering}
\end{table}

\subsubsection{Effect of Cost Matrix}
\label{Subsubsection:experiment_gkr_cost_effect}
An important module in our proposed method is the cost matrix $M$, which is constructed by the teacher's embedding network $\phi_{\mathrm{T}}$. In this part, we check whether this embedding network can characterize class semantics well. Specifically, we conduct a clustering experiment on CIFAR-100~\cite{CIFAR}. CIFAR-100 contains $20$ superclasses and $5$ classes in each superclass. Classes belonging to same superclass share similar semantic information. For a given class overlap ratio $g$, we compute the embedding centers of the student's classes by $\phi_{\mathrm{T}}$ and a randomly initialized network and then perform K-means clustering on them. The superclass of each class center is considered as the ground truth, and we report normalized mutual information~(NMI) in \tabref{Table:clustering}. We can see that the teacher's embedding network successfully capture semantic information of unseen classes.

\subsubsection{Ablation Study on Cost Matrix}
\label{Subsubsection:experiment_gkr_cost_ablation}
In this part, we conduct some important ablation studies about the cost matrix under the setting of generalized knowledge distillation. In our proposed method, we construct the cost matrix $M$ by the teacher's embedding network $\phi_{\mathrm{T}}$. Now we try some other cost matrices and report model performance on both CIFAR-100 and CUB. Specifically, these kinds of cost matrices are considered:
\begin{itemize}
\item \textbf{I.} Constant cost matrix. $M_{mn}=1$ if $m\neq n$ and $M_{mn}=0$ if $m=n$. In this case, the semantic relationships between source classes and target classes are totally ignored, which is a naive baseline;
\item \textbf{II.} Random cost matrix. We first compute the class centers by a randomly initialized embedding network which has a same structure as $\phi_{\mathrm{T}}$, and then compute the Euclidean distances between class centers to determine $M$. The computing process is same as \equref{Equation:cost_matrix};
\item \textbf{III.} Superclass-based cost matrix. CIFAR-100 dataset contains $20$ superclasses, and we set $M_{ij}=0$ if class $i$ and class $j$ come from a same superclass. Otherwise, we set $M_{ij}=1$. This setting is not applicable to CUB;
\item \textbf{IV.} Shallow cost matrix. We train a shallower teacher model, i.e., WideResNet-(16,2) for CIFAR-100 and MobileNet-0.5 for CUB, and compute the cost matrix.
\end{itemize}

Experiment results are shown in \tabref{Table:gkd_ablation}. This experiment is performed using class overlap ratio $g=60\%$ for CIFAR-100 and $g=50\%$ for CUB. We can see that our proposed cost matrix brings us best performances. This is because $M$ constructed by the teacher's embedding network can capture semantic information of both source classes and target classes well. Constant cost matrix and random cost matrix are two naive baselines, and they cannot describe the semantic relationships between source classes and target classes. Superclasses in CIFAR-100 bring similar classes together and offer instructive supervision. A shallower teacher can also capture class relatedness to some extent, but its comparison ability is weaker than the deep teacher.

\begin{table}[t]
    \centering
    \caption{\small Average test accuracy on DOG. The teacher is trained on CUB. Best results are in \textbf{bold}.}
    \begin{tabular}{@{}l|cccc@{}}
	\toprule
	Channel Width                & 1.0   & 0.75  & 0.5   & 0.25  \\ \midrule
	Student                      & 72.35 & 70.69 & 70.11 & 68.57 \\
	FitNet~\cite{Fitnets}        & 71.14 & 68.37 & 69.90 & 68.41 \\
	RKD~\cite{RKD}               & 72.38 & 70.15 & 69.90 & 68.42 \\ 
	ReFilled~\cite{ReFilled}     & 73.07 & 70.23 & 69.35 & 68.18 \\ 
	WCoRD~\cite{WCoRD}           & 72.84 & 69.98 & 69.42 & 68.20 \\ 
	MGD~\cite{MGD}           & 73.25 & 70.14 & 69.77 & 68.69 \\
	\midrule
	Ours                      & \bf 73.55 & \bf 70.74 & \bf 70.35 & \bf 68.60 \\ \bottomrule
	\end{tabular}
    \label{Table:gap}
\end{table}

\subsubsection{Extremely Large Task Gap}
\label{Subsubsection:experiment_gkr_gap}
Since we assume that a fixed teacher is given, an interesting question is: what will happen if this teacher is irrelevant to the current task? Ideally, the student's performance will not drop compared to training without the teacher. To verify this, we additionally train student models on Stanford Dog~\cite{DOG} with assistance of a teacher trained on the whole CUB dataset. Results are listed in \tabref{Table:gap}. Some methods that fit the cross-task setting are compared. We can see that our method does not suffer from performance drop while most of the compared methods do harm to student's performance to some extent. This means our method is robust to the task gap between teacher and student.

\begin{figure*}[t]
	\centering
	\includegraphics[width=\textwidth]{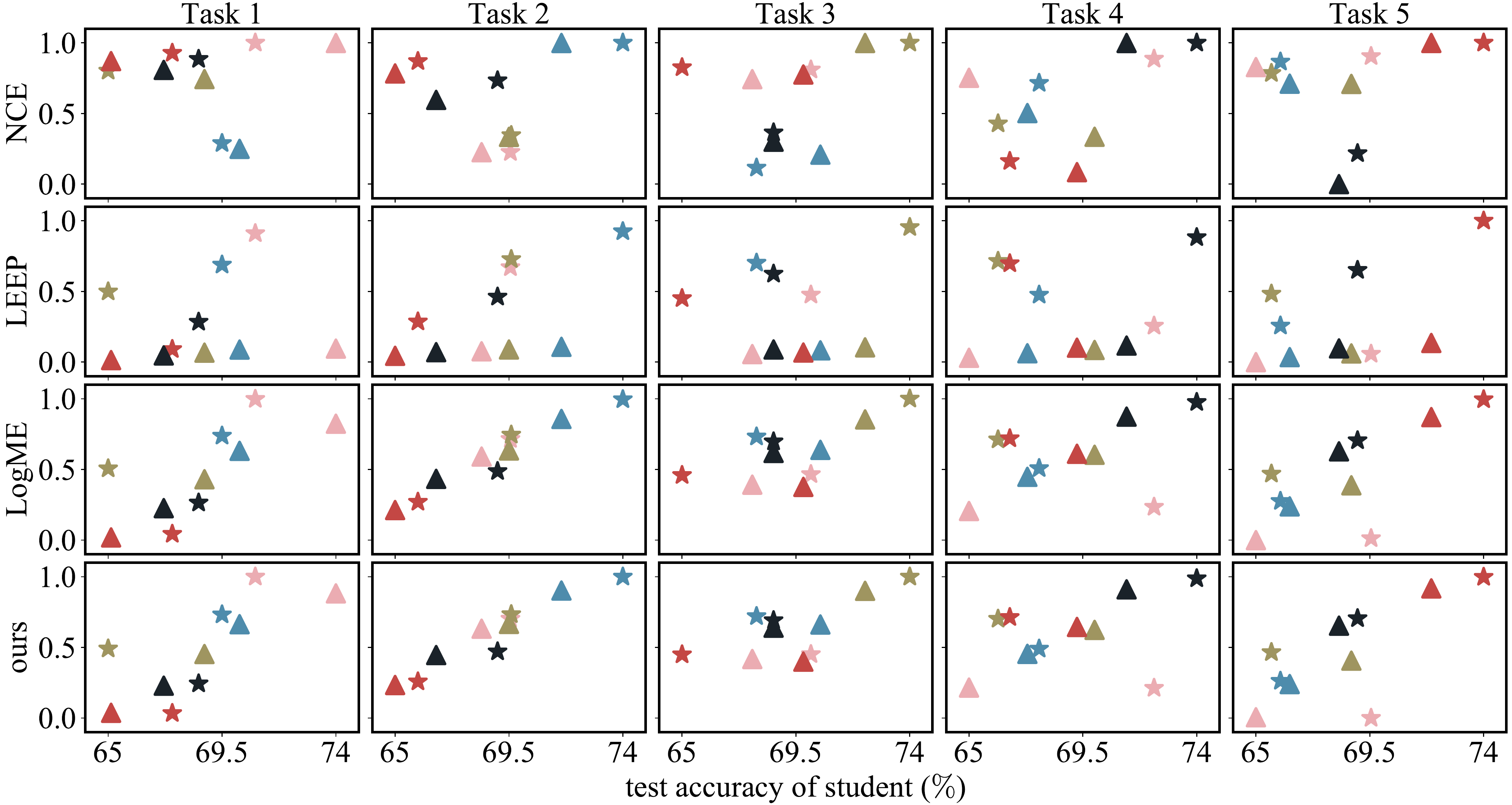}
	\caption{\small Correlation between four metrics and the student accuracy. $\bigstar$ means MobileNet, and $\blacktriangle$ means ResNet. We have normalized each metric into $[0,1]$. Different colors denote the teacher models trained on different class subsets. We list all the values for drawing this figure in Appendix 3.2.}
	\label{Figure:ta_cub200}
\end{figure*}

\subsubsection{Hyper-Parameter Analysis}
\label{Subsubsection:experiment_gkr_hyper}
In this part, we study the influence of several important hyper-parameters in our proposed method. To be specific, there are three hyper-parameters under our consideration, i.e., temperature $\tau$, weight of distillation term $\lambda$, and regularization strength $\epsilon$ in Sinkhorn distance. All these hyper-parameters appear in our objective \equref{Equation:generalized_kd}. Now, we range $\tau\in\{0.1, 1, 2, 3\}$, $\lambda\in\{1, 10, 100, 1000\}$, $\epsilon\in\{0.01, 0.1, 1\}$, and show their influences on model performance. This experiment is conducted on CIFAR-100 (class overlap ratio $g=60\%$) and CUB (class overlap ratio $g=50\%$). The student architecture is fixed as WideResNet-(16,1) for CIFAR-100 and MobileNet-0.25 for CUB. Results are shown in \figref{Figure:gkd_hyper_tau}, \figref{Figure:gkd_hyper_lambda}, and \figref{Figure:gkd_hyper_epsilon}. We can see that weight of distillation term has a remarkable influence on model performance, and we set it to different values for different datasets. $\epsilon$ controls the strength of entropy regularization term in Sinkhorn distance, and a large $\epsilon$ tends to decrease the effect of semantic transport. $\tau$ smooths the output probability distributions of teacher and student, and a proper $\tau$ will improve model performance to some extent.

\subsection{Teacher Assessment} \label{Subsection:ta}
In the second part, we construct a group of teachers to check the ability of our method to assess teachers. Several questions are under consideration: \textbf{(1)} Can our proposed metric successfully rank all the teachers according to their contributions? \textbf{(2)} What is the influence of the approximation in computing our metric? \textbf{(3)} Is the metric efficient enough to be applied in practical applications? \textbf{(4)} Can the metric work well in both coarser-grained and fine-grained teacher selection?

\subsubsection{Double Sliding Window Protocol}
\label{Subsubsection:experiment_ta_sliding}
In order to generate multiple teachers, we expand the aforementioned sliding window protocol to ``double sliding window protocol'', which contains two sliding windows for the teacher and the student, respectively, as shown in Figure~\ref{Figure:double_slide}. Now both the teacher's and the student's subsets can change with their own sliding window. The window sizes for CIFAR-100 and CUB are $50$ classes and $100$ classes respectively. The step sizes of both sliding windows are $10$ classes and $25$ classes for CIFAR-100 and CUB respectively.

\subsubsection{Implementation Details}
\label{Subsubsection:experiment_ta_implementation}
In this subsection, owing to the time consumption of constructing the model repository, we only conduct experiments on CUB dataset. The architecture of student is fixed to MobileNet-0.25. We consider another architecture family~(ResNet~\cite{ResNet}) to enrich the repository of teacher models. In total, we have $5$ class subsets and $2$ architectures, which means we can construct $10$ different teachers on CUB. Training details are same as those described in \secref{Subsubsection:experiment_gkr_implementation}.

\begin{table}[t]
\centering
\caption{\small Pearson correlation coefficients between metrics and ground-truth accuracy on $5$ tasks. Best results are in \textbf{bold}.}
\begin{tabular}{@{}l|cccccc@{}}
\toprule
Metrics            & Task 1 & Task 2 & Task 3 & Task 4 & Task 5 & Average \\ \midrule
NCE~\cite{NCE}     & -0.03  & 0.16   & 0.35   & 0.50   & 0.19   & 0.23    \\
LEEP~\cite{LEEP}   & 0.14   & 0.60   & 0.17   & 0.14   & 0.50   & 0.31    \\
LogME~\cite{LogME} & 0.72   & \bf 0.96   & 0.66   & 0.40   & 0.73   & 0.69    \\ \midrule
Ours               & \bf 0.77   & \bf 0.96   & \bf 0.69   & \bf 0.43   & \bf 0.75   & \bf 0.72    \\ \bottomrule
\end{tabular}
\label{Table:pearson}
\end{table}

\subsubsection{Teacher Assessment}
\label{Subsubsection:experiment_ta_assessment}
In this part, we study whether our proposed metric can precisely select the most contributive teacher compared to other metrics including NCE~\cite{NCE}, LEEP~\cite{LEEP}, and LogME~\cite{LogME}. Assuming that we have $H$ teachers, given a target task (a class subset for student), we denote by $G^{h}$ the test accuracy of the student trained with assistance of the $h$-th teacher. We use $Q^{h}$ to represent some evaluation metric of the $h$-th teacher. Ideally, $Q$ and $G$ should be highly correlated.

In our assessment method, we set $Q^{h}=-\mathcal{M}(f_{\mathrm{T}}^{h})$ where $\mathcal{M}$ is defined by \equref{Equation:metric}. To make the computation of \equref{Equation:metric} efficient, we also introduce several approximations in \secref{Section:assessment}, and we use Approximation \textbf{II} by default. On CUB dataset, we have $H=10$, and we show the values of $G$ and $Q$ in \figref{Figure:ta_cub200}. Note that we have normalized each metric into $[0, 1]$. We can see that LogME and our proposed metric are positively associated with the student accuracy while other two metrics fail. In this figure, we can see that LEEP is architecture-sensitive since all triangles~(ResNets) are assigned low confidence. In addition, Pearson correlation coefficients between each metric and the ground-truth accuracy on $5$ tasks are listed in \tabref{Table:pearson}. Our metric achieves the best results on $5$ tasks, showing that the proposed method can rank the teachers according to contributions.

\begin{figure}[t]
	\centering
	\includegraphics[width=\columnwidth]{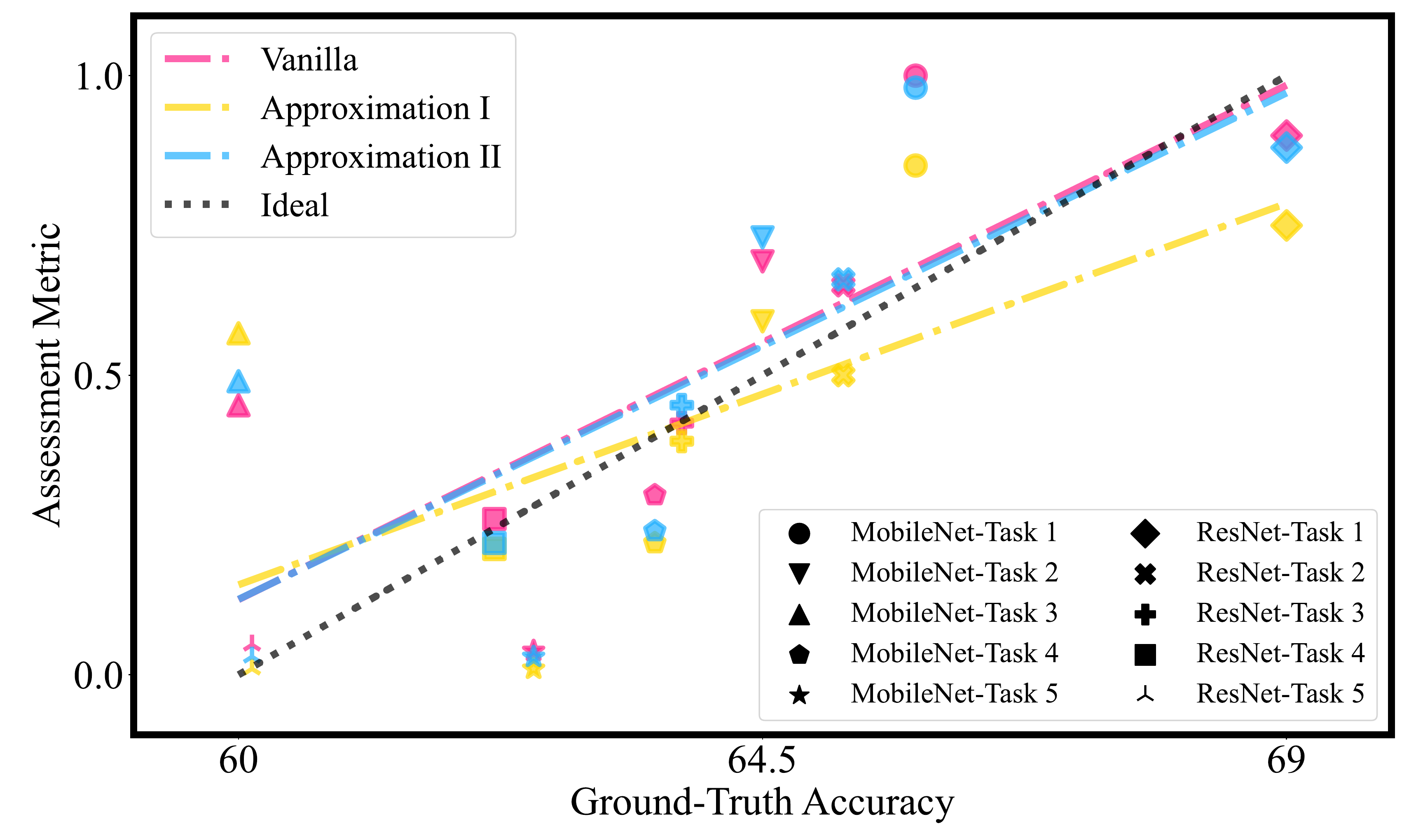}
	\caption{\small Assessment metrics computed using vanilla formulation, Approximation \textbf{I}, and \textbf{II}. We have normalized the metrics into $[0,1]$. We fit a linear regression model for each metric to check its correlation with ground-truth accuracy. The black dotted line is the ideal model, i.e., the metric is completely correlated with the ground-truth accuracy. Different colors stand for different metrics. Different marker shapes denote different teachers.}
	\label{Figure:approximation}
\end{figure}

\begin{table}[t]
\centering
\caption{\small KL divergence between the outputs of $f_{\mathrm{S}}^{h}$ and $f_{\mathrm{F}}^{h}$.}
\begin{tabular}{@{}l|ccccc@{}}
\toprule
Teacher Index $h$                                     & 1    & 2    & 3    & 4    & 5    \\ \midrule
$\mathbb{KL}(f_{\mathrm{S}}^{h}||f_{\mathrm{F}}^{h})$ & 0.32 & 0.41 & 0.39 & 0.47 & 0.53 \\ \midrule\midrule
Teacher Index $h$                                     & 6    & 7    & 8    & 9    & 10   \\ \midrule
$\mathbb{KL}(f_{\mathrm{S}}^{h}||f_{\mathrm{F}}^{h})$ & 0.36 & 0.42 & 0.47 & 0.56 & 0.59 \\ \bottomrule
\end{tabular}
\label{Table:kl}
\end{table}

\begin{figure}[t]
	\centering
	\begin{minipage}[b]{0.48\columnwidth}
		\centering
		\includegraphics[width=\linewidth]{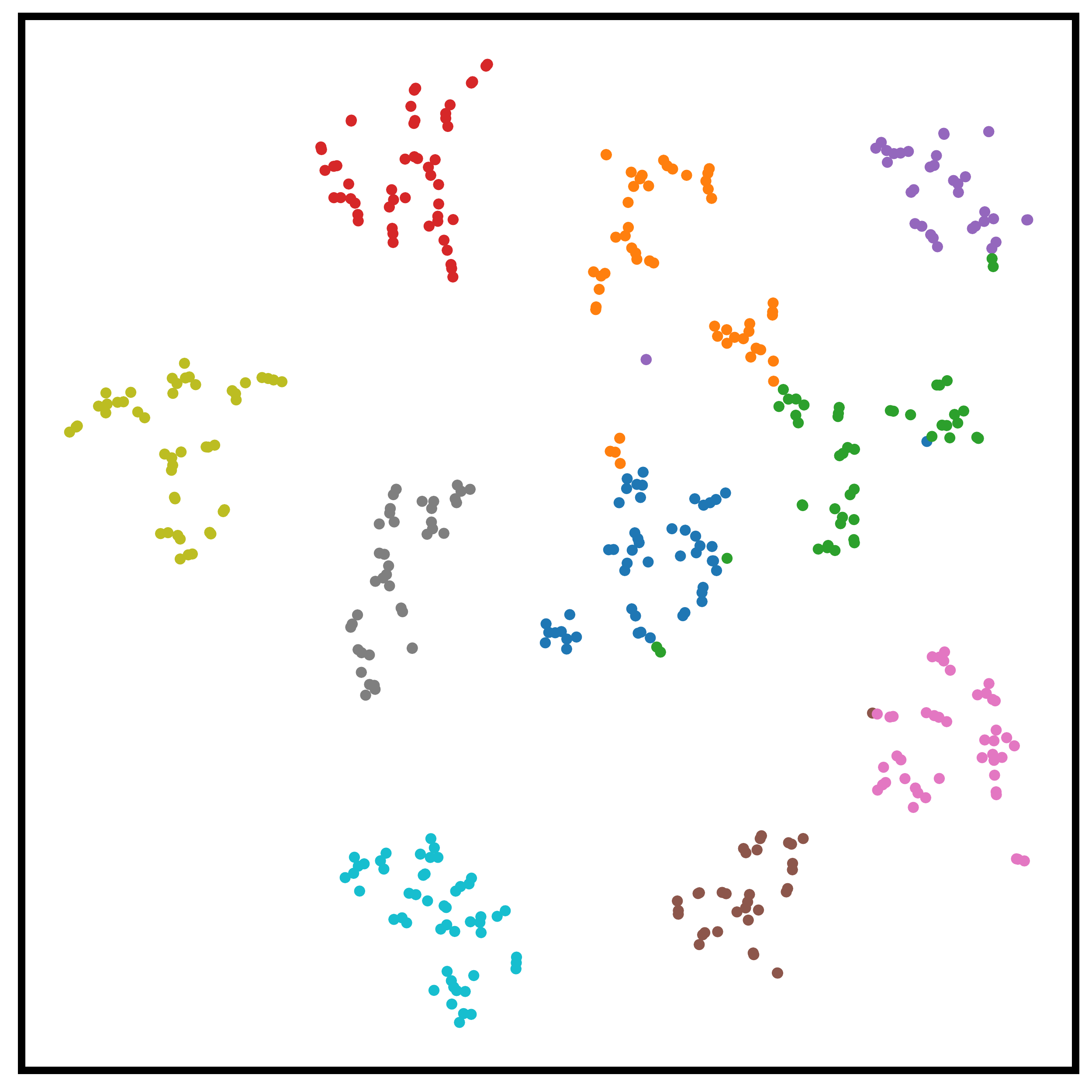}
		\subcaption{\small Embedding of $f_{\mathrm{S}}^{1}$.}
		\label{Figure:tsne_true}
	\end{minipage}
	\begin{minipage}[b]{0.48\linewidth}
		\centering
		\includegraphics[width=\linewidth]{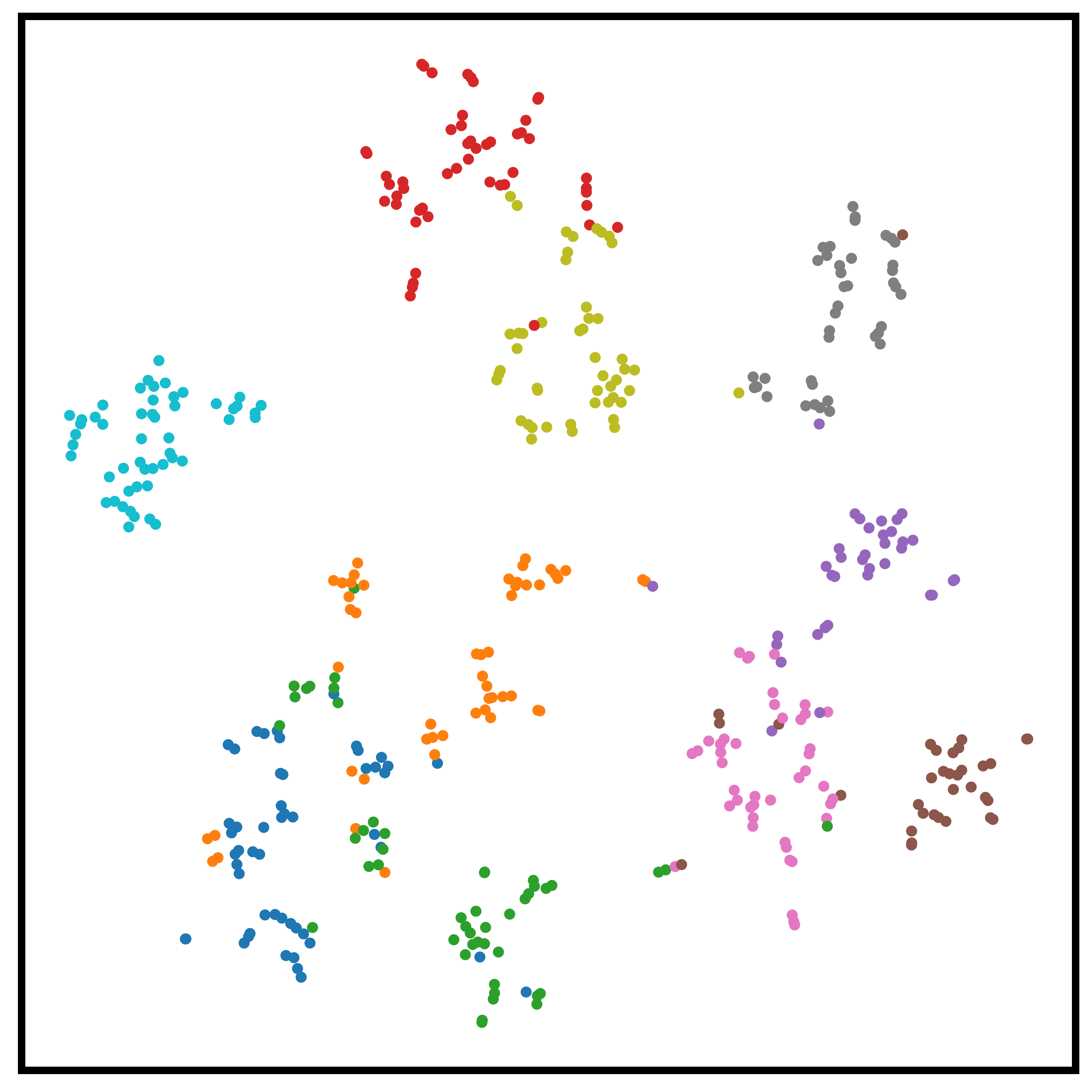}
		\subcaption{\small Embedding of $f_{\mathrm{F}}^{1}$.}
		\label{Figure:tsne_fictitious}
	\end{minipage}
	\caption{\small Visualization of instance representations extracted by the true student and the fictitious student. For simplicity, we only show the figures corresponding to the first teacher and randomly sample $10$ classes from Task 1.}
	\label{Figure:tsne}
\end{figure}

\subsubsection{Influence of Approximations}
\label{Subsubsection:experiment_ta_approximation}
In \secref{Section:assessment}, we have proposed two approximations to compute \equref{Equation:metric}, and in this part we study the influence of these approximations. For simplicity, we fix the student's task to Task 1 of CUB and check the assessment metrics of $10$ teachers using three different computation methods.

In \figref{Figure:approximation}, we show the values of $\{\mathcal{M}(f_{\mathrm{T}}^{h})\}_{h=1}^{10}$ using vanilla formulation, Approximation \textbf{I}, and Approximation \textbf{II} and further fit a linear regression model for each metric to check the correlation between the metric and the ground-truth accuracy. We can see that vanilla formulation and Approximation \textbf{II} both induce high correlation between the ground-truth accuracy and the metric, and they behave similarly to the ideal metric. On the contrary, the performance of Approximation \textbf{I} is not satisfying because it only trains a single student for all the teachers during the assessment procedure. Thus, Approximation \textbf{II} achieves a good trade-off between precision and efficiency.

Moreover, we check whether the fictitious student $f_{\mathrm{F}}^{h}$ in Approximation \textbf{II} can mimic the true student $f_{\mathrm{S}}^{h}$. We still fix the student's task to Task 1 and list the KL divergence between the outputs of $f_{\mathrm{S}}^{h}$ and $f_{\mathrm{F}}^{h}$ in \tabref{Table:kl}. The gap between two outputs is acceptable. Teacher $1-5$ are MobileNets trained on Task $1-5$, and teacher $6-10$ are ResNets trained on Task $1-5$. We can see that $\mathbb{KL}(f_{\mathrm{S}}^{h}||f_{\mathrm{F}}^{h})$ increases when the task semantic gap increases, which conforms to our intuition. Instance representations extracted by the two students are also visualized in \figref{Figure:tsne}. We randomly sample $10$ classes from Task 1 and use t-SNE~\cite{TSNE} to reduce the dimension of instance representations to $2$. We can see that the fictitious student can split each class well.

\begin{table}[t]
    \centering
    \caption{\small Time consumption of $4$ metrics.}
    \begin{tabular}{@{}l|cccc@{}}
	\toprule
	Method            & NCE  & LEEP & LogME & Ours \\ \midrule
	Time per Model~(s) & 0.12 & 0.11 & 0.42  & 0.74 \\ \bottomrule
	\end{tabular}
    \label{Table:time}
\end{table}

\subsubsection{Time Consumption}
\label{Subsubsection:experiment_ta_time}
We list the time consumption for computing different assessment metrics in \tabref{Table:time}. NCE and LEEP are two efficient metrics, but they fail to evaluate teacher qualities in our experiment. LogME and our proposed metric are slower but effective. The extra time cost of our metric mainly comes from training the linear classifier. In general, our metric can efficiently select the most relevant teacher.

\begin{figure}[t]
    \centering
    \caption{\small \textbf{Left}: Our proposed metric of $5$ teachers trained on diverse domains for $5$ target tasks. \textbf{Right}: Test accuracy of $5$ students trained with and without the selected teacher.}
    \includegraphics[width=\linewidth]{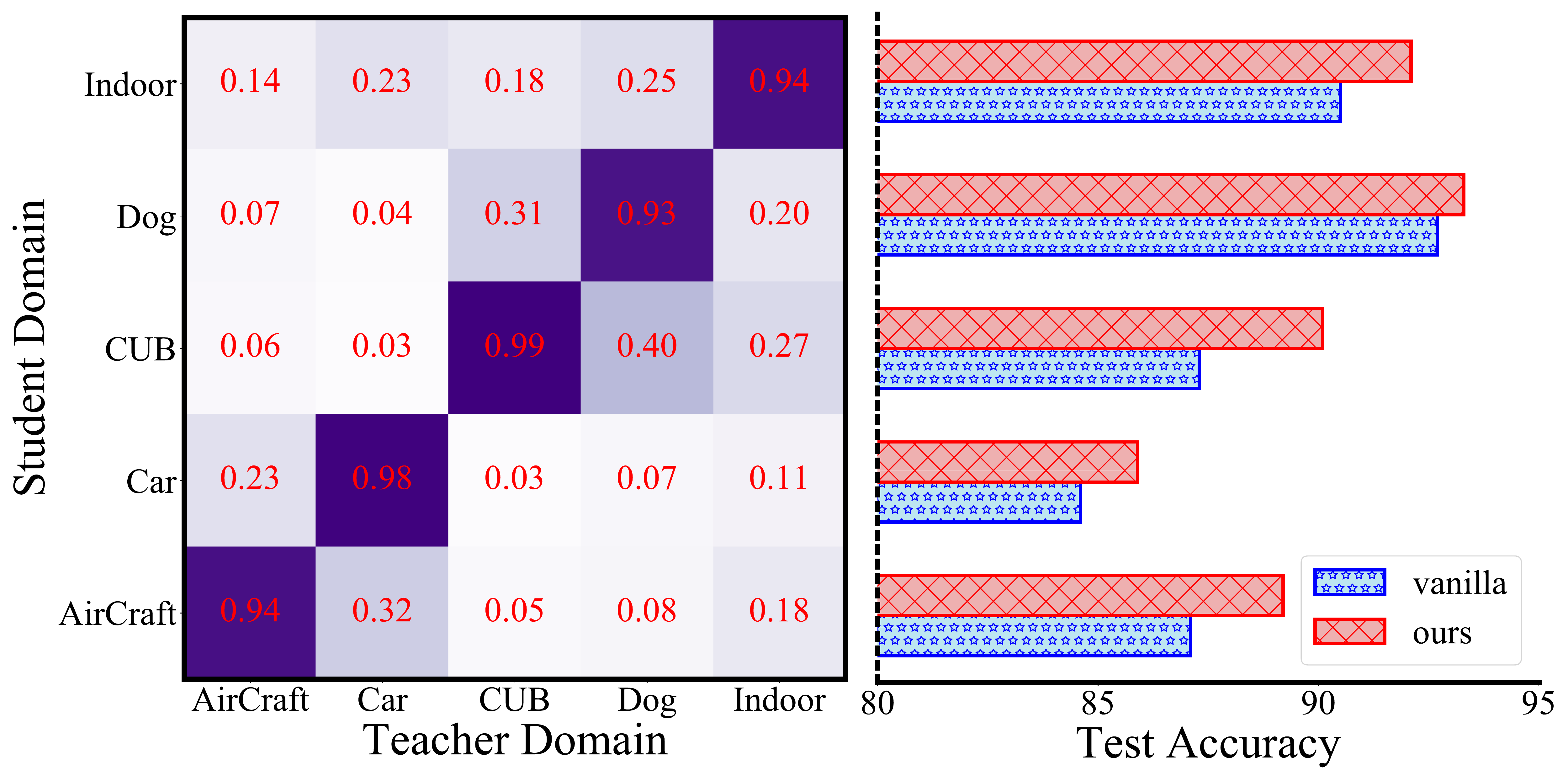}
    \label{Figure:domain}
\end{figure}

\subsubsection{Coarse-Grained Teacher Selection}
\label{Subsubsection:experiment_ta_coarse}
We have tried to select a suitable teacher from a fine-grained repository, i.e., the class subsets of all the teachers are sampled from a same dataset. In this part, we further evaluate our method on a coarse-grained repository containing $5$ teachers trained on MIT Indoor Scenes~(Indoor)~\cite{Indoor}, Stanford Dog~(Dog)~\cite{DOG}, Caltech-UCSD Birds-200-2011~(CUB)~\cite{CUB}, Stanford Car-196~(Car)~\cite{Car}, and FGVC-AirCraft~(AirCraft)~\cite{AirCraft}. These $5$ datasets constitute $5$ separate domains and the semantic gap between them is large.

In fact, this coarse-grained setting is more practical in real-world applications~\cite{LogME,WhichModel} since it is unlikely that the teachers in a model repository are trained on the subsets sampled from a dataset. However, the fine-grained setting constructed by sliding window protocol is harder than the common coarse-grained setting because the semantic similarities between tasks are higher, and testing our method under fine-grained setting is reasonable.

As for the simpler coarse-grained setting, we use our method to select the teacher for each target domain, and train $5$ student models. In \figref{Figure:domain}, we can see that our proposed metric easily finds the corresponding domain, and the student performance is improved a lot after distillation.

\section{Future Work}
\label{Section:future}
In our method, we only reuse the knowledge of one selected teacher, and reusing multiple models simultaneously is an interesting topic~\cite{Zoo-Tuning,B-Tuning,MultiTeacher}. Actually, we can distill multiple teachers by adding Sinkhorn distance terms in our method, but the weights of these terms are not easy to determined. A possible solution is weighting them by their assessment scores, which is a meaningful future work.

Another significant future work is finding vectorized representations of models and tasks~\cite{RKME,learnware,Task2Vec,Dataset2Vec}. Although our proposed method does not require to train the student model repeatedly, all the existing works~\cite{LogME,NCE,LEEP,N-LEEP,OTCE,WhichModel,GBC} including ours need to perform forward process of each teacher, which is infeasible for a huge repository. If each teacher and task is represented by a vector, we can directly match them in the embedding space to realize scalable model recommendation. 

In our paper, we assume that all the teacher models in the repository are well trained. If an offensive client upload a bad model to the teacher repository~\cite{Nasty}, it may be evaluated as a ``related'' model to the target task while does harm to student's performance. Defending adversarial attacks on the repository is also an essential future work.

\appendices

\section{More Experiment Details}
\label{Appendix:details}
\subsection{Experiment Details about \uline{Tab. 2}}
In this experiment, we try three different methods to test a model. We first train a WideResNet-(40,2)~\cite{WideResNet} on CIFAR-10~\cite{CIFAR}. We train the model for $200$ epochs. The model is optimized using SGD optimizer with initial learning rate $0.1$. The learning rate is multiplied by $0.2$ after $50$, $100$, and $150$ epochs. We use batch size $256$. As for the optimizer hyper-parameters, weight decay is set to $0.0005$ and momentum is set to $0.9$. The temperature $\tau$ of softmax function is set to $3$. After training the model, we test the model in three different methods, i.e., using the whole model, using NCM with true class centers, and using NCM with approximate class centers. Results are listed in \uline{Tab. 2}.

\subsection{Experiment Details about \uline{Fig. 3}}
In this experiment, we use a well-trained WideResNet-(40,2)~\cite{WideResNet} as teacher model and a WideResNet-(16,1) as student model. We randomly sample a mini-batch of instances~($256$ instances) from the training split of CIFAR-100, which forms $256$ optimal transport problems. The cost matrix $M$ is computed by the teacher's embedding network $\phi_{\mathrm{T}}$. For the $i$-th instance, the teacher's output probability is $\bm{\rho}_{\tau}(f_{\mathrm{T}}(\mathbf{x}_i))$, and the student's output probability is $\bm{\rho}_{\tau}(f_{\mathrm{S}}(\mathbf{x}_i))$. Hyper-parameter $\tau$ is set to $3$ and hyper-parameter $\epsilon$ is set to $0.1$. We need to solve the dual form of optimal transport problem, i.e., \uline{Equ. (13)}. When using Sinkhorn algorithm to solve \uline{Equ. (13)}, we run $100$ iterations and record $\nabla_{\mathbf{p}_{\mathrm{S}}}^{(t)}S_{\epsilon}(\mathbf{p}_{\mathrm{T}},\mathbf{p}_{\mathrm{S}})$ where $t=0,1,\ldots,100$. $\nabla_{\mathbf{p}_{\mathrm{S}}}S_{\epsilon}(\mathbf{p}_{\mathrm{T}},\mathbf{p}_{\mathrm{S}})$ is set to $\nabla_{\mathbf{p}_{\mathrm{S}}}^{(100)}S_{\epsilon}(\mathbf{p}_{\mathrm{T}},\mathbf{p}_{\mathrm{S}})$. After obtaining this sequence, we draw the three curves in \uline{Fig. 3}.

\section{Proof of \uline{Prop 4.3}}
\label{Appendix:proof}
Our proof is built on previous convergence analysis of potential vectors in Sinkhorn algorithm~\cite{OT-ML} and a fundamental theorem proved in~\cite{Foundation}. In order to give the detailed proof of \uline{Prop. 4.3}, we first introduce some basic concepts.

\begin{definition}[Hilbert Projective Metric]
Let $\mathbf{x}$ and $\mathbf{y}$ be two vectors with positive values, i.e., $\mathbf{x},\mathbf{y}\in\mathbb{R}_{++}^{R}$. The Hilbert projective metric over $\mathbf{x}$ and $\mathbf{y}$ is defined as \equref{Equation:hilbert}:
\begin{equation}
d_{\mathrm{HP}}(\mathbf{x},\mathbf{y})=\log\max_{i,j}\frac{\mathbf{x}_i\mathbf{y}_j}{\mathbf{x}_j\mathbf{y}_i}\;.
\label{Equation:hilbert}
\end{equation}
\end{definition}

We can see that $d_{\mathrm{HP}}(\mathbf{x},\mathbf{y})=0$ if and only if there exists some $t>0$ such that $\mathbf{x}=t\mathbf{y}$. An important property of Hilbert projective metric is invariance with respect to element-wise division, i.e., $d_{\mathrm{HP}}(\mathbf{x},\mathbf{y})=d_{\mathrm{HP}}(\mathbf{x}./\mathbf{y},\mathbf{1})=d_{\mathrm{HP}}(\mathbf{1}./\mathbf{x},\mathbf{1}./\mathbf{y})$. This can be easily proven from the definition of Hilbert projective metric. Besides, we have $d_{\mathrm{HP}}(\mathbf{x},\mathbf{y})=\|\log(\mathbf{x})-\log(\mathbf{y})\|_{\mathrm{var}}$ which is the variation seminorm of the difference between logarithm of two vectors. This relationship can be shown based on the definitions of variation seminorm and Hilbert projective metric.

\begin{theorem}[Proved by~\cite{Foundation}]
Let $K\in\mathbb{R}_{++}^{R_1\times R_2}$ be a matrix with positive values. Define $\psi(K)=\max_{i,j,k,l}\frac{K_{ik}K_{jl}}{K_{jk}K_{il}}$ and $\kappa(K)=\frac{\sqrt{\psi(K)}-1}{\sqrt{\psi(K)}+1}$. For any pair of vectors $\mathbf{x},\mathbf{y}\in\mathbb{R}_{++}^{R_2}$,
\begin{equation}
d_{\mathrm{HP}}(K\mathbf{x},K\mathbf{y})\leq \kappa(K)d_{\mathrm{HP}}(\mathbf{x},\mathbf{y})\;.
\label{Equation:foundation}
\end{equation}
\label{Theorem:foundation}
\end{theorem}

This theorem enables us to bound the Hilbert projective metric between two linear products by the Hilbert projective metric between two vectors along with a constant about matrix $K$, and this theorem is a cornerstone of our proof.

Recall that Sinkhorn's fixed point iteration requires us to update $\mathbf{u}^{(t+1)}=\frac{\mathbf{p}_{\mathrm{T}}}{K\mathbf{v}^{(t)}}$ and $\mathbf{v}^{(t+1)}=\frac{\mathbf{p}_{\mathrm{S}}}{K^\top\mathbf{u}^{(t+1)}}$, and the gradient w.r.t. student's output is $\nabla_{\mathbf{p}_{\mathrm{S}}}S_{\epsilon}(\mathbf{p}_{\mathrm{T}},\mathbf{p}_{\mathrm{S}})$. From the derivation of Sinkhorn algorithm, we have $\nabla_{\mathbf{p}_{\mathrm{S}}}S_{\epsilon}(\mathbf{p}_{\mathrm{T}},\mathbf{p}_{\mathrm{S}})=\epsilon\log\mathbf{v}^\star$, and we can approximate the gradient with $\epsilon\log\mathbf{v}^{(t)}$ if only $t$ iterations are performed. Now we want to bound the difference between $\nabla^{(t+1)}_{\mathbf{p}_{\mathrm{S}}}S_{\epsilon}(\mathbf{p}_{\mathrm{T}},\mathbf{p}_{\mathrm{S}})$ and $\nabla_{\mathbf{p}_{\mathrm{S}}}S_{\epsilon}(\mathbf{p}_{\mathrm{T}},\mathbf{p}_{\mathrm{S}})$ with the difference between $\nabla^{(t)}_{\mathbf{p}_{\mathrm{S}}}S_{\epsilon}(\mathbf{p}_{\mathrm{T}},\mathbf{p}_{\mathrm{S}})$ and $\nabla_{\mathbf{p}_{\mathrm{S}}}S_{\epsilon}(\mathbf{p}_{\mathrm{T}},\mathbf{p}_{\mathrm{S}})$, we have
\begin{equation}
\begin{aligned}
  & \|\nabla^{(t+1)}_{\mathbf{p}_{\mathrm{S}}}S_{\epsilon}(\mathbf{p}_{\mathrm{T}},\mathbf{p}_{\mathrm{S}})-\nabla_{\mathbf{p}_{\mathrm{S}}}S_{\epsilon}(\mathbf{p}_{\mathrm{T}},\mathbf{p}_{\mathrm{S}})\|_{\mathrm{var}} \\
=\; & \|\epsilon\log\mathbf{v}^{(t+1)}-\epsilon\log\mathbf{v}^\star\|_{\mathrm{var}} =\;\epsilon\;\|\log\mathbf{v}^{(t+1)}-\log\mathbf{v}^\star\|_{\mathrm{var}} \\
=\; & \epsilon\;d_{\mathrm{HP}}(\mathbf{v}^{(t+1)},\mathbf{v}^\star) =\; \epsilon\;d_{\mathrm{HP}}\left(\frac{\mathbf{p}_{\mathrm{S}}}{K^\top\mathbf{u}^{(t+1)}},\frac{\mathbf{p}_{\mathrm{S}}}{K^\top\mathbf{u}^\star}\right) \\
=\; & \epsilon\;d_{\mathrm{HP}}(K^\top\mathbf{u}^{(t+1)},K^\top\mathbf{u}^\star)\leq\; \epsilon\;\kappa(K)d_{\mathrm{HP}}(\mathbf{u}^{(t+1)},\mathbf{u}^\star)\;.
\end{aligned}
\label{Equation:proof2-1}
\end{equation}
Similarly, we have
\begin{equation}
\begin{aligned}
 & \epsilon\;d_{\mathrm{HP}}(\mathbf{u}^{(t+1)},\mathbf{u}^\star) \\
=\; & \;\epsilon\;d_{\mathrm{HP}}\left(\frac{\mathbf{p}_{\mathrm{T}}}{K\mathbf{v}^{(t)}},\frac{\mathbf{p}_{\mathrm{T}}}{K\mathbf{v}^\star}\right) = \epsilon\;d_{\mathrm{HP}}(K\mathbf{v}^{(t)},K\mathbf{v}^\star)\\
\leq\;& \kappa(K)\epsilon\;d_{\mathrm{HP}}(\mathbf{v}^{(t)},\mathbf{v}^\star)=\kappa(K)\epsilon\;\|\log\mathbf{v}^{(t)}-\log\mathbf{v}^\star\|_{\mathrm{var}} \\
=\;& \kappa(K)\|\epsilon\log\mathbf{v}^{(t)}-\epsilon\log\mathbf{v}^\star\| \\
=\;&\kappa(K)\|\nabla^{(t)}_{\mathbf{p}_{\mathrm{S}}}S_{\epsilon}(\mathbf{p}_{\mathrm{T}},\mathbf{p}_{\mathrm{S}})-\nabla_{\mathbf{p}_{\mathrm{S}}}S_{\epsilon}(\mathbf{p}_{\mathrm{T}},\mathbf{p}_{\mathrm{S}})\|_{\mathrm{var}}\;.
\end{aligned}
\label{Equation:proof2-2}
\end{equation}
Substituting \equref{Equation:proof2-2} into \equref{Equation:proof2-1} yields that
\begin{equation}
\frac
{\|\nabla_{\mathbf{p}_{\mathrm{S}}}^{(t+1)}S_{\epsilon}(\mathbf{p}_{\mathrm{T}},\mathbf{p}_{\mathrm{S}})-\nabla_{\mathbf{p}_{\mathrm{S}}}S_{\epsilon}(\mathbf{p}_{\mathrm{T}},\mathbf{p}_{\mathrm{S}})\|_{\mathrm{var}}}
{\|\nabla_{\mathbf{p}_{\mathrm{S}}}^{(t)}S_{\epsilon}(\mathbf{p}_{\mathrm{T}},\mathbf{p}_{\mathrm{S}})-\nabla_{\mathbf{p}_{\mathrm{S}}}S_{\epsilon}(\mathbf{p}_{\mathrm{T}},\mathbf{p}_{\mathrm{S}})\|_{\mathrm{var}}}\leq\kappa(K)^{2}\;,
\end{equation}
which ends the proof of \uline{Prop. 4.3}.

\section{Complete Experiment Results}
\label{Appendix:complete}
\subsection{Complete Data in \uline{Fig.~5}}
In this part, we list all values for drawing \uline{Fig.~5} in the main body. \tabref{Table:gkd_cifar100} and \tabref{Table:gkd_cub200} are results of generalized knowledge distillation on CIFAR-100 and CUB respectively.

\begin{table}[t]
\centering
\caption{\small Average test accuracies of the students on CIFAR-100. Teacher architecture is WideResNet-(40,2). The class subset of teacher is fixed while the class subset of student changes. Best results are in \textbf{bold}.}
\scalebox{0.9}{
\begin{tabular}{@{}ccccc@{}}
\toprule
\multicolumn{5}{c}{Class Overlap Ratio $g$ = $0\%$}                                 \\ \midrule
\multicolumn{1}{c|}{(Depth, Width)}  & (40, 2) & (16, 2) & (40, 1) & (16, 1) \\ \midrule
\multicolumn{1}{c|}{Student}         & 81.02   & 78.94   & 78.98   & 73.70   \\
\multicolumn{1}{c|}{RKD}             & 81.46   & 79.23   & 78.80   & 73.45   \\
\multicolumn{1}{c|}{AML}             & 79.99   & 79.11   & 78.99   & 73.68   \\
\multicolumn{1}{c|}{ReFilled}        & \bf 82.60   & 80.70   & 80.18   & \bf 74.42   \\
\multicolumn{1}{c|}{MGD}             & 79.24   & 79.31   & 77.74   & 73.79   \\
\multicolumn{1}{c|}{WCoRD}           & 80.29   & 80.38   & 77.83   & 73.90   \\
\multicolumn{1}{c|}{Ours}         & 81.75   & \bf 80.91   & \bf 81.04   & 74.33   \\ \midrule \midrule
\multicolumn{5}{c}{Class Overlap Ratio $g$ = $20\%$}                                \\ \midrule
\multicolumn{1}{c|}{(Depth, Width)}  & (40, 2) & (16, 2) & (40, 1) & (16, 1) \\ \midrule
\multicolumn{1}{c|}{Student}         & 80.34   & 76.12   & 75.43   & 70.84   \\
\multicolumn{1}{c|}{RKD}             & 80.56   & 76.77   & 75.82   & 71.05   \\
\multicolumn{1}{c|}{AML}             & 80.05   & 76.95   & 75.37   & 70.79   \\
\multicolumn{1}{c|}{ReFilled}        & \bf 81.40   & 77.82   & 77.24   & 72.28   \\
\multicolumn{1}{c|}{MGD}             & 80.25   & 76.25   & 75.57   & 71.22   \\
\multicolumn{1}{c|}{WCoRD}           & 80.50   & 77.19   & 75.77   & 71.52   \\
\multicolumn{1}{c|}{Ours}         & 81.28   & \bf 78.14   & \bf 77.84   & \bf 72.45   \\ \midrule \midrule
\multicolumn{5}{c}{Class Overlap Ratio $g$ = $40\%$}                                \\ \midrule
\multicolumn{1}{c|}{(Depth, Width)}  & (40, 2) & (16, 2) & (40, 1) & (16, 1) \\ \midrule
\multicolumn{1}{c|}{Student}         & 78.86   & 75.67   & 74.98   & 69.36   \\
\multicolumn{1}{c|}{RKD}             & 79.33   & 75.58   & 74.69   & 69.07   \\
\multicolumn{1}{c|}{AML}             & 79.98   & 75.84   & 74.28   & 68.87   \\
\multicolumn{1}{c|}{ReFilled}        & 80.40   & \bf 76.26   & 75.62   & 70.08   \\
\multicolumn{1}{c|}{MGD}             & 79.44   & 75.24   & 74.08   & 69.13   \\
\multicolumn{1}{c|}{WCoRD}           & 81.18   & 76.22   & 74.50   & 69.37   \\
\multicolumn{1}{c|}{Ours}         & \bf 81.33   & 76.14   & \bf 75.99   & \bf 70.90   \\ \midrule \midrule
\multicolumn{5}{c}{Class Overlap Ratio $g$ = $60\%$}                                \\ \midrule
\multicolumn{1}{c|}{(Depth, Width)}  & (40, 2) & (16, 2) & (40, 1) & (16, 1) \\ \midrule
\multicolumn{1}{c|}{Student}         & 78.90   & 76.37   & 75.14   & 68.48   \\
\multicolumn{1}{c|}{RKD}             & 78.69   & 76.20   & 75.50   & 68.23   \\
\multicolumn{1}{c|}{AML}             & 79.32   & 76.45   & 75.23   & 68.64   \\
\multicolumn{1}{c|}{ReFilled}        & \bf 80.66   & 76.66   & 76.52   & 69.50   \\
\multicolumn{1}{c|}{MGD}             & 78.95   & 75.55   & 75.03   & 68.30   \\
\multicolumn{1}{c|}{WCoRD}           & 78.92   & 75.25   & 76.48   & 69.72   \\
\multicolumn{1}{c|}{Ours}         & 80.51   & \bf 76.92   & \bf 77.87   & \bf 70.34   \\ \midrule \midrule
\multicolumn{5}{c}{Class Overlap Ratio $g$ = $80\%$}                                \\ \midrule
\multicolumn{1}{c|}{(Depth, Width)}  & (40, 2) & (16, 2) & (40, 1) & (16, 1) \\ \midrule
\multicolumn{1}{c|}{Student}         & 80.50   & 77.43   & 76.96   & 72.16   \\
\multicolumn{1}{c|}{RKD}             & 81.21   & 77.65   & 77.34   & 72.05   \\
\multicolumn{1}{c|}{AML}             & 81.06   & 77.20   & 77.06   & 72.35   \\
\multicolumn{1}{c|}{ReFilled}        & \bf 82.56   & \bf 78.76   & 79.28   & 73.92   \\
\multicolumn{1}{c|}{MGD}             & 81.46   & 76.60   & 77.86   & 71.78   \\
\multicolumn{1}{c|}{WCoRD}           & 81.54   & 77.16   & 77.80   & 73.09   \\
\multicolumn{1}{c|}{Ours}         & 82.33   & 78.29   & \bf 80.08   & \bf 74.17   \\ \midrule \midrule
\multicolumn{5}{c}{Class Overlap Ratio $g$ = $100\%$}                               \\ \midrule
\multicolumn{1}{c|}{(Depth, Width)}  & (40, 2) & (16, 2) & (40, 1) & (16, 1) \\ \midrule
\multicolumn{1}{c|}{Student}         & 80.66   & 77.94   & 76.35   & 71.56   \\
\multicolumn{1}{c|}{RKD}             & 80.52   & 78.03   & 76.82   & 72.04   \\
\multicolumn{1}{c|}{AML}             & 80.73   & 78.24   & 77.15   & 71.79   \\
\multicolumn{1}{c|}{ReFilled}        & 81.58   & \bf 78.60   & \bf 78.04   & 73.52   \\
\multicolumn{1}{c|}{MGD}             & 80.13   & 77.64   & 76.35   & 71.09   \\
\multicolumn{1}{c|}{WCoRD}           & 81.50   & 77.78   & 76.85   & 71.94   \\
\multicolumn{1}{c|}{Ours}         & \bf 82.32   & 78.33   & 77.85   & \bf 74.17   \\ \bottomrule
\end{tabular}}
\label{Table:gkd_cifar100}
\end{table}

\begin{table}[t]
\centering
\caption{\small Average test accuracies of the students on CUB. Teacher architecture is MobileNet-1.0. The class subset of teacher is fixed while the class subset of student changes. Best results are in \textbf{bold}.}
\scalebox{0.9}{
\begin{tabular}{@{}ccccc@{}}
\toprule
\multicolumn{5}{c}{Overlap Ratio = $0\%$}                          \\ \midrule
\multicolumn{1}{c|}{Width Multiplier} & 1.0   & 0.75  & 0.5   & 0.25  \\ \midrule
\multicolumn{1}{c|}{Student}          & 71.25 & 67.56 & 66.85 & 64.48 \\
\multicolumn{1}{c|}{RKD}              & 72.24 & 68.42 & 66.85 & 65.74 \\
\multicolumn{1}{c|}{AML}              & 72.86 & 68.79 & 68.59 & 66.83 \\
\multicolumn{1}{c|}{ReFilled}         & \bf 75.13 & \bf 71.67 & 71.06 & 68.22 \\
\multicolumn{1}{c|}{MGD}              & 72.24 & 69.43 & 69.78 & 67.13  \\
\multicolumn{1}{c|}{WCoRD}            & 72.78 & 69.81 & 69.20 & 67.70  \\
\multicolumn{1}{c|}{Ours}          & 75.03 & 71.44 & \bf 71.35 & \bf 69.66 \\ \midrule \midrule
\multicolumn{5}{c}{Overlap Ratio = $25\%$}                         \\ \midrule
\multicolumn{1}{c|}{Width Multiplier} & 1.0   & 0.75  & 0.5   & 0.25  \\ \midrule
\multicolumn{1}{c|}{Student}          & 71.30 & 71.08 & 68.56 & 65.71 \\
\multicolumn{1}{c|}{RKD}              & 72.07 & 71.70 & 68.56 & 66.43 \\
\multicolumn{1}{c|}{AML}              & 72.35 & 72.05 & 70.37 & 67.20 \\
\multicolumn{1}{c|}{ReFilled}         & 75.09 & 73.92 & \bf 72.99 & 70.04 \\
\multicolumn{1}{c|}{MGD}              & 73.41 & 72.52 & 69.99 & 67.99  \\
\multicolumn{1}{c|}{WCoRD}            & 73.97 & 72.99 & 70.89 & 67.28  \\
\multicolumn{1}{c|}{Ours}          & \bf 75.52 & \bf 74.33 & 72.85 & \bf 71.12 \\ \midrule \midrule
\multicolumn{5}{c}{Overlap Ratio = $50\%$}                         \\ \midrule
\multicolumn{1}{c|}{Width Multiplier} & 1.0   & 0.75  & 0.5   & 0.25  \\ \midrule
\multicolumn{1}{c|}{Student}          & 68.20 & 66.11 & 65.23 & 62.26 \\
\multicolumn{1}{c|}{RKD}              & 68.72 & 66.82 & 65.58 & 62.79 \\
\multicolumn{1}{c|}{AML}              & 67.94 & 67.34 & 66.29 & 63.64 \\
\multicolumn{1}{c|}{ReFilled}         & 70.25 & \bf 68.39 & \bf 68.50 & 65.33 \\
\multicolumn{1}{c|}{MGD}              & 68.10 & 67.90 & 66.81 & 64.24  \\
\multicolumn{1}{c|}{WCoRD}            & 68.45 & 67.42 & 66.89 & 63.95  \\
\multicolumn{1}{c|}{Ours}          & \bf 70.94 & 68.13 & 67.78 & \bf 66.17 \\ \midrule \midrule
\multicolumn{5}{c}{Overlap Ratio = $75\%$}                         \\ \midrule
\multicolumn{1}{c|}{Width Multiplier} & 1.0   & 0.75  & 0.5   & 0.25  \\ \midrule
\multicolumn{1}{c|}{Student}          & 65.53 & 66.73 & 64.10 & 60.81 \\
\multicolumn{1}{c|}{RKD}              & 65.89 & 67.28 & 64.66 & 61.35 \\
\multicolumn{1}{c|}{AML}              & 66.32 & 66.92 & 65.03 & 62.09 \\
\multicolumn{1}{c|}{ReFilled}         & 67.28 & 68.35 & \bf 66.72 & 63.03 \\
\multicolumn{1}{c|}{MGD}              & 67.01 & 67.56 & 65.43 & 63.29  \\
\multicolumn{1}{c|}{WCoRD}            & 66.89 & 67.60 & 66.00 & 63.11  \\
\multicolumn{1}{c|}{Ours}          & \bf 67.79 & \bf 69.20 & 66.46 & \bf 64.17 \\ \midrule \midrule
\multicolumn{5}{c}{Overlap Ratio = $100\%$}                        \\ \midrule
\multicolumn{1}{c|}{Width Multiplier} & 1.0   & 0.75  & 0.5   & 0.25  \\ \midrule
\multicolumn{1}{c|}{Student}          & 67.76 & 67.98 & 64.91 & 62.17 \\
\multicolumn{1}{c|}{RKD}              & 67.23 & 68.25 & 65.73 & 62.04 \\
\multicolumn{1}{c|}{AML}              & 67.06 & 68.35 & 66.27 & 62.69 \\
\multicolumn{1}{c|}{ReFilled}         & 68.77 & 69.10 & 68.44 & 63.33 \\
\multicolumn{1}{c|}{MGD}              & 67.46 & 68.35 & 67.31 & 62.50 \\
\multicolumn{1}{c|}{WCoRD}            & 67.72 & 68.75 & 68.07 & 63.24 \\
\multicolumn{1}{c|}{Ours}          & \bf 70.03 & \bf 70.87 & \bf 69.55 & \bf 65.72 \\ \bottomrule
\end{tabular}}
\label{Table:gkd_cub200}
\end{table}

\subsection{Complete Data in \uline{Fig.~7}}
In this section, we list all the values for drawing \uline{Fig.~7} in the main body. For each target task~(each column of \uline{Fig.~7}), we give the accuracies of the students trained with assistance of $10$ different teachers~(x-axis values in each column of \uline{Fig.~7}) and four metrics to evaluate $10$ teachers~(y-axis values in each column). All the metrics are normalized into $[0,1]$ for convenience. These values are listed in \tabref{Table:cub_ta_task1}, \tabref{Table:cub_ta_task2}, \tabref{Table:cub_ta_task3}, \tabref{Table:cub_ta_task4}, and \tabref{Table:cub_ta_task5}.

\begin{table}[t]
\caption{\small Values for drawing the first column of \uline{Fig.~7}.}
\centering
\begin{tabular}{@{}cccccc@{}}
\toprule
\multicolumn{6}{c}{Target Task 1}                                                                                                          \\ \midrule
\multicolumn{1}{c|}{Teacher Id} & \multicolumn{1}{c|}{Accuracy} & NCE~\cite{NCE} & LEEP~\cite{LEEP} & LogME~\cite{LogME} & Ours \\ \midrule
\multicolumn{1}{c|}{1}          & \multicolumn{1}{c|}{65.72}            & 1.00           & 0.91             & 0.99               & 1.00    \\
\multicolumn{1}{c|}{2}          & \multicolumn{1}{c|}{64.47}            & 0.28           & 0.68             & 0.73               & 0.73    \\
\multicolumn{1}{c|}{3}          & \multicolumn{1}{c|}{60.19}            & 0.80           & 0.49             & 0.50               & 0.49    \\
\multicolumn{1}{c|}{4}          & \multicolumn{1}{c|}{63.59}            & 0.88           & 0.28             & 0.26               & 0.24    \\
\multicolumn{1}{c|}{5}          & \multicolumn{1}{c|}{62.60}            & 0.92           & 0.09             & 0.04               & 0.03    \\
\multicolumn{1}{c|}{6}          & \multicolumn{1}{c|}{68.75}            & 0.99           & 0.09             & 0.82               & 0.88    \\
\multicolumn{1}{c|}{7}          & \multicolumn{1}{c|}{65.13}            & 0.24           & 0.08             & 0.63               & 0.66    \\
\multicolumn{1}{c|}{8}          & \multicolumn{1}{c|}{63.81}            & 0.74           & 0.06             & 0.43               & 0.45    \\
\multicolumn{1}{c|}{9}          & \multicolumn{1}{c|}{62.28}            & 0.80           & 0.04             & 0.22               & 0.22    \\
\multicolumn{1}{c|}{10}         & \multicolumn{1}{c|}{60.30}            & 0.87           & 0.01             & 0.01               & 0.03    \\ \bottomrule
\end{tabular}
\label{Table:cub_ta_task1}
\end{table}

\begin{table}[t]
\caption{\small Values for drawing the second column of \uline{Fig.~7}.}
\centering
\begin{tabular}{@{}cccccc@{}}
\toprule
\multicolumn{6}{c}{Target Task 2}                                                                                                          \\ \midrule
\multicolumn{1}{c|}{Teacher Id} & \multicolumn{1}{c|}{Accuracy} & NCE~\cite{NCE} & LEEP~\cite{LEEP} & LogME~\cite{LogME} & Ours \\ \midrule
\multicolumn{1}{c|}{1}          & \multicolumn{1}{c|}{64.17}            & 0.22           & 0.67             & 0.71               & 0.70    \\
\multicolumn{1}{c|}{2}          & \multicolumn{1}{c|}{69.59}            & 1.00           & 0.92             & 0.99               & 0.99    \\
\multicolumn{1}{c|}{3}          & \multicolumn{1}{c|}{64.21}            & 0.34           & 0.72             & 0.74               & 0.73    \\
\multicolumn{1}{c|}{4}          & \multicolumn{1}{c|}{63.55}            & 0.73           & 0.46             & 0.48               & 0.47    \\
\multicolumn{1}{c|}{5}          & \multicolumn{1}{c|}{59.71}            & 0.86           & 0.28             & 0.26               & 0.25    \\
\multicolumn{1}{c|}{6}          & \multicolumn{1}{c|}{62.78}            & 0.22           & 0.07             & 0.59               & 0.63    \\
\multicolumn{1}{c|}{7}          & \multicolumn{1}{c|}{66.63}            & 0.99           & 0.10             & 0.85               & 0.90    \\
\multicolumn{1}{c|}{8}          & \multicolumn{1}{c|}{64.10}            & 0.33           & 0.08             & 0.63               & 0.66    \\
\multicolumn{1}{c|}{9}          & \multicolumn{1}{c|}{60.59}            & 0.59           & 0.07             & 0.43               & 0.44    \\
\multicolumn{1}{c|}{10}         & \multicolumn{1}{c|}{58.61}            & 0.78           & 0..04            & 0.21               & 0.23    \\ \bottomrule
\end{tabular}
\label{Table:cub_ta_task2}
\end{table}

\begin{table}[t]
\caption{\small Values for drawing the third column of \uline{Fig.~7}.}
\centering
\begin{tabular}{@{}cccccc@{}}
\toprule
\multicolumn{6}{c}{Target Task 3}                                                                                                          \\ \midrule
\multicolumn{1}{c|}{Teacher Id} & \multicolumn{1}{c|}{Accuracy} & NCE~\cite{NCE} & LEEP~\cite{LEEP} & LogME~\cite{LogME} & Ours \\ \midrule
\multicolumn{1}{c|}{1}          & \multicolumn{1}{c|}{66.17}            & 0.80           & 0.47             & 0.46               & 0.45    \\
\multicolumn{1}{c|}{2}          & \multicolumn{1}{c|}{63.36}            & 0.11           & 0.70             & 0.73               & 0.72    \\
\multicolumn{1}{c|}{3}          & \multicolumn{1}{c|}{71.28}            & 1.00           & 0.95             & 1.00               & 0.99    \\
\multicolumn{1}{c|}{4}          & \multicolumn{1}{c|}{64.24}            & 0.36           & 0.62             & 0.69               & 0.69    \\
\multicolumn{1}{c|}{5}          & \multicolumn{1}{c|}{59.51}            & 0.82           & 0.45             & 0.45               & 0.44    \\
\multicolumn{1}{c|}{6}          & \multicolumn{1}{c|}{63.14}            & 0.74           & 0.05             & 0.39               & 0.41    \\
\multicolumn{1}{c|}{7}          & \multicolumn{1}{c|}{66.66}            & 0.20           & 0.08             & 0.63               & 0.66    \\
\multicolumn{1}{c|}{8}          & \multicolumn{1}{c|}{68.97}            & 0.99           & 0.10             & 0.85               & 0.90    \\
\multicolumn{1}{c|}{9}          & \multicolumn{1}{c|}{64.24}            & 0.30           & 0.08             & 0.61               & 0.64    \\
\multicolumn{1}{c|}{10}         & \multicolumn{1}{c|}{65.78}            & 0.77           & 0.06             & 0.37               & 0.39    \\ \bottomrule
\end{tabular}
\label{Table:cub_ta_task3}
\end{table}

\begin{table}[t]
\caption{\small Values for drawing the fourth column of \uline{Fig.~7}.}
\centering
\begin{tabular}{@{}cccccc@{}}
\toprule
\multicolumn{6}{c}{Target Task 4}                                                                                                          \\ \midrule
\multicolumn{1}{c|}{Teacher Id} & \multicolumn{1}{c|}{Accuracy} & NCE~\cite{NCE} & LEEP~\cite{LEEP} & LogME~\cite{LogME} & Ours \\ \midrule
\multicolumn{1}{c|}{1}          & \multicolumn{1}{c|}{71.12}            & 0.88           & 0.25             & 0.23               & 0.21    \\
\multicolumn{1}{c|}{2}          & \multicolumn{1}{c|}{66.81}            & 0.71           & 0.47             & 0.50               & 0.48    \\
\multicolumn{1}{c|}{3}          & \multicolumn{1}{c|}{65.27}            & 0.42           & 0.71             & 0.71               & 0.70    \\
\multicolumn{1}{c|}{4}          & \multicolumn{1}{c|}{72.72}            & 1.00           & 0.88             & 0.97               & 0.98    \\
\multicolumn{1}{c|}{5}          & \multicolumn{1}{c|}{65.71}            & 0.16           & 0.69             & 0.71               & 0.71    \\
\multicolumn{1}{c|}{6}          & \multicolumn{1}{c|}{64.18}            & 0.75           & 0.03             & 0.20               & 0.21    \\
\multicolumn{1}{c|}{7}          & \multicolumn{1}{c|}{66.37}            & 0.50           & 0.06             & 0.44               & 0.45    \\
\multicolumn{1}{c|}{8}          & \multicolumn{1}{c|}{68.89}            & 0.33           & 0.08             & 0.60               & 0.62    \\
\multicolumn{1}{c|}{9}          & \multicolumn{1}{c|}{70.09}            & 0.99           & 0.11             & 0.87               & 0.91    \\
\multicolumn{1}{c|}{10}         & \multicolumn{1}{c|}{68.23}            & 0.08           & 0.10             & 0.61               & 0.64    \\ \bottomrule
\end{tabular}
\label{Table:cub_ta_task4}
\end{table}

\begin{table}[t]
\caption{\small Values for drawing the fifth column of \uline{Fig~7}.}
\centering
\begin{tabular}{@{}cccccc@{}}
\toprule
\multicolumn{6}{c}{Target Task 5}                                                                                                          \\ \midrule
\multicolumn{1}{c|}{Teacher Id} & \multicolumn{1}{c|}{Accuracy} & NCE~\cite{NCE} & LEEP~\cite{LEEP} & LogME~\cite{LogME} & Ours \\ \midrule
\multicolumn{1}{c|}{1}          & \multicolumn{1}{c|}{69.66}            & 0.90           & 0.05             & 0.01               & 0.00    \\
\multicolumn{1}{c|}{2}          & \multicolumn{1}{c|}{66.44}            & 0.86           & 0.25             & 0.27               & 0.26    \\
\multicolumn{1}{c|}{3}          & \multicolumn{1}{c|}{66.12}            & 0.78           & 0.48             & 0.46               & 0.46    \\
\multicolumn{1}{c|}{4}          & \multicolumn{1}{c|}{69.18}            & 0.21           & 0.65             & 0.70               & 0.70    \\
\multicolumn{1}{c|}{5}          & \multicolumn{1}{c|}{73.66}            & 1.00           & 1.00             & 0.99               & 0.99    \\
\multicolumn{1}{c|}{6}          & \multicolumn{1}{c|}{65.57}            & 0.82           & 0.00             & 0.00               & 0.00    \\
\multicolumn{1}{c|}{7}          & \multicolumn{1}{c|}{66.77}            & 0.71           & 0.03             & 0.23               & 0.24    \\
\multicolumn{1}{c|}{8}          & \multicolumn{1}{c|}{68.96}            & 0.70           & 0.06             & 0.38               & 0.40    \\
\multicolumn{1}{c|}{9}          & \multicolumn{1}{c|}{68.52}            & 0.00           & 0.09             & 0.62               & 0.65    \\
\multicolumn{1}{c|}{10}         & \multicolumn{1}{c|}{71.80}            & 0.99           & 0.13             & 0.87               & 0.91    \\ \bottomrule
\end{tabular}
\label{Table:cub_ta_task5}
\end{table}

\ifCLASSOPTIONcompsoc
  \section*{Acknowledgments}
\else
  \section*{Acknowledgment}
\fi
This work is supported by the National Science Foundation of China~(61921006).

\ifCLASSOPTIONcaptionsoff
  \newpage
\fi

\bibliographystyle{IEEEtran}
\bibliography{references}

\end{document}